\documentclass{article}

\usepackage[preprint]{neurips_2025}

\usepackage[utf8]{inputenc} 
\usepackage[T1]{fontenc}    
\usepackage{hyperref}       
\usepackage{url}            
\usepackage{booktabs}       
\usepackage{amsfonts}       
\usepackage{nicefrac}       
\usepackage{microtype}      
\usepackage{xcolor}         
\usepackage{multirow}
\usepackage{tabularx}
\usepackage{makecell}
\usepackage{amsmath}
\usepackage{graphicx}
\usepackage{algorithm}
\usepackage{algorithmic}
\usepackage{wrapfig}
\usepackage{subcaption}

\title{Feed Two Birds with One Scone: \\
	Exploiting Function-Space Regularization for Both\\
	  OOD Robustness and ID Fine-Tuning Performance }

\author{%
  Xiang Yuan \\
  Xi'an Jiaotong University \\
  \texttt{relojeffrey@gmail.com} \\
  \And
  Jun Shu \\
  Xi'an Jiaotong University \\
  \texttt{xjtushujun@gmail.com} \\
  \AND
  Deyu Meng \\
  Xi'an Jiaotong University \\
  \texttt{dymeng@mail.xjtu.edu.cn} \\
  \And
  Zongben Xu \\
  Xi'an Jiaotong University \\
  \texttt{zbxu@mail.xjtu.edu.cn} \\
}

\begin{document}

\maketitle

\begin{abstract}
    Robust fine-tuning aims to achieve competitive in-distribution (ID) performance while maintaining the out-of-distribution (OOD) robustness of a pre-trained model when transferring it to a downstream task. To remedy this, most robust fine-tuning methods aim to preserve the pretrained weights, features, or logits. However, we find that these methods cannot always improve OOD robustness for different model architectures. This is due to the OOD robustness requiring the model function to produce stable prediction for input information of downstream tasks, while existing methods might serve as a poor proxy for the optimization in the function space. Based on this finding, we propose a novel regularization that constrains the distance of fine-tuning and pre-trained model in the function space with the simulated OOD samples, aiming to preserve the OOD robustness of the pre-trained model. Besides, to further enhance the OOD robustness capability of the fine-tuning model, we introduce an additional consistency regularization to promote stable predictions of perturbed samples. Extensive experiments demonstrate our approach could consistently improve both downstream task ID fine-tuning performance and OOD robustness across a variety of CLIP backbones, outperforming existing regularization-based robust fine-tuning methods.
\end{abstract}

\section{Introduction}

Foundation models such as CLIP have achieved impressive performance on diverse domains via pre-train-fine-tuning paradigm \cite{bommasani2021opportunities,radford2021learning,brown2020language}. Leveraging the strong general knowledge acquired through
large-scale pre-training, foundation models can adapt to downstream tasks through zero-shot inference or fine-tuning. However, recent studies have shown that a naive fine-tuning approach comprises foundation models' strong out-of-distribution (OOD) generalization and robustness capability during adaptation to downstream in-distribution (ID) tasks \cite{wortsman2022robust,kumarfine2022}. For example, a model fine-tuned on ImageNet has better accuracy on ID data yet may underperform for OOD data such as ImageNet-A \cite{hendrycks2021natural}, ImageNet-R, or cross-task zero-shot performance.

To ensure robustness under distribution shifts, a wide range of robust fine-tuning methods \cite{wortsman2022robust,goyal2023finetune,lee2022surgical} have been recently developed to adapt models to ID while maintaining strong OOD robustness. 
The key insight of these methods is to explicitly constrain the distance between the fine-tuned and the pre-trained models, so as to maintain OOD robustness capabilities acquired during pre-training when fine-tuning downstream tasks. For example, 
L2-SP method \cite{xuhong2018explicit,mukhotifine} regularized the $\ell2$ distance of weights between
the fine-tuned model and the pre-trained model. 
To better preserve concepts of CLIP model, LDIFS method \cite{mukhotifine} proposed to compute the 
$\ell2$ distance of feature representations between the fine-tuned model and the pre-trained model. CAR-FT \cite{mao2024context} devised the context-awareness regularization to minimize the distance of context distributions induced by fine-tuned/pre-trained CLIP models. Afterward, Lipsum-FT \cite{nam2024lipsum} improved CAR-FT by computing $\ell2$ loss of logits in the context of an arbitrary discriminative model via a random text guidance strategy. Tian et al. \cite{tian2023trainable,tian2023fast} devised a trainable projection method to constrain the parameter space, and recently CaRot \cite{oh2024towards} introduced a self-distillation regularization based on the self-evolving EMA fine-tuning models.

Though these methods could alleviate the degeneracy of models' generalizability and robustness to some extent, they cannot always achieve OOD robustness of pretrained models for different model architectures, as shown in Figure 1. In fact, the OOD robustness of a model should consider how the model function's output changes with respect to varying input information, which is relative to the space of concerned functions themselves. While existing methods are limited because the change in the parameter/feature/logits space only serves as an imperfect proxy for that in function space \cite{benjamin2018measuring}. 

To address the limitation, a natural idea is to minimize distances between fine-tuned/pre-trained CLIP functions in an $L2$ Hilbert space , and thus one can maintain strong OOD robustness by directly limiting how much the
fine-tuning model function changes with respect to the pre-trained model function on the OOD samples. Besides, current insight on OOD robustness mainly attributes to the general knowledge of the pre-trained model, while ignoring the potential contribution of data in downstream tasks. To further enhance the OOD robustness capability of the fine-tuning model, we additionally introduce a consistency regularization in the function space to facilitate OOD generalization inspired by current domain adaptation and domain generalization community \cite{yang2023sample,sagawadistributionally}.

We summarize our main contribution as follows:

1) We point out that existing regularization-based fine-tuning methods do not adequately achieve satisfactory OOD robustness and ID fine-tuning performance simultaneously. 

2) We provide a perspective that robust fine-tuning should rely on the optimization on the function space. 

3) Based on this understanding, we devise a functional regularization robust fine-tuning (\textbf{FRR-FT}) method that minimize the $L2$ function distance of fine-tuning and pre-trained model to better maintain OOD robustness capability of pre-trained CLIP models, and a consistency regularization to promote stable predictions during fine-tuning in function space. 

4) Entensive validation shows that the proposed method consistently improves existing regularization-based robust fine-tuning methods across a variety of CLIP backbones on distribution shift and cross-task zero-shot performance in terms of both OOD robustness and downstream task ID fine-tuning performance.

\begin{figure}[tb]
	\centering
	\label{results_CLIP_B32}
	\begin{minipage}{0.32\textwidth}
		\centering
		\includegraphics[width=\textwidth]{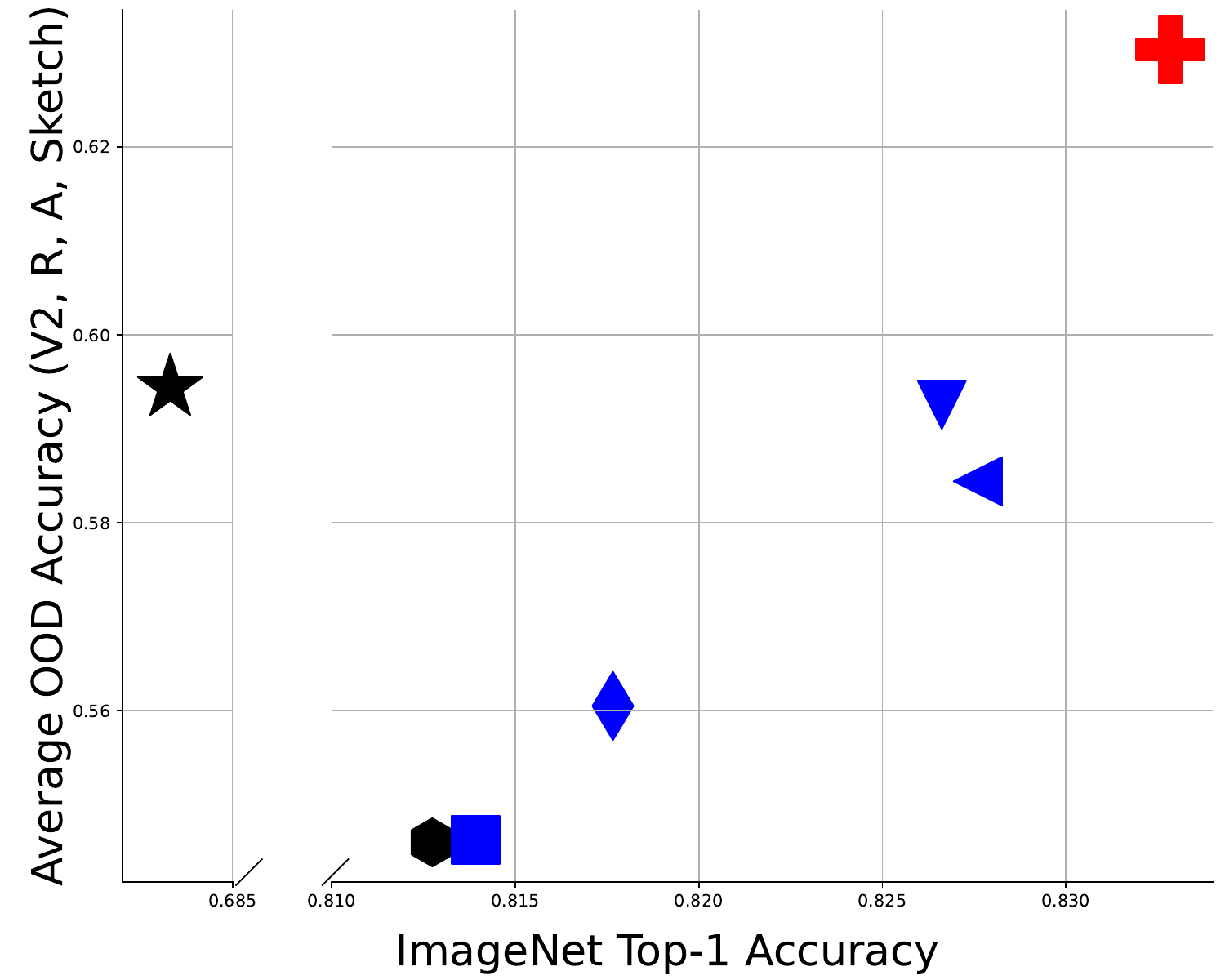}
		
	\end{minipage}
	\hfill
	\begin{minipage}{0.32\textwidth}
		\centering
		\includegraphics[width=\textwidth]{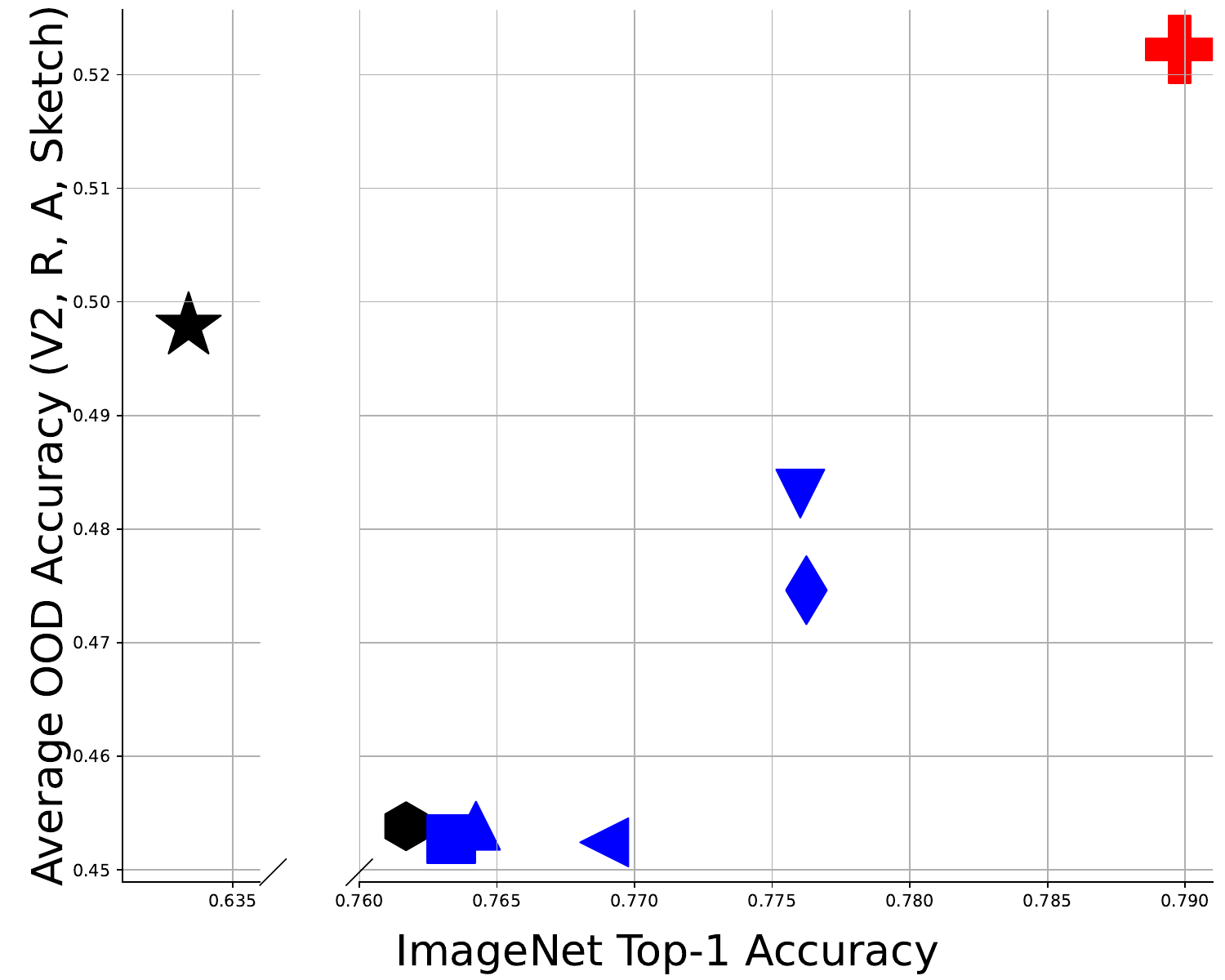}
	\end{minipage}
	\hfill
	\begin{minipage}{0.32\textwidth}
		\centering
		\includegraphics[width=\textwidth]{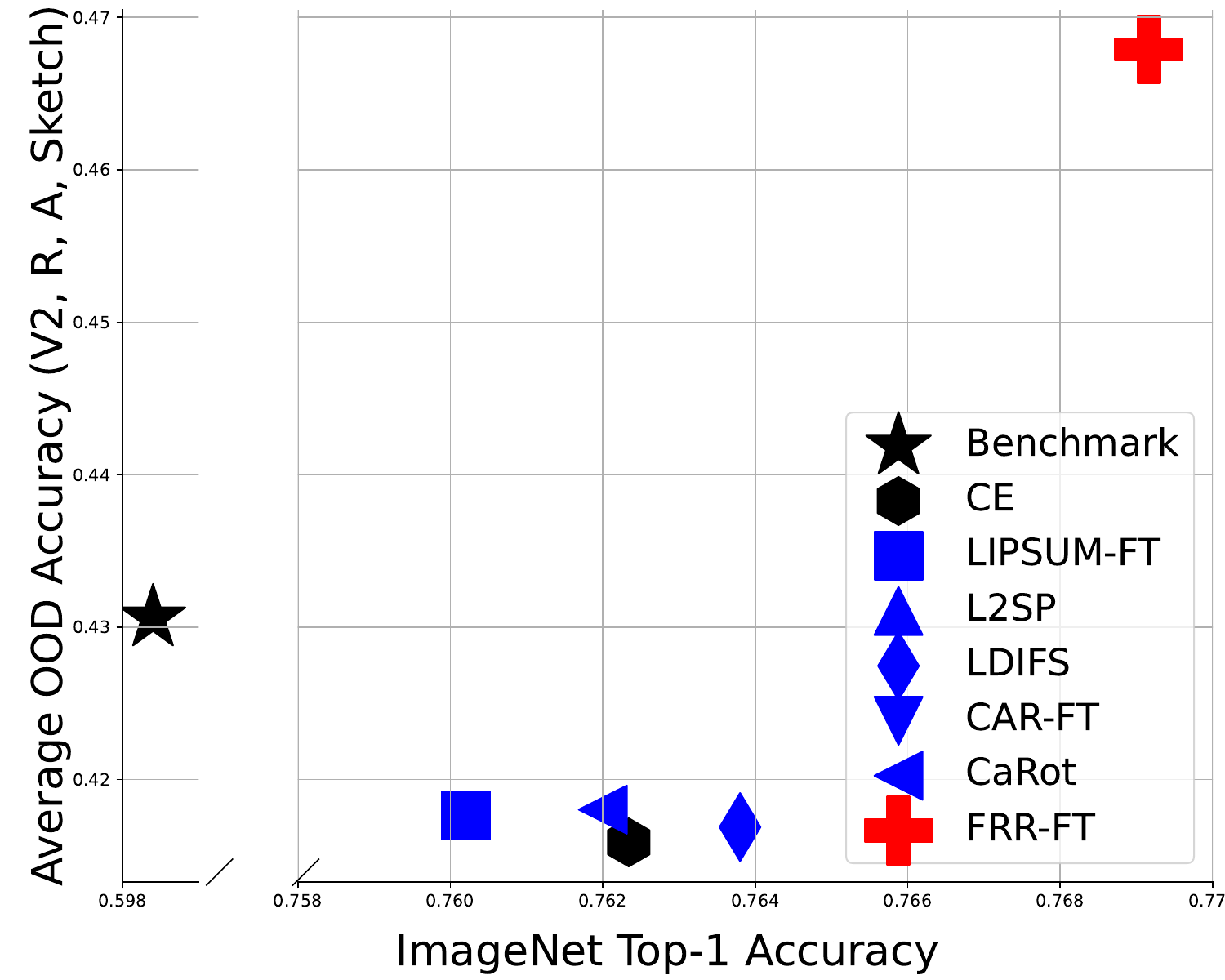}
	\end{minipage} \label{fig1}
	\caption{Performance comparison of existing regularization-based robust fine-tuning methods on ID and OOD datasets under different CLIP architectures including CLIP VIT-B/32 (Left), VIT-B/16 (Middle), CLIP-RN50 (Right).} 
\end{figure}

\section{Preliminary}

\textbf{Classification for OOD tasks.} For robust fine-tuning evaluation, we commonly consider a covariate shift scenario for the classification task, where both ID and OOD tasks share the same label space $\mathcal{Y}$, and have the same conditional distribution over target labels, but have different marginal distributions over input \(\mathcal{X}\). That is,
\(P_{S}(y|x) = P_{T}(y|x)\), but \(P_{S}(x) \neq P_{T}(x)\). Robust fine-tuning evaluates the CLIP model that is fine-tuned on a training split of the ID task, on a test split of ID, and on OOD tasks. 

\textbf{Regularization-based robust fine-tuning methods.} Robust fine-tuning aims to achieve consistently high performance on data from both fine-tuning downstream task (ID performance, source task) and related but different test distributions (OOD robustness, target task). The pre-trained CLIP includes a pair of encoders: a visual or image encoder \(f_{\theta_{v_0}}(\cdot)\) and a text encoder \(f_{\theta_{t_0}}(\cdot)\), trained on large-scale image--text pairs to align both modalities into a shared space $\mathbb{R}^D$.
For image classification, given an input image ${x}$ and a set of $K$ class names $\{c_i\}_{i=1}^K$ as natural language text, the encodings for each class $\psi_{\theta_{t_0}}({t}_i) = f_{\theta_{t_0}}({t}_i)$ ($t_i$ is the text related to class $c_i$) and the image $\phi_{\theta_{v_0}}({x}) = f_{\theta_{v_0}}({x})$ are first obtained. The text encodings $\psi_{\theta_{t_0}}({t}_i)$ are then used as parameters of a $K$-class linear classifier, and the classification inference on ${x}$ is performed as $\arg\max_i \psi_{\theta_{t_0}}({t}_i)^{\top} \phi_{\theta_{v_0}}({x})$. This is known as the zero-shot (\textbf{ZS}) prediction and CLIP's pre-trained model has been shown to have competitive ZS performance \cite{radford2021learning}. The vanilla fine-tuning (\textbf{FT}) method aims to fine-tune the entire model $\theta = \{\theta_v, \mathbf{w}\}$ (including the image encoder parameters $\theta_v$ and linear head $\mathbf{w}: \mathbb{R}^D \to \mathbb{R}^K$)  using a cross-entropy loss $\mathcal{L}_{\text{CE}} $ in an end-to-end manner. Given data $\mathcal{D} = \{(x_i,y_i)_{i=1}^N, y = 1,2, \cdots, K\}$ of downstream task, the loss could be written as $ \mathcal{L}_{\text{CE}} (\theta_{v},\mathbf{w}; \mathcal{D})= \frac{1}{N}\sum_{i=1}^N \mathcal{L}_{\text{CE}} (\mathbf{w}^\top f_{\theta_{v}}({x_i}),y_i)$.

Such FT method could yield substantial improvements on the ID data, while it may often undermine the model's performance on OOD data. To address this issue, regularization-based robust fine-tuning methods \cite{wortsman2022robust} have been recently developed to adapt models to ID while maintaining strong OOD robustness. Specifically, they additionally introduce a regularization term \(R(\theta_v,\mathbf{w})\) for preserving the knowledge of the pre-trained models by modifying the fine-tuning procedure as follows \cite{wortsman2022robust,goyal2023finetune,lee2022surgical}
\begin{equation}
	\theta_{v}^*,\mathbf{w}^* = \arg\min_{\theta_{v},\mathbf{w}}  \mathcal{L}_{\text{CE}} (\theta_{v},\mathbf{w}; \mathcal{D}) + \lambda R(\theta_v).
\end{equation}
The detailed forms of \(R(\theta_v)\) are summarized in Table~\ref{tab:reg_methods}, including strategies to minimize the distance of model weights, features, or logits between fine-tuned and pre-trained models. For the detailed related work, please refer to Appendix.

\begin{table}[tb]
	\centering
	\small
	\setlength{\tabcolsep}{4pt}
	\renewcommand{\arraystretch}{1.5} 
	\setlength{\extrarowheight}{3pt} 
	
	\caption{Summary of regularization-based robust fine-tuning methods.}
	\resizebox{0.98\textwidth}{!}{
		\label{tab:reg_methods}
		\begin{tabular}{lccc}

			\hline
			Method     & Venue        & $R(\theta_v)$  & Type \\
			\hline
			L2-SP \cite{mukhotifine}      & TMLR ’24     & $\|\theta_v - \theta_{v_0}\|_2^2 $ & parameter alignment\\
			CAR-FT \cite{mao2024context}    & IJCV ’24     & 
			$ \mathrm{KL}\left( \mathbf{w}_{ctx}^\top \phi_{\theta_{v_0}}(x)\,\|\,\mathbf{w}_{ctx}^\top \phi_{\theta_{v}}(x)\right)
			$  & logits alignment \\
			LDIFS  \cite{mukhotifine}     & TMLR ’24     & $\|\phi_{\theta_{v}}(x) - \phi_{\theta_{v_0}}(x)\|_2^2$ & feature alignment \\
			Lipsum-FT \cite{nam2024lipsum} & ICLR ’24     & 
			$\displaystyle
			\frac{1}{2M} \sum_{i=1}^M\bigl\|\langle {\psi}_{\theta_{t_0}}(t^{i}), \phi_{\theta_{v}}(x) \rangle - \langle {\psi}_{\theta_{t_0}}(t^i), \phi_{\theta_{v_0}}(x)\rangle\bigr\|_2^2 \;
			$  & logits alignment \\
			CaRot \cite{oh2024towards}     & NeurIPS ’24  &
			\parbox[t]{0.55\linewidth}{\centering\scriptsize$\displaystyle
				\mathrm{KL}\left( \phi_{\theta_{v_{ema}}}(x)\,\|\,\phi_{\theta_{v}}(x)\right)
		+\mathrm{KL}\left( \psi_{\theta_{t_{ema}}}(x)\,\|\,\psi_{\theta_{t}}(x)\right)
				$} & feature alignment\\
					\hline
			FRR-FT & This manuscript &   $\frac{1}{N} \sum_{i=1}^N \left[|f_{\theta}(\tilde{x}_i) - f_0(\tilde{x}_i)|^2\right]+\frac{1}{N} \sum_{i=1}^N[KL(f_{\theta}(x_i) ||f_{\theta}(\tilde{x}_i))]$                       &  functional alignment \\
			\hline
		\end{tabular}
	}
\end{table}

\section{Proposed Method}

Our goal is to achieve good OOD robustness during the adaptation to downstream ID tasks of pretrained models. As shown in Fig. 1, existing regularization-based robust fine-tuning methods cannot always achieve OOD robustness of pretrained models for different model architectures. Different from existing methods  
aligning model weights, features, or logits between fine-tuned and pre-trained models, we explore aligning the prediction function between fine-tuned and pre-trained models for maintaining OOD robustness. The key insight stems from that the OOD robustness of a model should consider how the model function's output changes with respect to varying input information, which is related to the space of concerned functions themselves.
We firstly present the preliminary about learning in function space in Section \ref{lable1}. Then we show that the proposed functional regularization robsut fine-tuning (FRR-FT) method in Section \ref{lable2}.
Finally, we present some empirical analysis of the proposed FRR-FT method in Section \ref{lable3}.

\subsection{Learning in Function Space} \label{lable1}
Here, we study the fine-tuning process of CLIP models in the space of functions defined by the inner product \(\langle f, g \rangle = \int_{\mathcal{X}} f(x)g(x)d\mu(x)\), which yields the following norm:
\[
\|f\|^2 = \int_{\mathcal{X}} |f|^2 d\mu.
\]
Here \(\mu\) is a measure and corresponds to the probability density of the input \(\mathcal{X}\). It is important to note that this norm is computed with respect to an empirical data distribution, rather than a uniform distribution over the entire input space. The \(|\cdot|^2\)
operator denotes the 2-norm, which is applicable to vector-valued functions. Although we refer to this function space as a Hilbert space, we do not explicitly rely on an inner product structure; thus, the same discussion applies to any normed vector space, such as a Banach space. The distance between two functions \(f\) and \(g\) is given by 
\[
\|f - g\|^2 = \int_{\mathcal{X}} |f - g|^2 d\mu.
\]
Since \(\mu\) is a density, \(\int_{\mathcal{X}} d\mu = 1\), and we can write
\[
\|f - g\|^2 = \mathbb{E}_{x}[|f(x) - g(x)|^2].
\]
The expectation can be approximated as an empirical expectation over a batch of examples drawn from the input distribution:
\[
\|f - g\|^2 \approx \frac{1}{N} \sum_{i=1}^N |f(x_i) - g(x_i)|^2.
\]
In the following, we will use above approximation to compute functional regularization.

\subsection{Proposed functional regularization robust fine-tuning} \label{lable2}
To improve OOD robustness of fine-tuned models, we propose two functional regularizations presented in Section \ref{lable21} and \ref{lable22}, respectively, and overall algorithms presented in Section \ref{lable23}.

\subsubsection{Functional alignment regularization between fine-tuned and pre-trained models} \label{lable21}

We aim to constrain the distance between fine-tuned model \( f_{\theta} \) and pretrained model \( f_0 \) in the function space, ensuring that the fine-tuned model remains close to the pretrained one on OOD task, i.e., 
\begin{align*}
	\mathcal{R}_{FAR} = \|f_{\theta} - f_0\|^2 = \mathbb{E}_{x\sim \mathcal{D}_{ood}}[|f_{\theta}(x) - f_0(x)|^2].
\end{align*}
Generally, we only have access to ID data $\mathcal{D} = \{(x_i,y_i)_{i=1}^N\}$ of downstream task, while OOD data are unavailable. To address the issue, we use ID data $\mathcal{D}$ to help generate simulated OOD data through data augmentation as studied in domain generalization \cite{zhou2020learning,zhoudomain}. We use RandAugment technique \cite{cubuk2020randaugment} to generate potential simulated OOD data $\mathcal{D}_{ood} = \{(\tilde{x}_i)_{i=1}^N\}$. Now, we can approximate $\mathcal{R}_{FAR}$ by
\begin{align}
	\tilde{\mathcal{R}}_{FAR} = \frac{1}{N} \sum_{i=1}^N \left[|f_{\theta}(\tilde{x}_i) - f_0(\tilde{x}_i)|^2\right].
\end{align}
Above functional alignment regularization (FAR) aims to approximate the $L2$ function distance of fine-tuning and pre-trained models to better maintain OOD robustness capability of pre-trained CLIP models, and we empirically verify FAR could bring better OOD robustness
 (refer to Section \ref{lable3}).

\begin{algorithm}[t]
	\caption{Learning algorithms of functional regularization robust fine-tuning  (FRR-FT)}
	\label{alg:hilbert_gd}
	\begin{algorithmic}[1]
		\STATE \textbf{Input:} Initial parameters \(\theta_0\), learning rates \(\eta\), ID data $\mathcal{D}$, pretrained model \(f_0\)
		\WHILE{not converged}
		\STATE Sample a training mini-batch \(X \sim \mathcal{D}\)
		\STATE Generate the augmented OOD data $\tilde{X}$
		\STATE Compute gradient of the main loss in Eq.(\ref{eq1}): $\nabla_{\theta} \mathcal{L}$
		\STATE Update parameters: \(\theta \gets \theta - \eta \nabla_{\theta} \mathcal{L}\)
		\ENDWHILE
		\STATE \textbf{Output:} \(\theta\)
	\end{algorithmic}
\end{algorithm}

\subsubsection{Functional consisteny regularization during fine-tuning }\label{lable22}
Besides, current insight on OOD robustness mainly attributes to the general knowledge of pre-trained model, while ignoring the potential contribution of fine-tuning data in downstream tasks. To this goal, we propose the following functional consistency regularization (FCR) to
minimize the discrepancy between the model's predictions with perturbed samples, expressed as	
\begin{align*}
	\mathcal{R}_{FCR} = \mathbb{E}_{x\sim \mathcal{D},\tilde{x} \sim \mathcal{D}_{ood}}[KL(f_{\theta}(x) ||f_{\theta}(\tilde{x}))].
\end{align*}
FCR seeks to regularize the change in the model function's output distribution throughout varying the input distribution, in order to ensure
the fine-tuned model capable of behaving robustly for potential distribution shift during the fine-tuning process. Similar to FAR, we also use RandAugment technique to generate potential simulated OOD data $\mathcal{D}_{ood} = \{(\tilde{x}_i)_{i=1}^N\}$, then we can approximate $\mathcal{R}_{FAR}$ by
\begin{align}
	\tilde{\mathcal{R}}_{FCR} = \frac{1}{N} \sum_{i=1}^N[KL(f_{\theta}(x_i) ||f_{\theta}(\tilde{x}_i))].
\end{align}
FCR explicitly enforces smoothness of the fine-tuned model's prediction landscape with respect to input perturbations, which explores to directly enhance the model’s robustness and generalization capability beyond that mere alignment with the pre-trained CLIP function.

\subsubsection{Learning algorithm of FRR-FT}\label{lable23}
Based on proposed functional regularization, we use the following objective for robust fine-tuning 
\begin{align}\label{eq1}
	\mathcal{L} = \mathcal{L}_{CE} + \lambda_1 \tilde{\mathcal{R}}_{FAR} + \lambda_2  \tilde{\mathcal{R}}_{FCR},
\end{align}
where $\mathcal{L}_{CE}$ aims to obtain ID fine-tuning performance, $\tilde{\mathcal{R}}_{FAR}, \tilde{\mathcal{R}}_{FCR}$ expects to improve OOD robustness, and $\lambda_1,\lambda_2$ are hyperparameters making trade-off between OOD Robustness and ID Fine-Tuning Performance.
The overall algorithm is presented in Algorithm \ref{alg:hilbert_gd}.

\begin{figure}[ht]
	\centering
	\begin{subfigure}{0.48\linewidth}
		\centering
		\includegraphics[width=\linewidth]{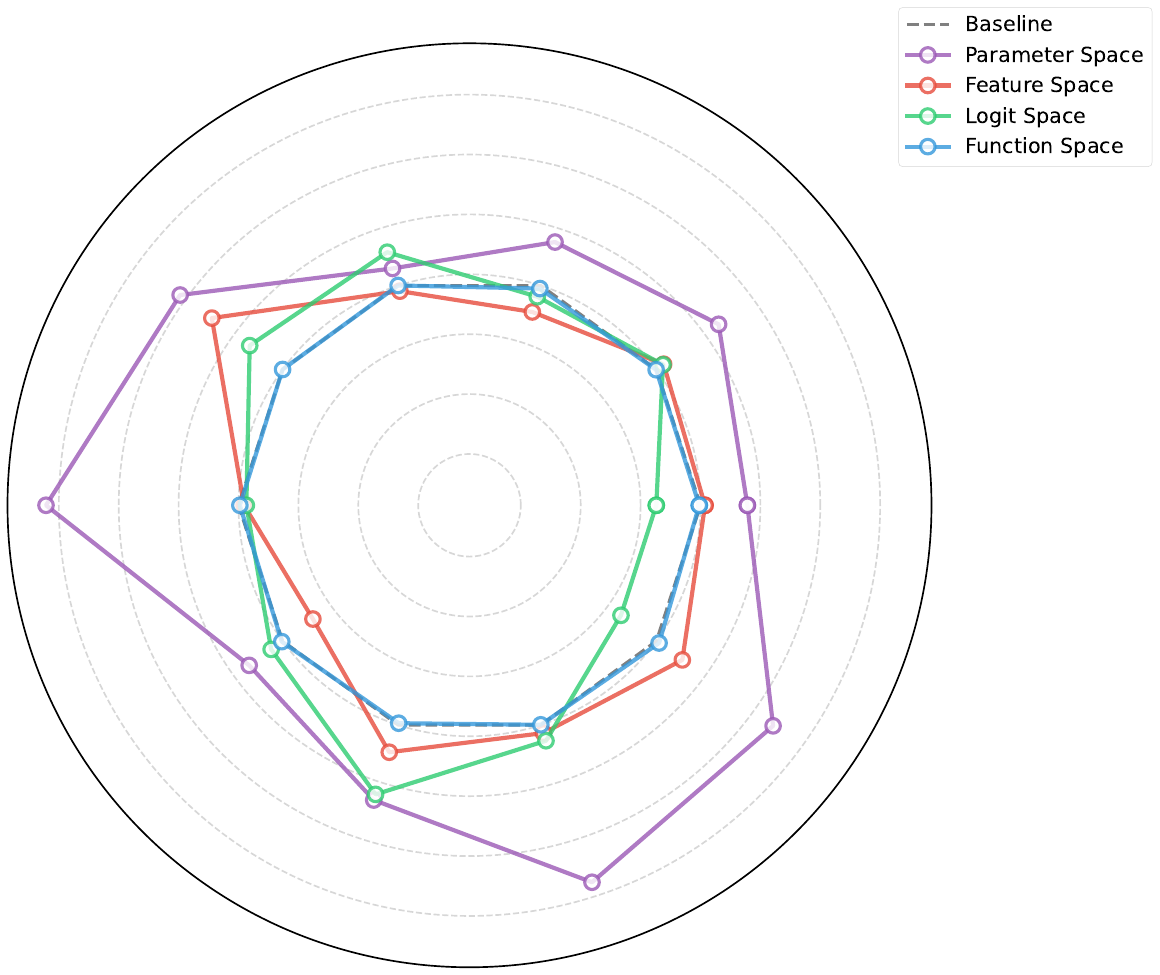}
		\subcaption{Loss of fine-tuned models}
		\label{fig:subfig1}
	\end{subfigure}%
	\hfill
	\begin{subfigure}{0.48\linewidth}
		\centering
		\includegraphics[width=\linewidth]{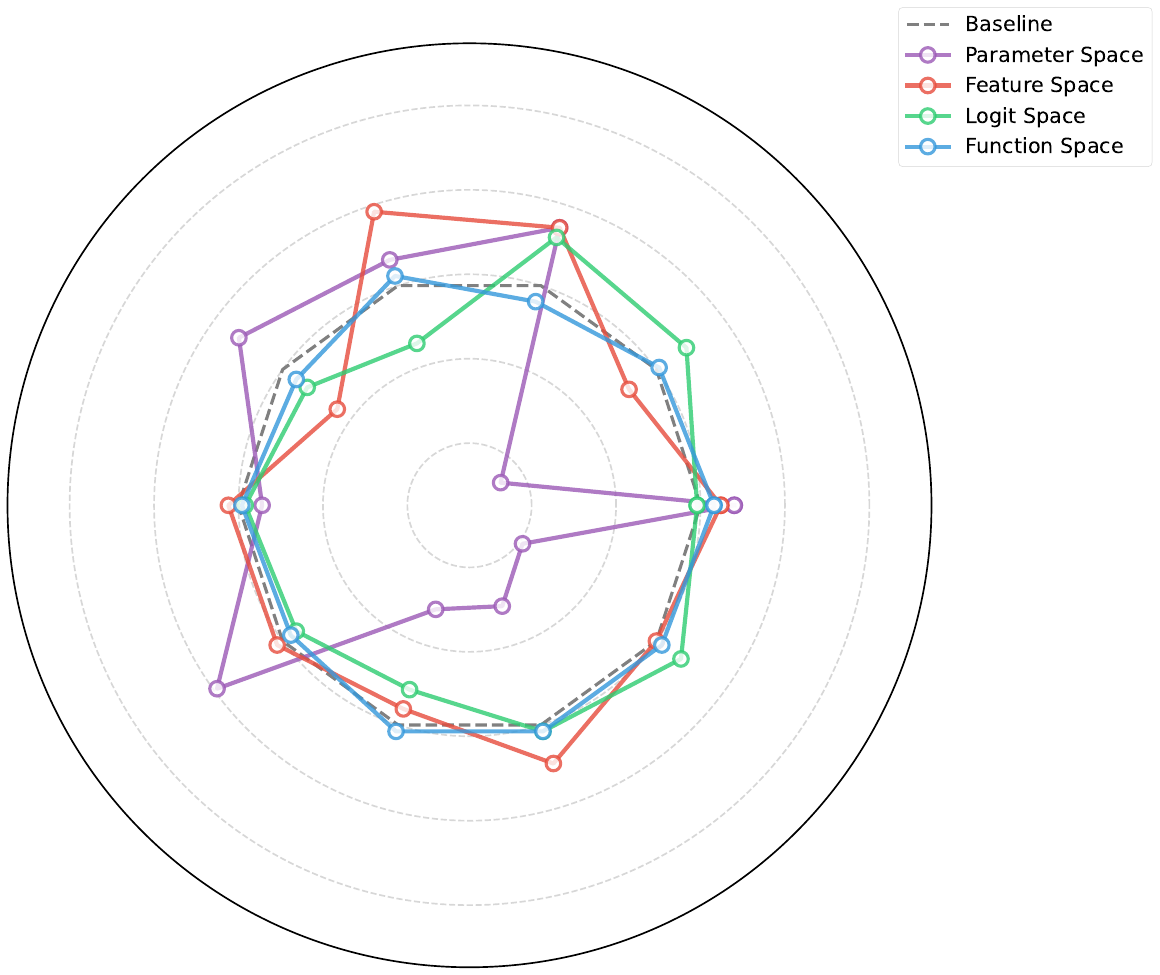}
		\subcaption{Test Accuracy of fine-tuned models}
		\label{fig:subfig2}
	\end{subfigure}
	\caption{Robustness analysis of optimization in parameter/feature/logit spaces and proposed function space. Figure shows the changes in test loss and top-1 accuracy when the pre-trained model is fine-tuned along ten randomly generated unitary perturbation orthogonal directions  directions. Each perturbation direction is scaled to 2\% of the model’s parameter norm, highlighting how small, structured changes in different spaces affect fine-tuned model's OOD performance.}
	\label{figx}
\end{figure}

\textbf{Finding 1:} \textit{The function distance of fine-tuning and pre-trained models could reasonably reflect the OOD robustness of the model, while distance in the model parameter/feature/logits space only serves as an imperfect proxy for that in the function space. } We use the pre-trained CLIP models as an initialized model, and update the model along 10 randomly generated unitary perturbation directions calculated in the parameter/feature/logits space and proposed function space, respectively. We keep the updated magnitude equal. We then measured the performance of obtained models on multiple datasets. As seen in the Fig. \ref{figx}, revising the pre-trained model in the function space could consistently maintain the OOD robustness of pre-trained models. However, the models obtained via fine-tuning pre-trained model in parameter/feature/logits spaces can not always ensure the satisfactory OOD performance.

\begin{figure}[t]
	\centering
	\begin{subfigure}[b]{\textwidth}
		\includegraphics[width=\linewidth]{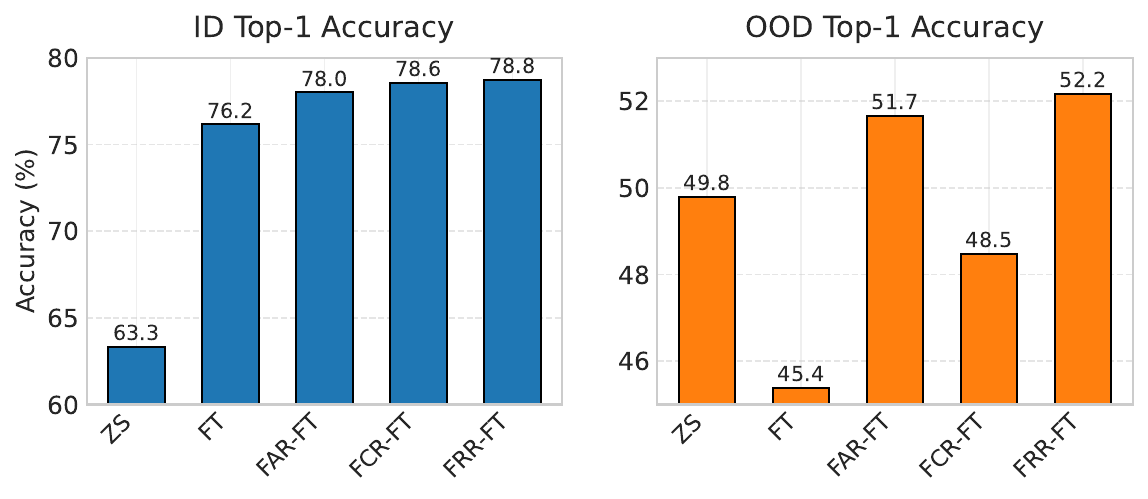}
	\end{subfigure}
	\caption{Relative performance gains over the pretrained model under full-parameter fine-tuning (FT) when incorporating the function-alignment regularizer (FT+FAR), the function-consistency regularizer (FT+FCR), and both regularizers together (FT+FAR+FCR). Both FAR and FCR individually improve upon the FT baseline, and combining FAR with FCR yields further performance gains.}
	\label{fig:paramters_function}
\end{figure}

\textbf{Finding 2:} \textit{Constraining the distance between fine-tuned and pre-trained models is relatively not enough for boosting OOD robustness.} Current insight on OOD robustness mainly attributes to the general knowledge of pre-trained model, while ignoring the potential contribution of data in downstream tasks. In the previous studies, the model's invariance takes an important role in improve model's OOD robustness \cite{liu2021towards,ahuja2021invariance}. To this goal, we conduct experiments on ImageNet benchmarks to show the impact on the OOD robustness of additional FCR objective. As shown in Fig. \ref{fig:paramters_function}, we can see that only FCR objective cannot achieve a better OOD performance than FAR objective, which may be due to the limited size of given data. While combining FAR and FCR could further boost the OOD robustness of FAR objective. This implies that traditional methods of improving model's OOD performance may also work in VLMs field. We hope this finding will inspire future research toward more attempts of previous researches on robust fine-tuning.

\textbf{Finding 3:}  \textit{The FRR-FT is model-agnostic, while existing regularization-based robust fine-tuning methods struggle with specific model architectures.} We conducted experiments across different CLIP model architectures, including VIT-L, VIT-B32, VIT-B16, and ResNet50, under the ImageNet experimental setup (detailed settings refer to Section \ref{labelex}). The results are presented in Table 2 and 3. Proposed FRR-FT method
consistently achieves improvements in both the downstream task ID performance (relative to fine-tuned models) and OOD performance (relative to the pre-trained model) on all CLIP model architectures. However, regularizing the distance of fine-tuning and pre-trained models in terms of parameter/feature/logits space could achieve OOD robustness and ID performance on specific model architectures. This demonstrates the advantage of 
FRR-FT for robust fine-tuning, highlighting the potential applicablility for real-life robust fine-tuning tasks.

\section{Experiments} \label{labelex}

\subsection{Distribution Shift Results}
To validate the effectiveness of our proposed fine-tuning method on multimodal models, we design experiments in 
class-invariant distribution shifts OOD scenarios.

All experiments are conducted on four models ViT-L/14, ViT-B/32, ViT-B/16, ResNet-50 based on CLIP Architecture. Under a unified training protocol we compare against SOTA baselines (1) zero-shot CLIP: following paper clip to zero-shot classification (2)FT: fine-tune the whole model parameters using supervised cross-entropy loss (3)Flyp\cite{goyal2023finetune}: fine-tune all the whole model parameters using image-text contrastive loss (4)L2SP\cite{xuhong2018explicit,mukhotifine}: Constrain the fine-tuning model parameters to be close to the pre-trained model parameters to limit the change of model parameters. (4)LDIFS\cite{mukhotifine}: Constrain the fine-tuning model to be close to the features of the ID dataset and the pre-trained model parameters to limit the feature changes of the model on the ID (5)CAR-ft\cite{mao2024context}: constraint the fine-tuned model is consistent with the pre-trained model on the domain classification proxy task (5)Lipsum-FT \cite{nam2024lipsum}: constraints the fine-tuned model’s ability to classify random text is consistent with the pre-trained model (6)CaRot\cite{oh2024towards}: increases the smallest singular value of the ID input covariance matrix to lower the upper bound of OOD calibration and generalization errors and get EMA model as the teacher model to provide supervised label information to design the cross entropy supervision loss.

Our model is fine-tuned on the ID dataset. Since the class of the distribution shift data belongs to the class of ID, we directly use the model to predict the OOD data label. Following the paper \cite{wortsman2022robust}, a simulation was constructed on ImageNet: each CLIP variant is fine-tuned end-to-end on ImageNet1K using AdamW with peak learning rate $3\times10^{-5}$ (500-step linear warm-up, cosine decay), weight decay $0.1$, and batch size 256, for 10 epochs.  After fine-tuning, we evaluate on four ImageNet-derived OOD benchmarks: ImageNet-V2 (a natural re-sampling of the original test set), ImageNet-R (200 classes of “rendition” images such as paintings and sculptures), ImageNet-Sketch (hand-drawn sketches of the 1 000 classes), and WILDS-iWildCam, a large camera-trap wildlife dataset. 
In addition, in order to further verify the robustness of the model, we verified it on the distribution shift of real scenes WILDS-iWildCam datasets.
In WILDS-iWildCam, each domain corresponds to a distinct trap location: the training split comprises approximately 201399 images from 323 locations, while the test split contains 60029 images from 91 held-out locations, spanning 182 animal species.  By design, no location in the test set appears during training, enabling a rigorous assessment of geographic domain shift.

As shown in Table~\ref{tab:ImageNet_comparison}, in our ImageNet fine-tuning experiments, we observe that (1) full-parameter tuning consistently boosts ID accuracy but incurs a substantial drop in OOD performance, regardless of backbone architecture. (2) Prior methods attempt to alleviate this trade-off by adding constraints in the parameter, feature or logit spaces. Although most of these approaches yield relative OOD improvements over vanilla fine-tuning, none can fully recover the pretrained models’ OOD accuracy across different architectures. Specifically, only LDIFS on ViT-B/16 and ViT-B/32 restores OOD performance to 60.49\% (versus 59.44\% for the pretrained model), and L2SP on ViT-B/16 (62.37\% v.s. 59.44\%) achieves pretrained-level OOD accuracy. We attribute this limitation to the weak coupling between those surrogate constraints and the model’s actual function mapping under distribution shift.
(3) By contrast, our method directly regularizes the mapping in function space, yielding simultaneous gains in both ID and OOD metrics. On ViT-B/32, we improve ID accuracy from 76.17\% to 78.98\%, surpassing the previous state-of-the-art (77.41\%), while  raising OOD accuracy from 48.69\% to 52.17\%, whereas competing fine-tuning schemes peak at only 50.09\%. On ViT-B/16, we achieve 83.28\% ID and 63.04\% OOD, exceeding the best reported pair (82.31\% ID / 62.37\% OOD) despite their method (L2SP) having sacrificed ID (75.52\%) to boost OOD. These results underscore the potency of constraining the learned function itself rather than its parameter or feature representations.
(4) Our approach also generalizes beyond transformers: when applied to a convolutional CLIP backbone, we raise ID accuracy from 76.23\% to 76.92\% and OOD accuracy from 43.56\% to 46.29\%. From the table, this is the only convolutional-CLIP fine-tuning method that improves OOD performance, further validating the broad applicability of our function-space regularization.

Due to the domain-specific nature of the WILDS datasets, the CLIP pretrained backbone exhibits poor baseline generalization (ID accuracy and OOD accuracy below 20\% )  as in the Table \ref{tab:WILDS-iWildCam_comparison} . Full-parameter fine-tuning immediately elevates these metrics often by more than 25 percentage points in ID and 20 points in OOD, demonstrating the necessity of adaptation on domain-unique data. Existing fine-tuning methods further improve upon this, typically yielding incremental gains. 
However, our function-space regularization delivers consistent and significant improvements across both transformer and convolutional CLIP backbones. Crucially, even when competing methods have already saturated their performance, our approach still pushes the frontier, underscoring its robustness under realistic distribution shifts.

To further validate generality, we extend our evaluation to (i) the FLOW domain-shift dataset and (ii) a long-tailed, extended-class classification task. In both scenarios, our method consistently outperforms state-of-the-art fine-tuning schemes (see Appendix for full experimental details). These results confirm that function-space optimization provides a powerful, broadly applicable framework for improving robustness on real-world, domain-specific datasets.

\begin{table}[t]
	\centering
	\caption{ViT-B/32, ViT-B/16, ResNet50 on ImageNet}
	\label{tab:ImageNet_comparison}
	\begin{tabular}{lcc|cc|cc}
		\toprule
		\multirow{2}{*}{Methods} 
		& \multicolumn{2}{c}{ViT-B/32} 
		& \multicolumn{2}{c}{ViT-B/16} 
		& \multicolumn{2}{c}{ResNet50} \\
		& ID    & OOD Avg.\ & ID    & OOD Avg.\ & ID    & OOD Avg.\ \\
		\midrule
		ZS        & 63.34 & 49.79     & 68.33 & 59.44     & 59.84 & 43.56     \\
		FT        & 76.17 & 45.39     & 81.27 & 54.10     & 76.23 & 41.59     \\
		FLYP  \cite{goyal2023finetune}    & 76.39    & 46.54       & 82.31    & 56.30       & 76.06    & 41.07   \\
		L2SP    \cite{xuhong2018explicit,mukhotifine}    & 77.62 & 47.46     & 75.52 & 62.37     & 64.42 & 42.42     \\
		LDIFS   \cite{mukhotifine}  & 77.33 & 50.09 & 81.43    & 60.49        & 76.38 & 41.69     \\
		CAR-FT \cite{mao2024context}   & 77.60 & 48.31     & 82.66 & 59.26     & 76.60 & 43.42     \\
		Lipsum-FT \cite{nam2024lipsum} & 76.33 & 45.27     & 81.39 & 54.62     & 76.02 & 41.77     \\
		CaRot  \cite{oh2024towards}   & 77.41 & 47.59     & 82.76 & 58.94     & 76.20   & 41.80       \\
		\midrule
		Ours      & \textbf{78.98} & \textbf{52.17}     & \textbf{83.28} & \textbf{63.04}     & \textbf{76.92} & \textbf{46.29}     \\
		\bottomrule
	\end{tabular}
\end{table}

\begin{table}[t]
	\centering
	\caption{Performance on WILDS-iWildCam: ID and OOD metrics for ViT-B/32, ViT-B/16, and ResNet50}
	\label{tab:WILDS-iWildCam_comparison}
	\small
	\resizebox{\textwidth}{!}{
		\begin{tabular}{lccccccccc}
			\toprule
			Method  & \multicolumn{3}{c}{ViT-B/32} & \multicolumn{3}{c}{ViT-B/16} & \multicolumn{3}{c}{ResNet50} \\
			\cmidrule(lr){2-4} \cmidrule(lr){5-7} \cmidrule(lr){8-10}
			& Acc    & Recall   & F1       & Acc     & Recall   & F1       & Acc     & Recall   & F1       \\
			\midrule
			\multicolumn{10}{c}{\textbf{ID Metrics}} \\
			ZS       & 7.46  & 8.64  & 8.03  & 10.55 & 10.22 & 8.81  & 6.09  & 8.23  & 7.23  \\
			FT      &    64.21    &    30.21   &  29.55     &   69.59     &    31.53   &  32.77     &     63.49  &    22.74   &    23.10   \\
			FLYP \cite{goyal2023finetune} &   64.31    &   30.47    &  59.76     &   71.47    &   32.19    &  32.51     &   63.21   &   20.14    &    24.19   \\
			L2SP \cite{xuhong2018explicit,mukhotifine} & 73.05 & 35.29 & 35.78 & 75.79 & 38.16 & 38.52 & 67.74 & 26.25 & 25.35 \\
			LDIFS  \cite{mukhotifine}   & 77.76 & 49.68 & 44.30 & 80.69 & 48.15 & 48.31 & 78.75 & 41.34 & 42.56 \\
			CAR-FT \cite{mao2024context}  & 77.00 & 40.29 & 40.87 & 80.35 & 44.92 & 45.64 & 78.88 & 42.08 & 42.42 \\
			Lipsum-FT \cite{nam2024lipsum} & 77.34 & 40.77 & 41.29 & 79.98 & 45.86 & 45.62 & 78.53 & 40.95 & 41.08 \\
			CaRot \cite{oh2024towards} &    77.53   &   40.95    &   41.57    &   79.47    &    45.14   &    45.27   &    78.61   &  41.83     &   41.09    \\
			\midrule
			Ours     & \textbf{79.87} & \textbf{40.99} & \textbf{41.70} & \textbf{81.06} & \textbf{51.09} & \textbf{51.69} & \textbf{81.13} & \textbf{46.43} & \textbf{45.94} \\
			\midrule
			\multicolumn{10}{c}{\textbf{OOD Metrics}} \\
			ZS       & 12.89 & 7.80  & 7.32  & 15.23 & 13.24 & 10.99 & 10.30 & 8.39  & 6.22  \\
			FT       &  60.15    &   19.40    &  20.41    &    64.13   &    21.11   &   22.30    &   58.14    &  15.96    &   15.43    \\
			FLYP   \cite{goyal2023finetune}   &   60.21    &   20.12    &   20.08    &   65.90     &   22.15    &    22.08   &   27.61    &   17.49    &  18.05     \\
			L2SP \cite{xuhong2018explicit,mukhotifine}     & 68.94 & 27.87 & 28.74 & 68.03 & 27.33 & 26.07 & 63.52 & 17.70 & 17.37 \\
			LDIFS \cite{mukhotifine}   & 65.21 & 22.49 & 22.34 & 74.25 & 36.03 & 34.36 & 69.21 & 27.60 & 27.17 \\
			CAR-FT  \cite{mao2024context} & 65.55 & 27.88 & 26.81 & 74.85 & 34.47 & 33.67 & 69.31 & 29.42 & 29.03 \\
			Lipsum-FT \cite{nam2024lipsum}  & 63.77 & 24.48 & 24.75 & 75.16 & 38.01 & 36.91 & 69.17 & 28.62 & 25.53 \\
			CaRot \cite{oh2024towards}   &    65.74   &   28.91    &   28.88    &   75.33    &   38.13    &   37.24    &  69.14   &   29.51    &    26.33   \\
			\midrule
			Ours     & \textbf{72.49} & \textbf{31.58} & \textbf{32.41} & \textbf{78.29} & \textbf{38.56} & \textbf{39.31} & \textbf{69.44} & \textbf{31.52} & \textbf{29.07} \\
			\bottomrule
		\end{tabular}
	}
\end{table}

\subsection{Ablation Study}
Besides, an ablation study confirms the complementary roles of our two regularizers. Introducing the function-alignment term alone already recovers the pretrained OOD level while further boosting ID. For CLIP ViT-B/32 Architecture as seen in Fig. \ref{fig:paramters_function}, the consistency regularizer by itself raises ID from 76.2\% to 78.6\% and OOD from 45.4\% to 48.5\%, though it remains below the pretrained OOD accuracy of 49.8\%. When both regularizers are combined, we observe additional gains, ID climbs to 78.8\% from 76.2\% and OOD to 52.2\%, demonstrating that these two components work synergistically to preserve and enhance robustness under distribution shift.
To better understand the effectiveness of each component in our method, we conduct more ablation studies in the appdix, confirming that removing either component results in noticeable degradation in either OOD robustness or ID accuracy, demonstrating the complementary roles of the two regularization strategies.
We observe that incorporating the functional regularization alone improves the model’s ability to generalize under distribution shifts, indicating that constraining the model in function space is crucial for maintaining OOD behavior. Meanwhile, the consistency regularization enhances model stability under perturbations and leads to improvements across OOD datasets.

Notably, combining both regularizations leads to the best performance across all metrics. This suggests that (1) constraining the model in function space effectively preserves the robustness of the pre-trained model, while (2) promoting output consistency further reinforces generalization by leveraging structure from downstream data. Together, they address the two core challenges in robust fine-tuning: functional deviation from the pre-trained model and prediction instability on unseen domains.
Our ablation study confirms the necessity of both components in our framework. The functional distance regularization preserves the OOD-aware generality of the pre-trained model, while the consistency regularization encourages stable predictions across perturbed inputs. Their synergy enables us to achieve superior ID and OOD performance simultaneously.

\section{Conclusion}
In this work, we have delved into a new perspective from function-space regularization for robust fine-tuning of CLIP models. Current regularization-based robust fine-tuning methods aim to minimize the distance between fine-tuned and pre-trained models in the parameter/feature/logits space to preserve OOD robustness capabilities of pre-trained models. Though achieving some success in some cases, they cannot always achieve satisfactory OOD robustness for different model architectures. In our view, the optimization in parameter/feature/logits space only serves as an imperfect proxy for that in the function space. Comparatively, fine-tuning models in the function space is more effective for maintaining OOD robustness of pre-trained models. Based on this understanding, we propose a functional alignment regularization (FAR) for robust fine-tuning, which could consistently maintain the OOD robustness of pre-trained models for different CLIP model architectures. Besides, we find that constraining the distance between fine-tuned and pre-trained models is relatively not enough for boosting OOD robustness. As a step toward addressing this issue, we propose a functional consistency regularization (FCR) borrowed from previous researches on models' invariance in improving model's OOD robustness. Combining FAR and FCR achieves a synergistic effect.
Extensive experiments on multiple CLIP backbones and challenging distribution shift benchmarks demonstrate that our method consistently outperforms state-of-the-art regularization-based fine-tuning approaches. Ablation studies confirm the necessity and synergy of both regularizations in achieving superior robustness and generalization.
Overall, our framework offers a principled and practical solution for improving both OOD robustness and downstream task ID fine-tuning performance of CLIP-like models when adapting to downstream tasks. We believe FRR-FT can be broadly beneficial in improving the robust fine-tuing performance of learning tasks using pre-trained initialization.

\nocite{*}
\bibliography{reference}

\newpage
\appendix

\section{Related Work}
\subsection{Robust CLIP Fine-tuning}
Foundation models such as CLIP have demonstrated remarkable zero-shot and transfer learning capabilities across diverse vision-language tasks \cite{radford2021learning,bommasani2021opportunities,brown2020language}. However, fine-tuning these models on downstream in-distribution (ID) datasets often results in a notable degradation of out-of-distribution (OOD) robustness \cite{wortsman2022robust,kumarfine2022}. This inherent trade-off between improving ID accuracy and preserving OOD generalization has stimulated a substantial line of research dedicated to developing robust fine-tuning strategies that adapt foundation models to downstream tasks while retaining the broad generalization capabilities learned during pre-training.
Broadly speaking, these strategies fall into two major categories: regularization-based methods and model-ensemble based methods.
Regularization-based methods typically impose regularization constraints that encourage the fine-tuned model to remain close to the pre-trained model, either in the parameter space, in various representation spaces or context-aware. For instance, L2-SP \cite{xuhong2018explicit,mukhotifine} employs an $\ell_2$ penalty on the deviation between fine-tuned and pre-trained weights to mitigate catastrophic forgetting and achieve OOD robustness. Feature-level alignment methods such as LDIFS \cite{mukhotifine} promote consistency in intermediate feature representations of downtask datasets  between fine-tuned and pre-trained models, helping to stabilize learned semantic embeddings during fine-tuning. Context-aware regularization techniques like CAR-FT \cite{mao2024context} further propose proxy tasks to reduce distributional divergence in contextual embeddings between the fine-tuned and pre-trained models, while Lipsum-FT \cite{nam2024lipsum} maintains logits consistency through random textual prompt augmentation to preserve semantic alignment. More recently, CaRot \cite{oh2024towards} has combined self-distillation with covariance regularization to enhance the robustness and calibration of fine-tuned models.
Despite these advancements, the effectiveness of these methods often depends heavily on the underlying architecture and may struggle to simultaneously preserve both ID accuracy and OOD robustness across diverse model variants. Our proposed method achieves consistent improvements in downstream task performance and OOD robustness across a broad spectrum of CLIP architectures, thus pushing the frontier of robust fine-tuning.
Model-ensemble based methods, in contrast, leave the fine-tuning process unchanged and combine multiple trained checkpoints post hoc. Early work interpolated directly between pre-trained and fine-tuned weight vectors to balance ID accuracy and OOD robustness\cite{wortsman2022robust}, yielding concurrent gains on both metrics. Subsequent “weight averaging” approaches aggregate a cohort of fine-tuned models\cite{wortsman2022Modelsoups}, each obtained under different hyperparameter configurations, to outperform the single best checkpoint in large-scale sweeps on benchmarks such as ImageNet. Dawin\cite{oh2024dawin} refines this paradigm further by learning sample-specific ensemble coefficients, thereby optimizing per-example predictions.
Regularization-based and ensemble-based strategies have largely been studied in isolation and our work demonstrates that combining ensemble-based strategies with our proposed method achieves consistent improvements in both downstream ID accuracy and OOD robustness across a broad spectrum of CLIP architectures again, pushing the frontier of robust fine-tuning.(As in the Figure~\ref{fig:ensemble} )

\subsection{Functional Regularizations}
The regularization of neural networks has recently shifted from indirect parameter-space techniques (e.g., weight decay, dropout) toward methods that directly constrain the learned mapping in function space. This evolution addresses a key limitation of parameter-based penalties: they do not necessarily control how the model’s predictions behave across inputs. Function-space regularization, by contrast, more faithfully targets the ultimate goal of improving predictive performance, especially under distribution shift, in uncertainty calibration, and in continual or multi-task learning scenarios.
A principled way to perform function-space regularization is via priors defined directly over functions. For example, \cite{rudner2023function} introduced Function-Space Empirical Bayes (FS-EB), which interprets conventional parameter penalties as empirical priors and derives corresponding constraints on the predictive distribution via Bayesian inference. FS-EB employs a first-order Taylor expansion of the network to render variational inference tractable, yielding high-quality uncertainty estimates under covariate shift. Separately, \cite{titsias2020functional} applied a Gaussian process–based functional regularization in continual learning, demonstrating substantial mitigation of catastrophic forgetting by constraining changes to the model’s global function in response to new tasks.
Another line of work imposes explicit geometric constraints on the input–output mapping. \cite{benjamin2019measuring} showed that small parameter changes can induce large functional shifts, motivating Hilbert-Constrained Gradient Descent, which directly limits the stepwise change in expected outputs and thereby shortens optimization trajectories and reduces forgetting in sequential tasks. From a kernel perspective, \cite{bietti2019kernel} proposed regularizing the network’s RKHS norm induced by the empirical neural tangent kernel, effectively controlling its global functional complexity. Local smoothness has also been targeted via Jacobian-based penalties in other studies. More recently, \cite{chen2022ntk} leveraged the neural tangent kernel to define Mahalanobis-distance constraints in logit space, further improving robustness.
In this paper, we extend these insights to the fine-tuning of CLIP models. By enforcing function-space regularization during adaptation, we alleviate the common drop in out-of-distribution robustness that often accompanies standard fine-tuning, achieving superior OOD stability without sacrificing in-distribution accuracy.

\section{Empirical Analysis}

\begin{figure}[htbp]
	\centering
	\begin{subfigure}[b]{\textwidth}
		\centering
		\includegraphics[width=0.15\textwidth]{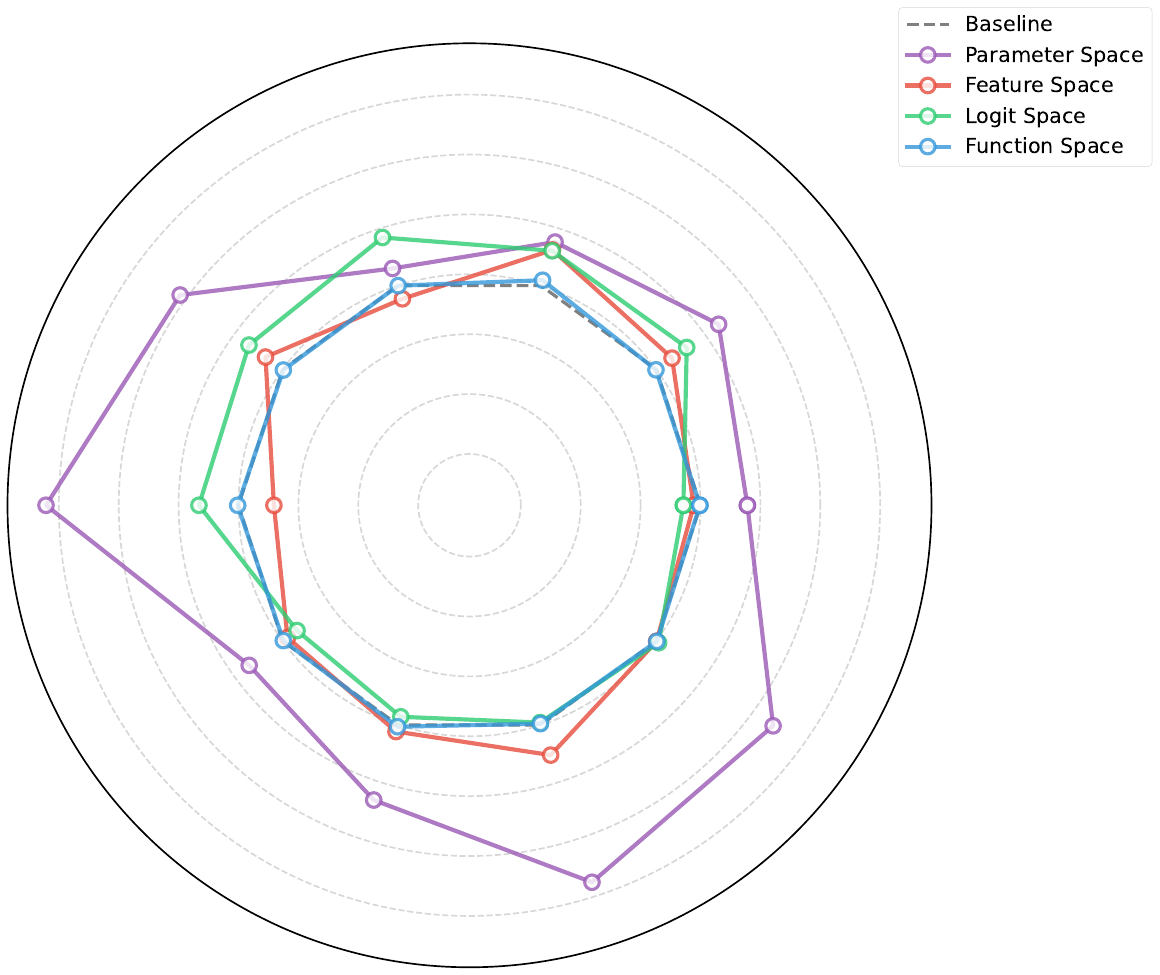}
		\includegraphics[width=0.15\textwidth]{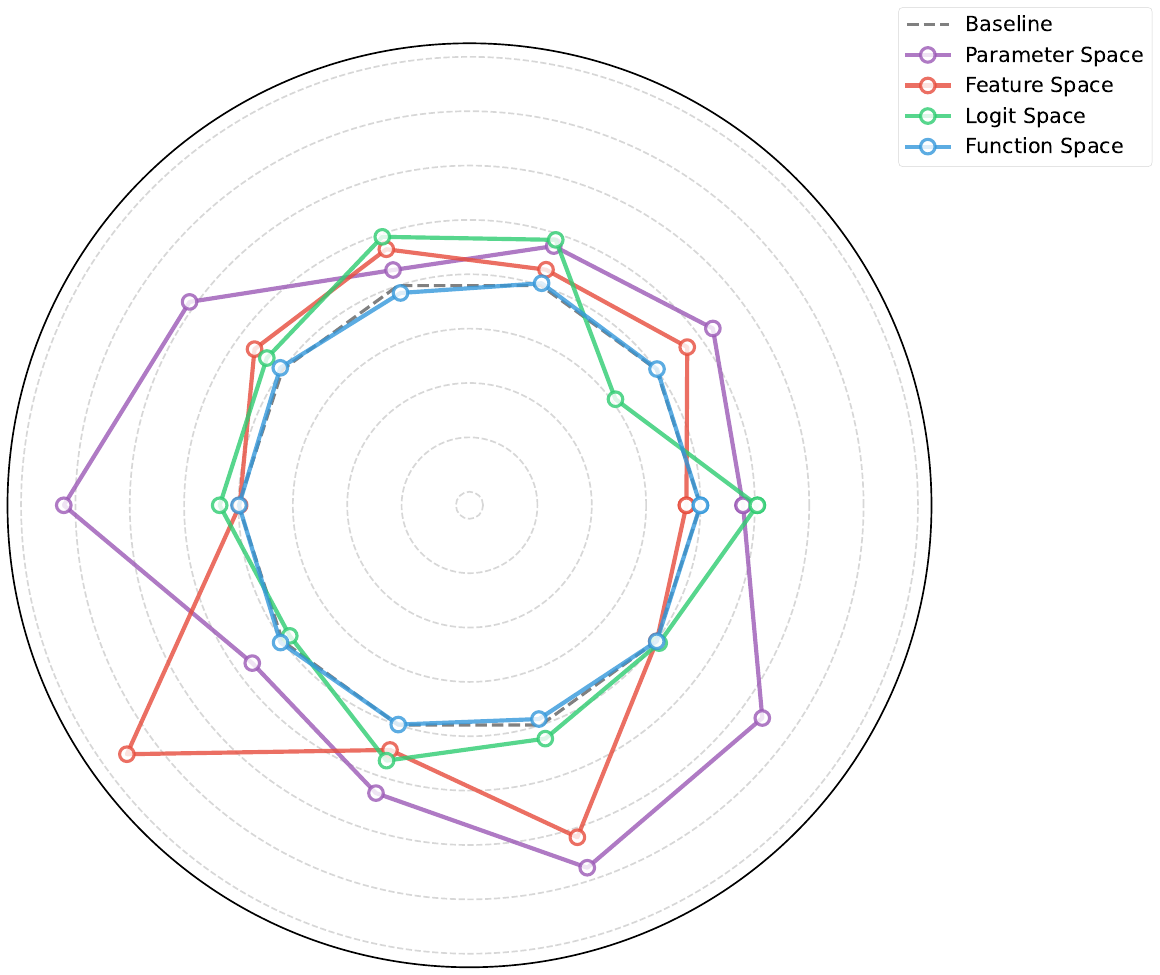}
		\includegraphics[width=0.15\textwidth]{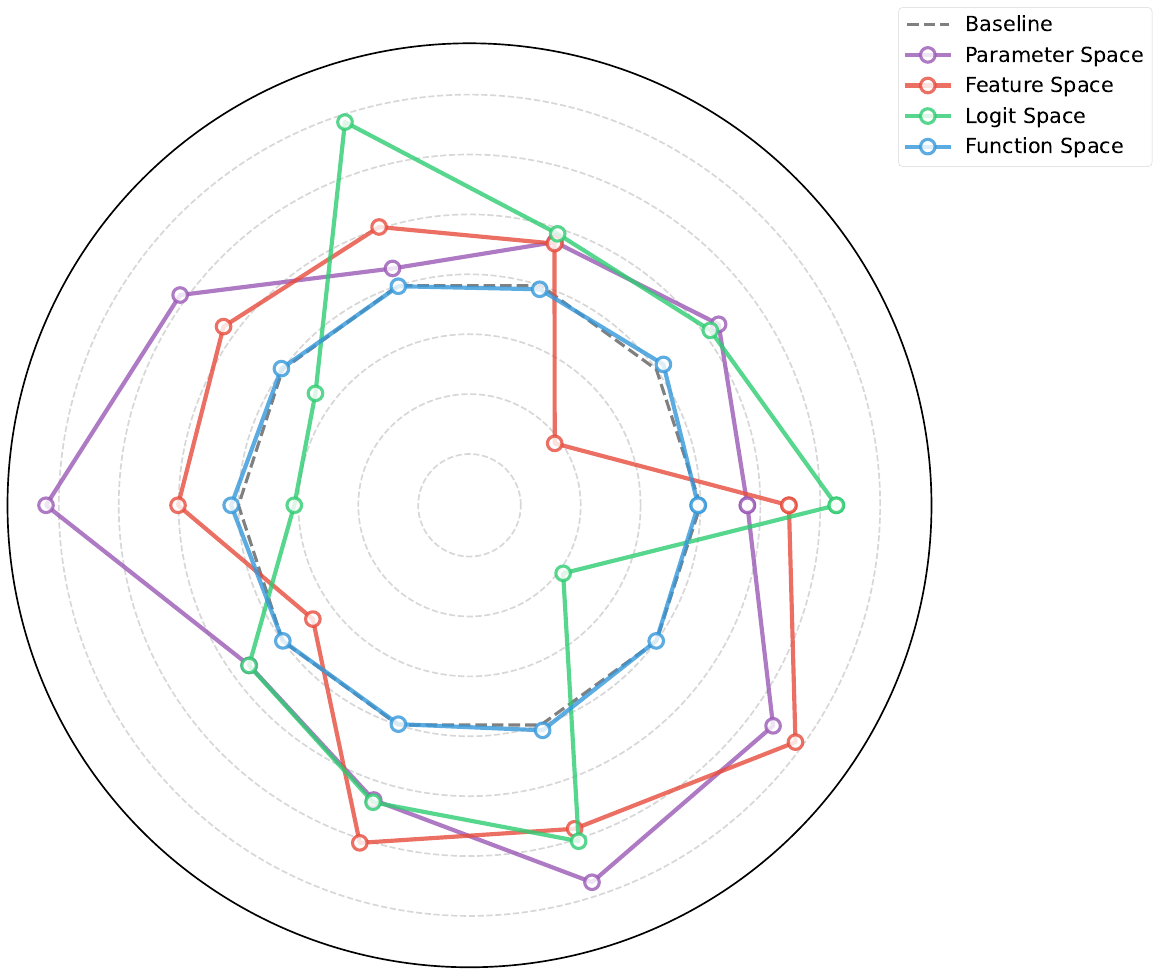}
		\includegraphics[width=0.15\textwidth]{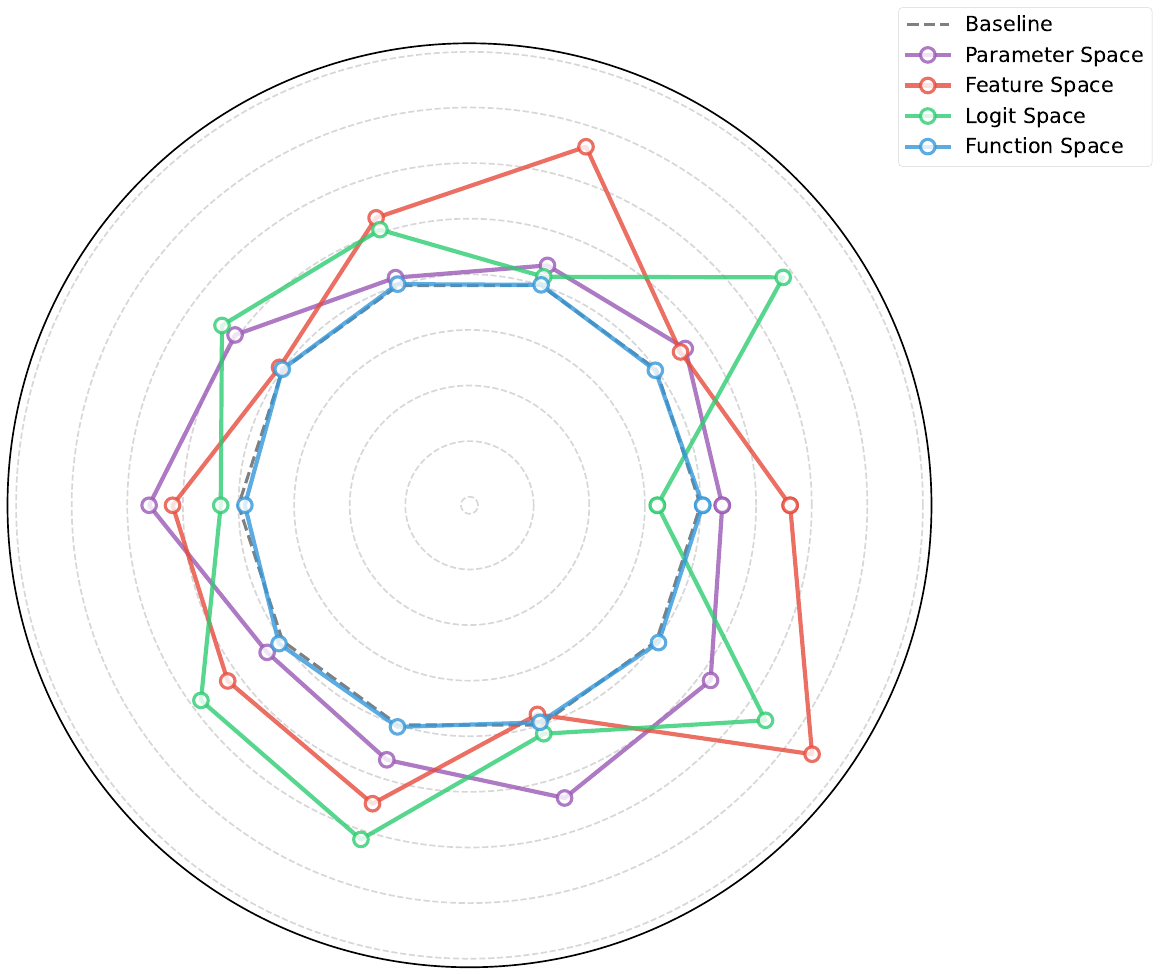}
		\includegraphics[width=0.15\textwidth]{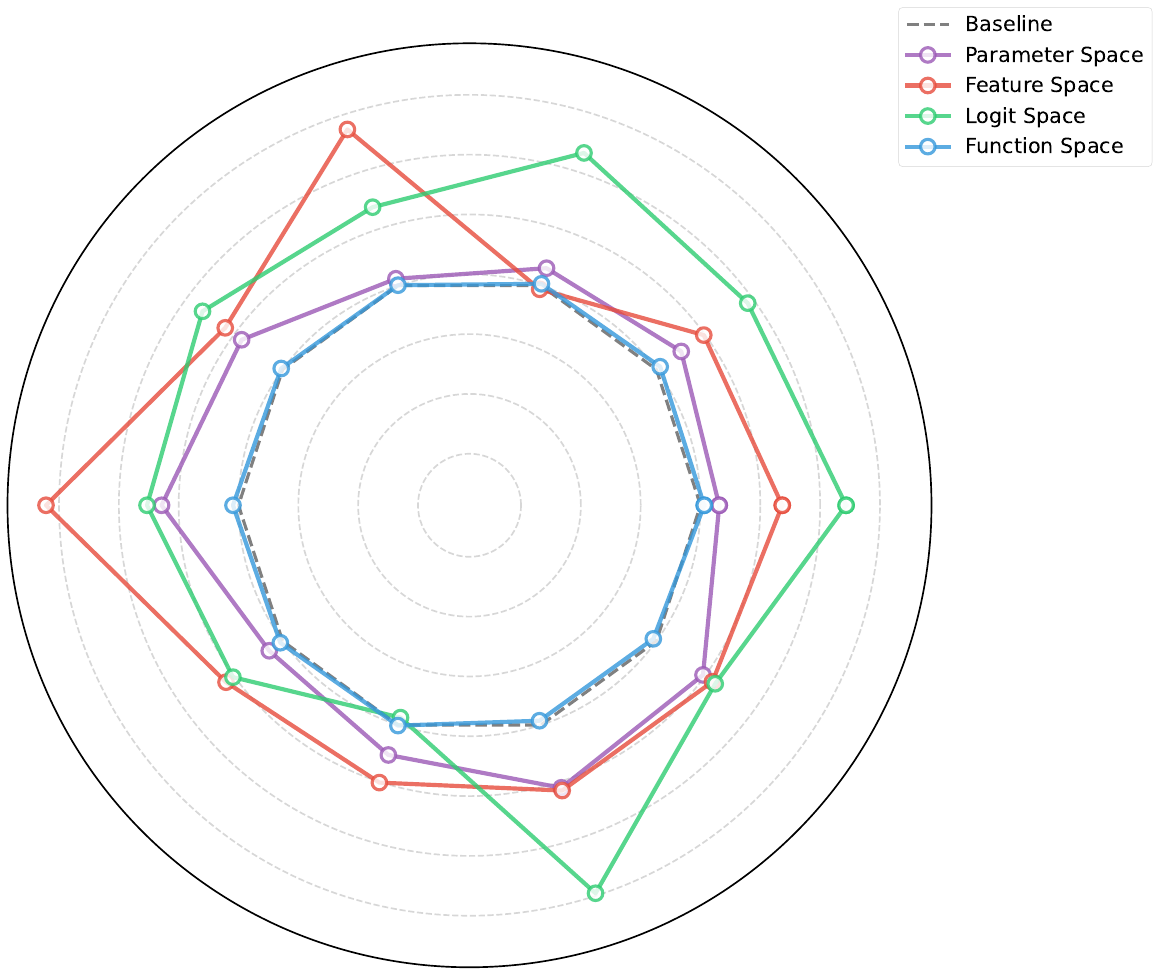}
		\includegraphics[width=0.15\textwidth]{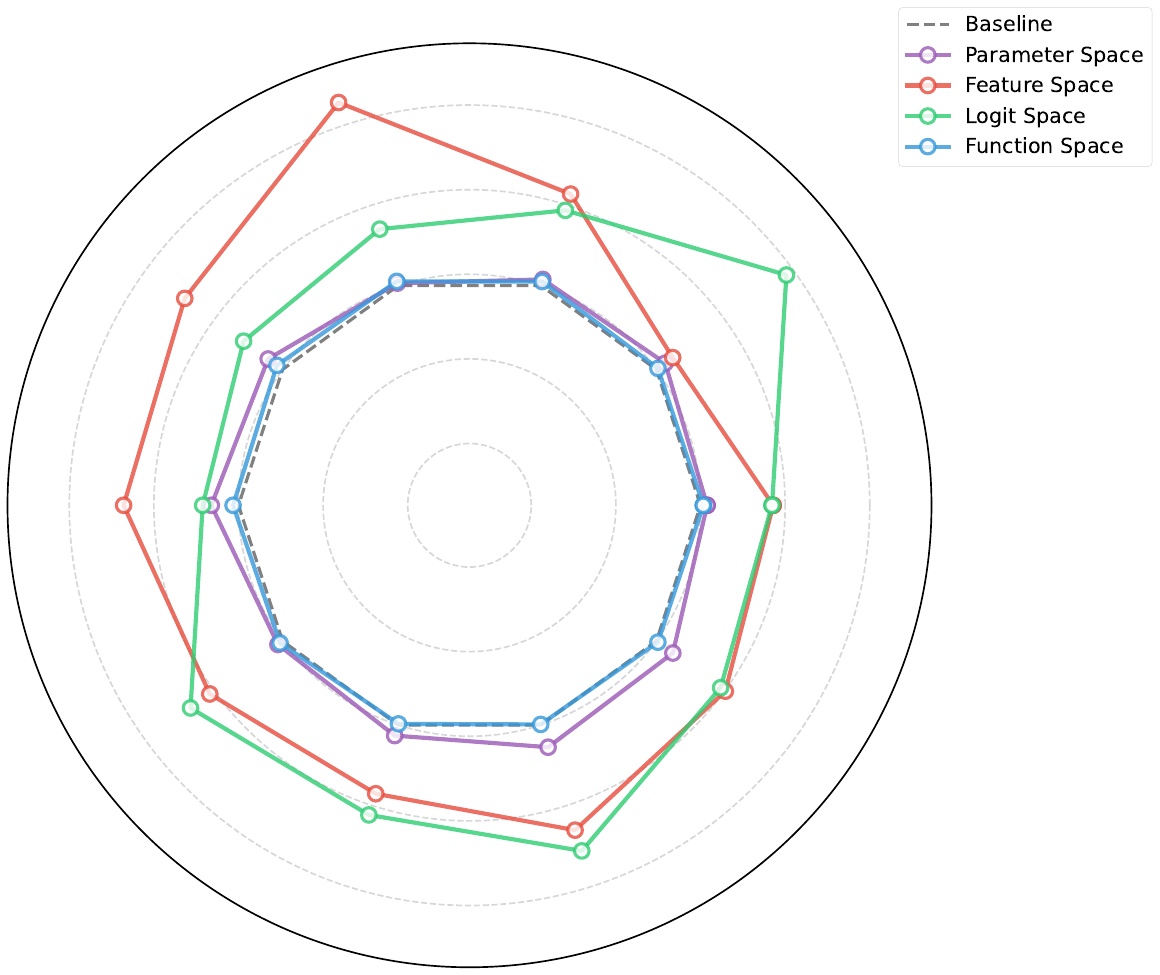}
		\subcaption{Effect of different perturbation magnitude on loss in ImageNet.}
		\label{fig:row1}
	\end{subfigure}
	\vspace{1ex}
	
	\begin{subfigure}[b]{\textwidth}
		\centering
		\includegraphics[width=0.15\textwidth]{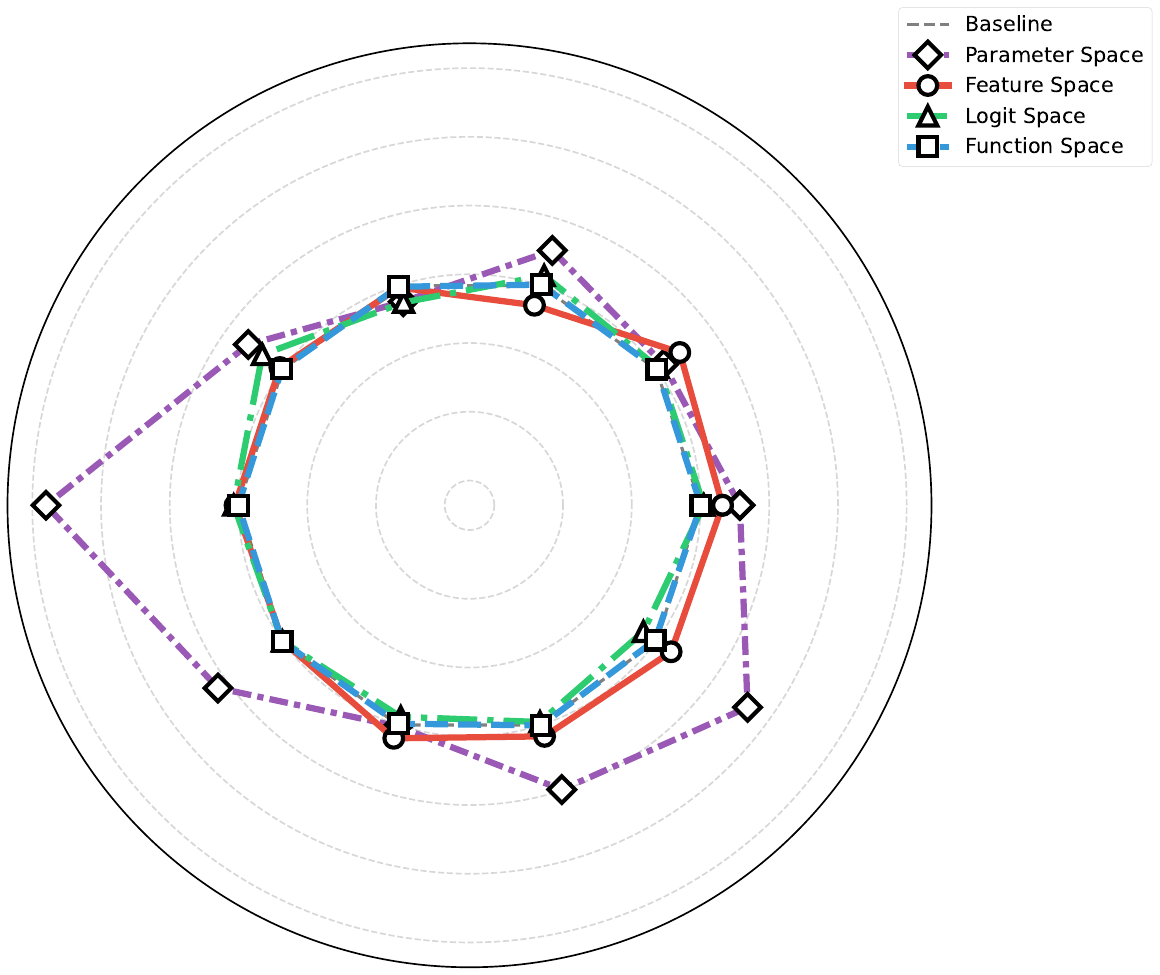}
		\includegraphics[width=0.15\textwidth]{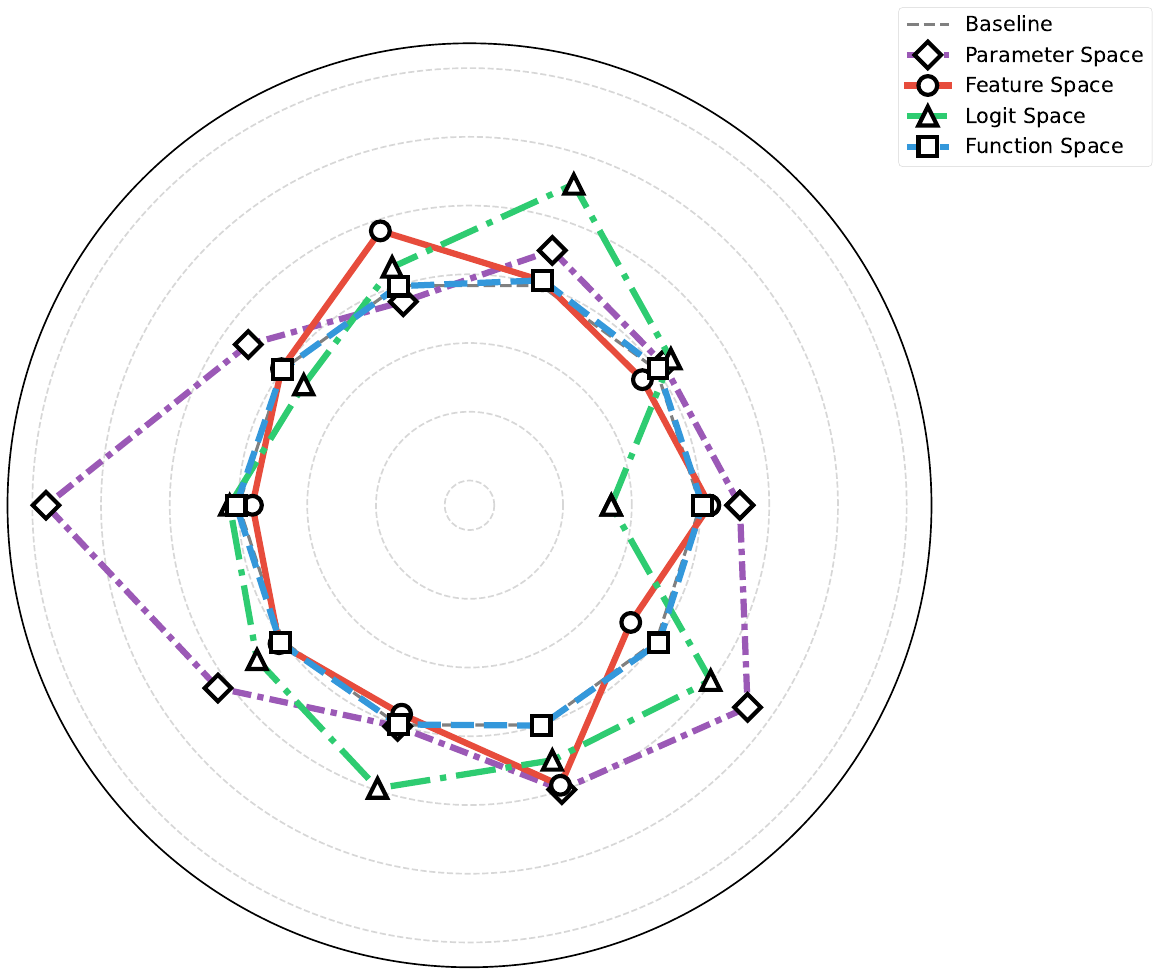}
		\includegraphics[width=0.15\textwidth]{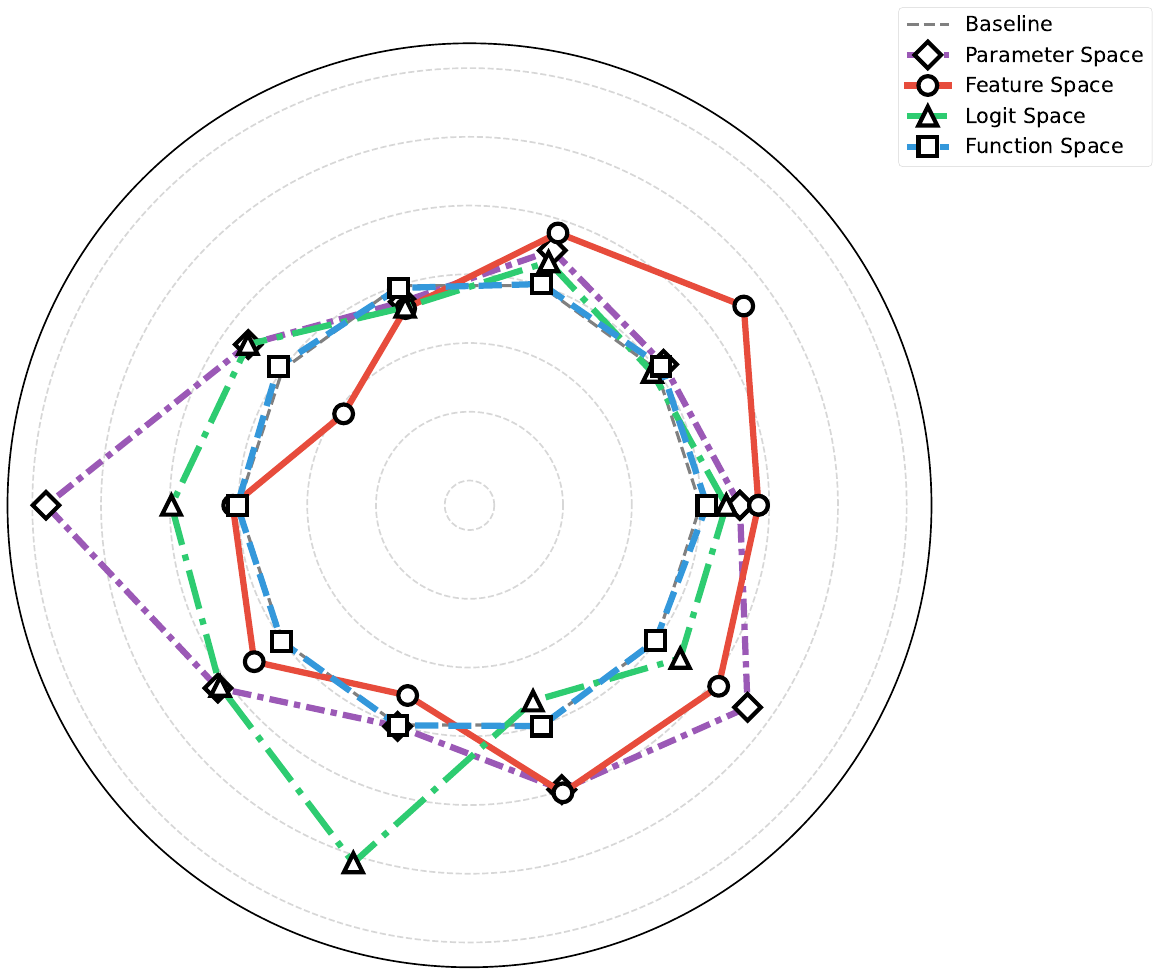}
		\includegraphics[width=0.15\textwidth]{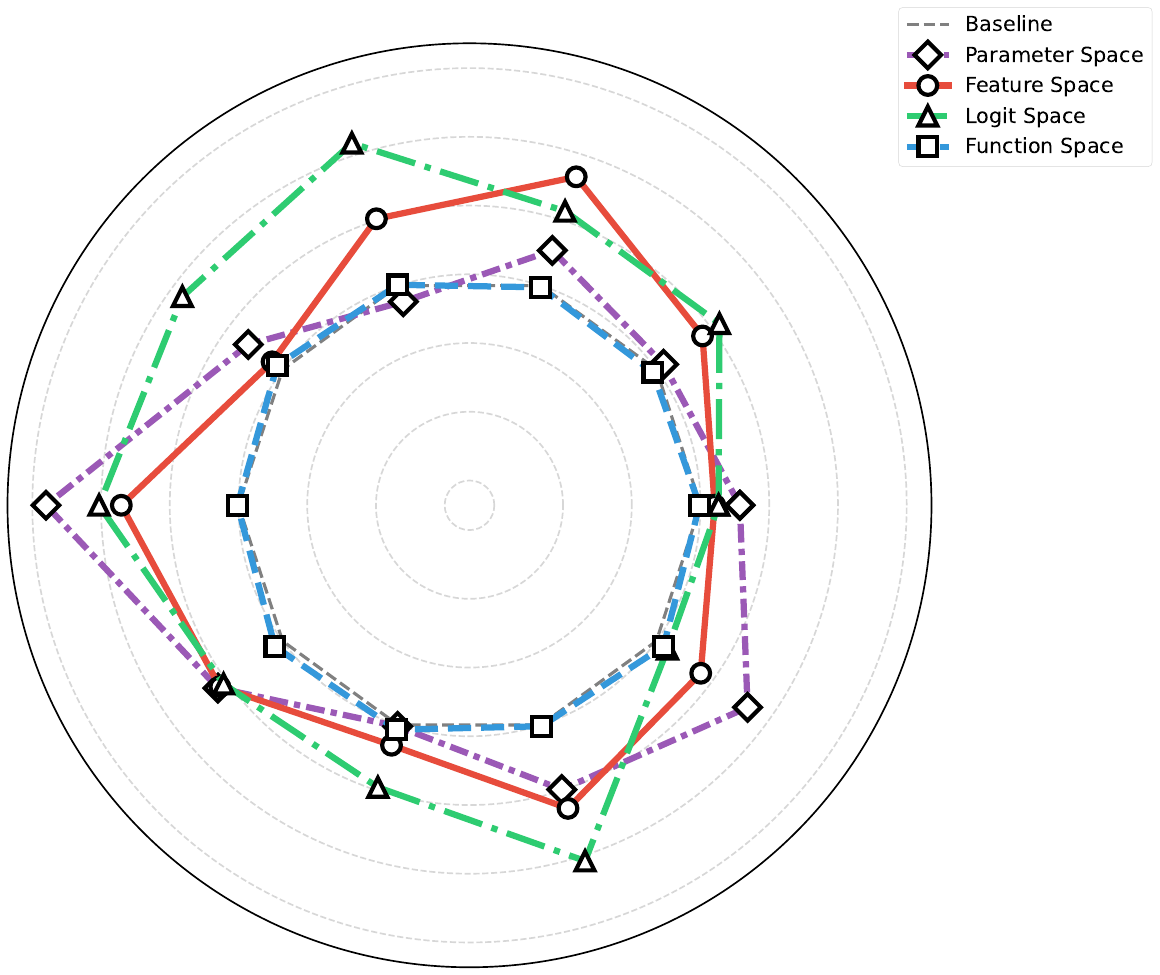}
		\includegraphics[width=0.15\textwidth]{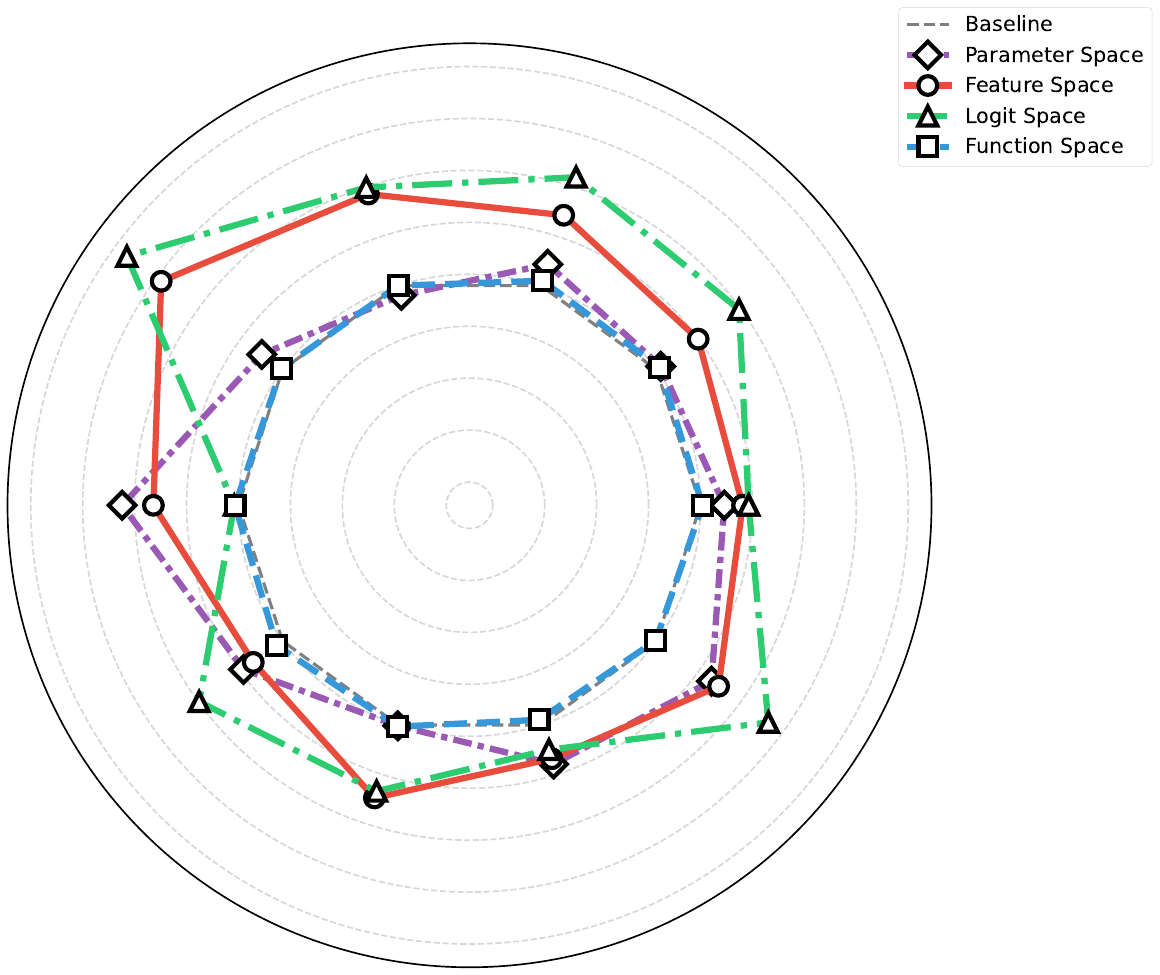}
		\includegraphics[width=0.15\textwidth]{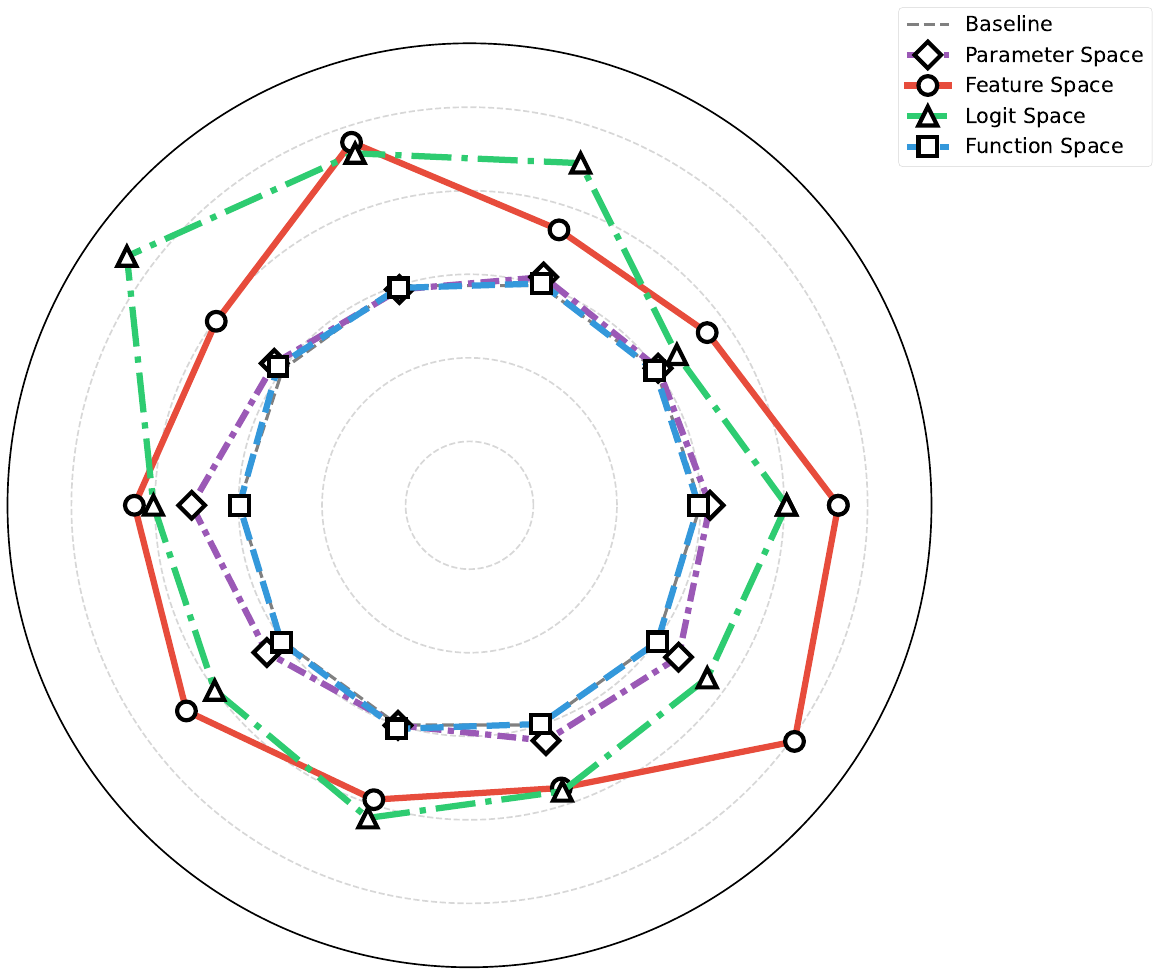}
		\subcaption{Effect of different perturbation magnitude on loss in ImageNetV2.}
		\label{fig:row1}
	\end{subfigure}
	\vspace{1ex}
	
	\begin{subfigure}[b]{\textwidth}
		\centering
		\includegraphics[width=0.15\textwidth]{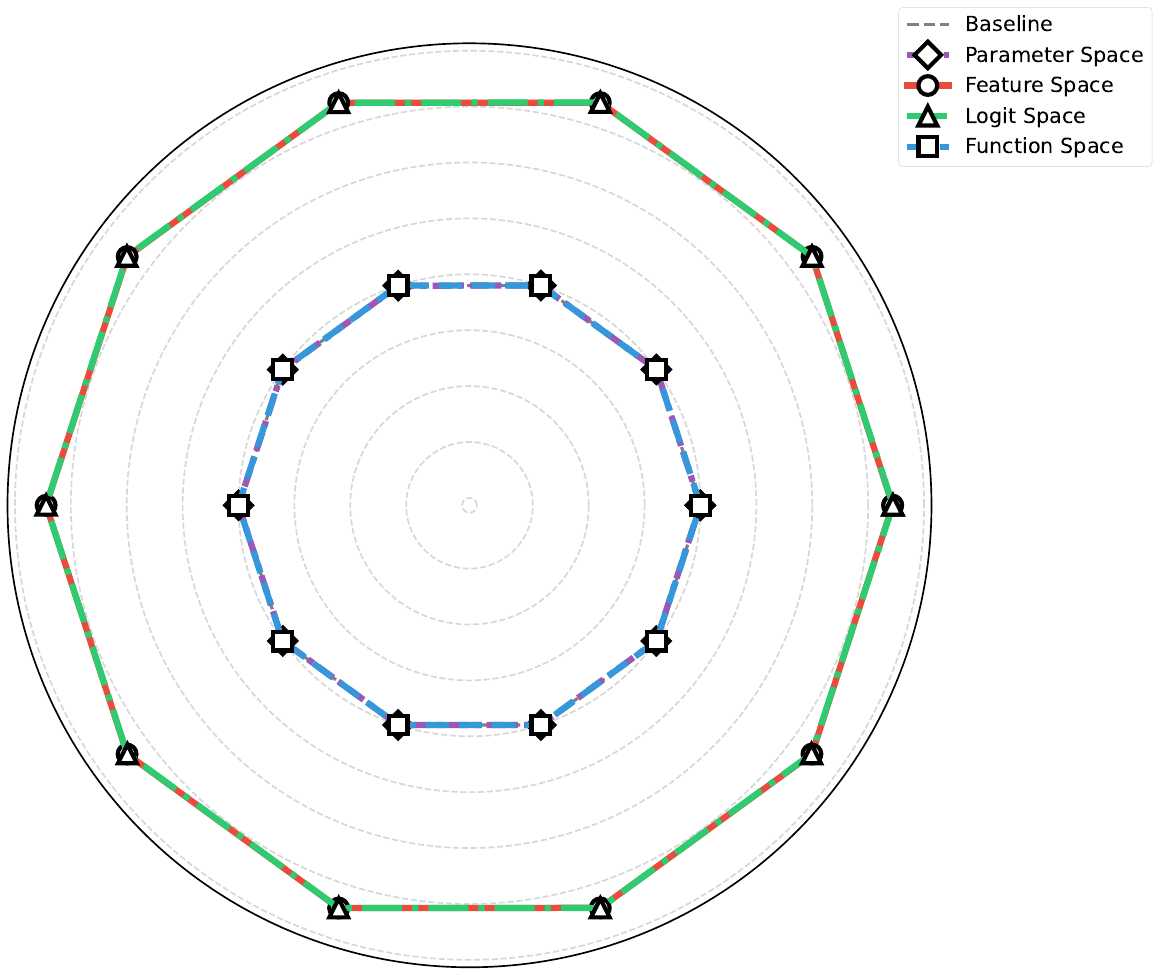}
		\includegraphics[width=0.15\textwidth]{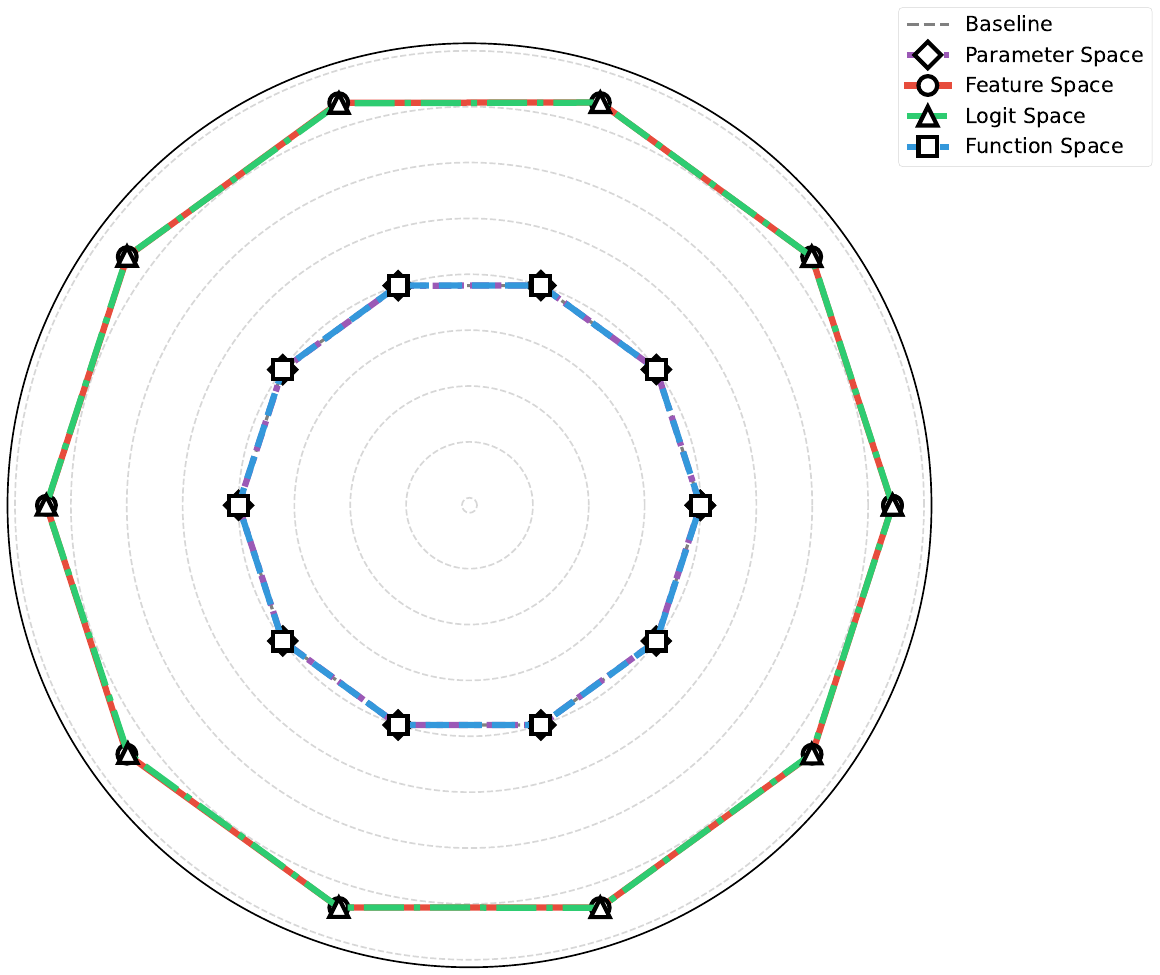}
		\includegraphics[width=0.15\textwidth]{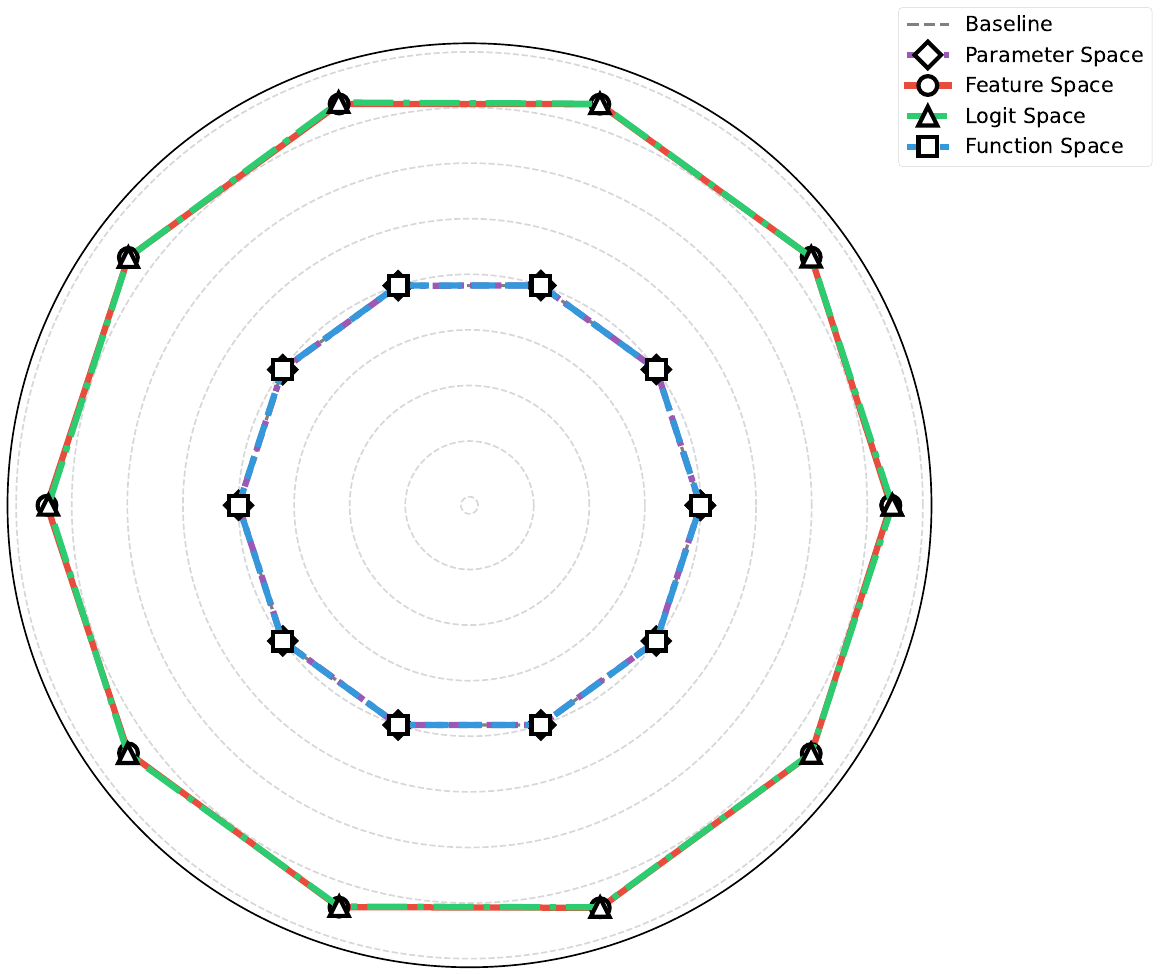}
		\includegraphics[width=0.15\textwidth]{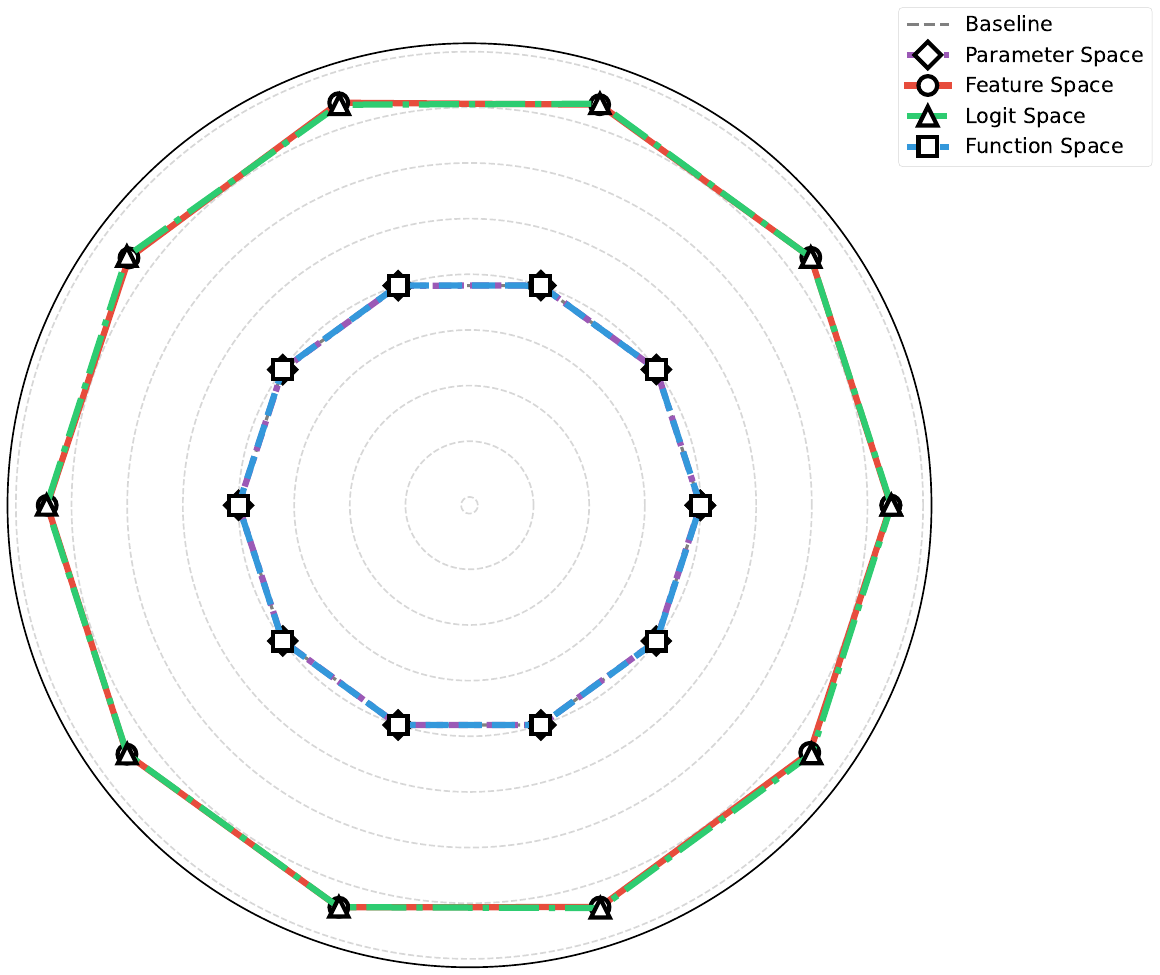}
		\includegraphics[width=0.15\textwidth]{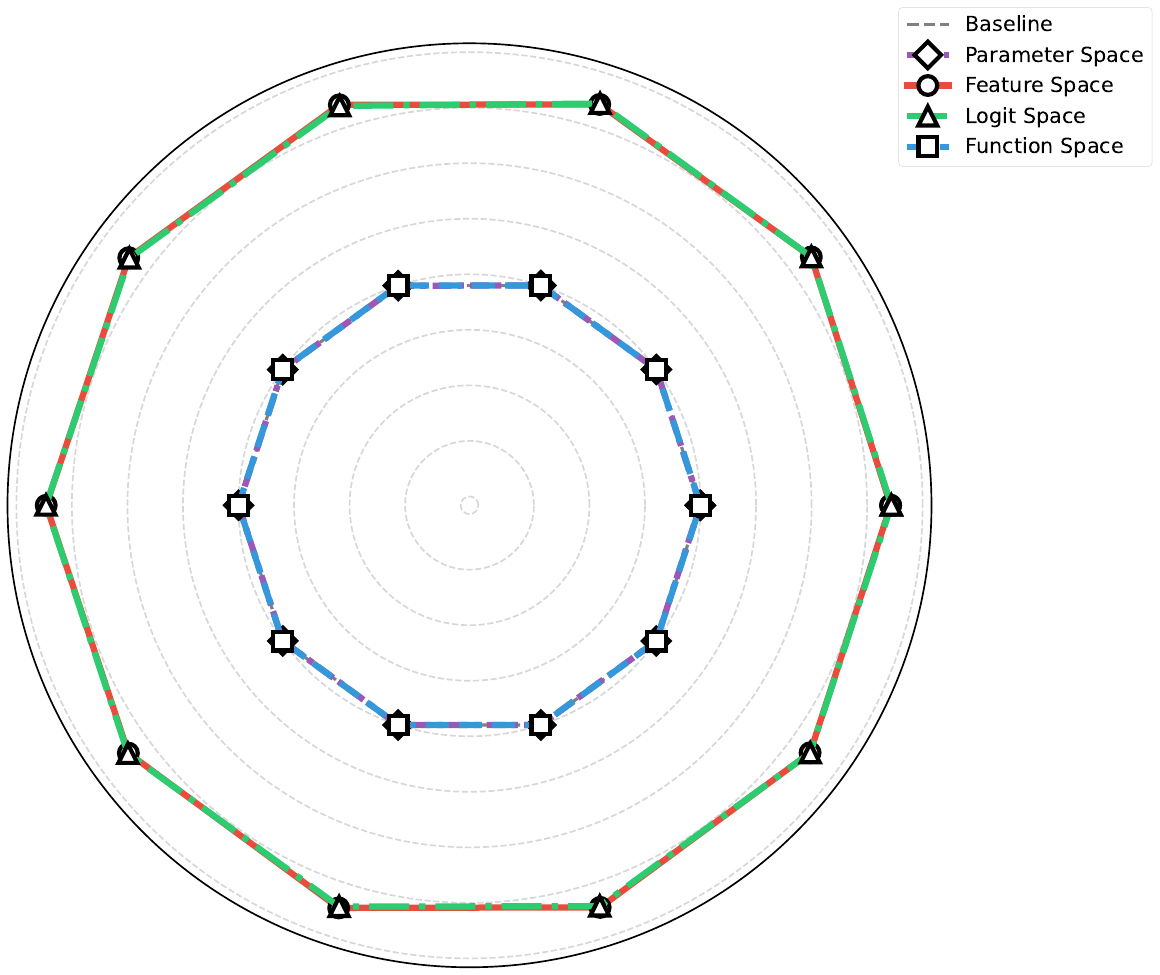}
		\includegraphics[width=0.15\textwidth]{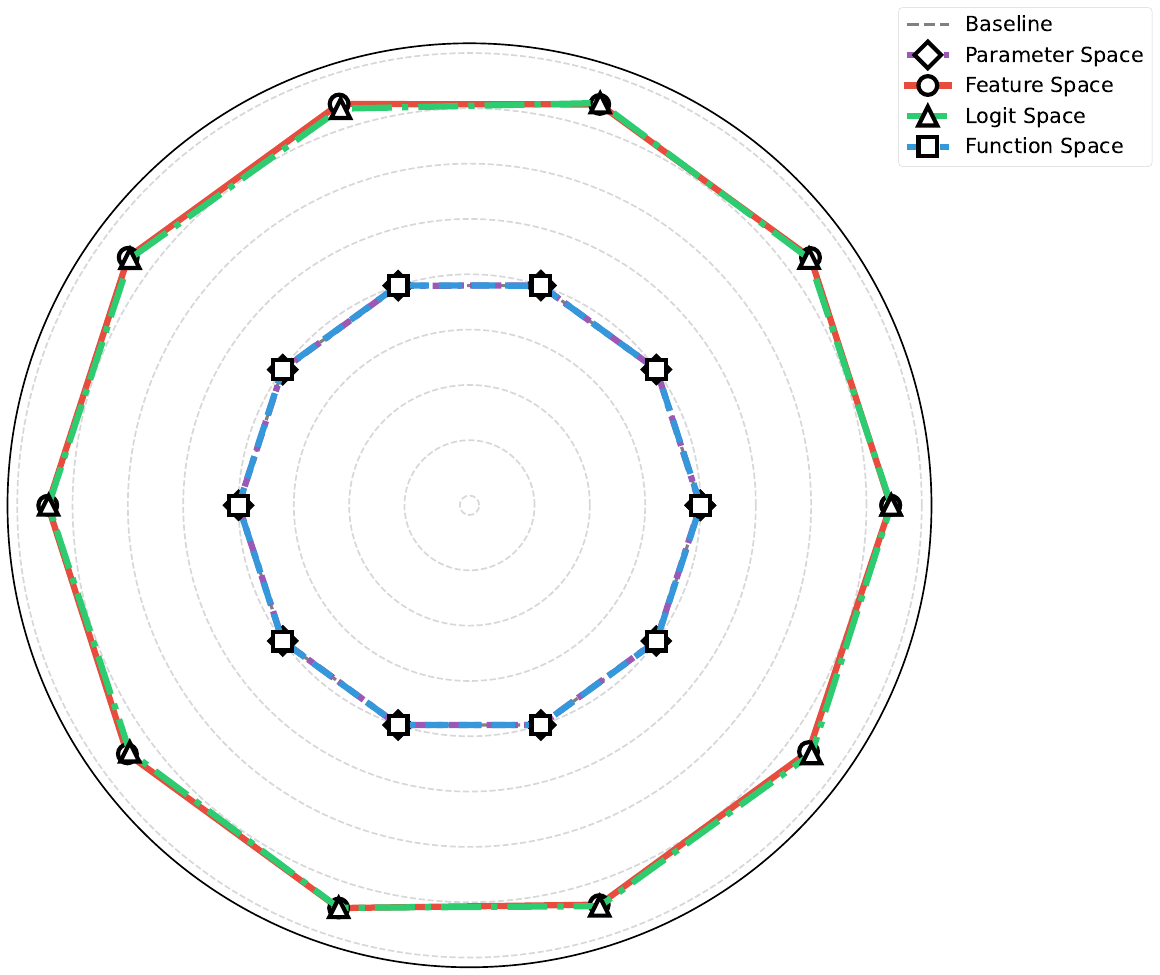}
		\subcaption{Effect of different perturbation magnitude on loss in ImageNetA.}
		\label{fig:row1}
	\end{subfigure}
	\vspace{1ex}
	
	\begin{subfigure}[b]{\textwidth}
		\centering
		\includegraphics[width=0.15\textwidth]{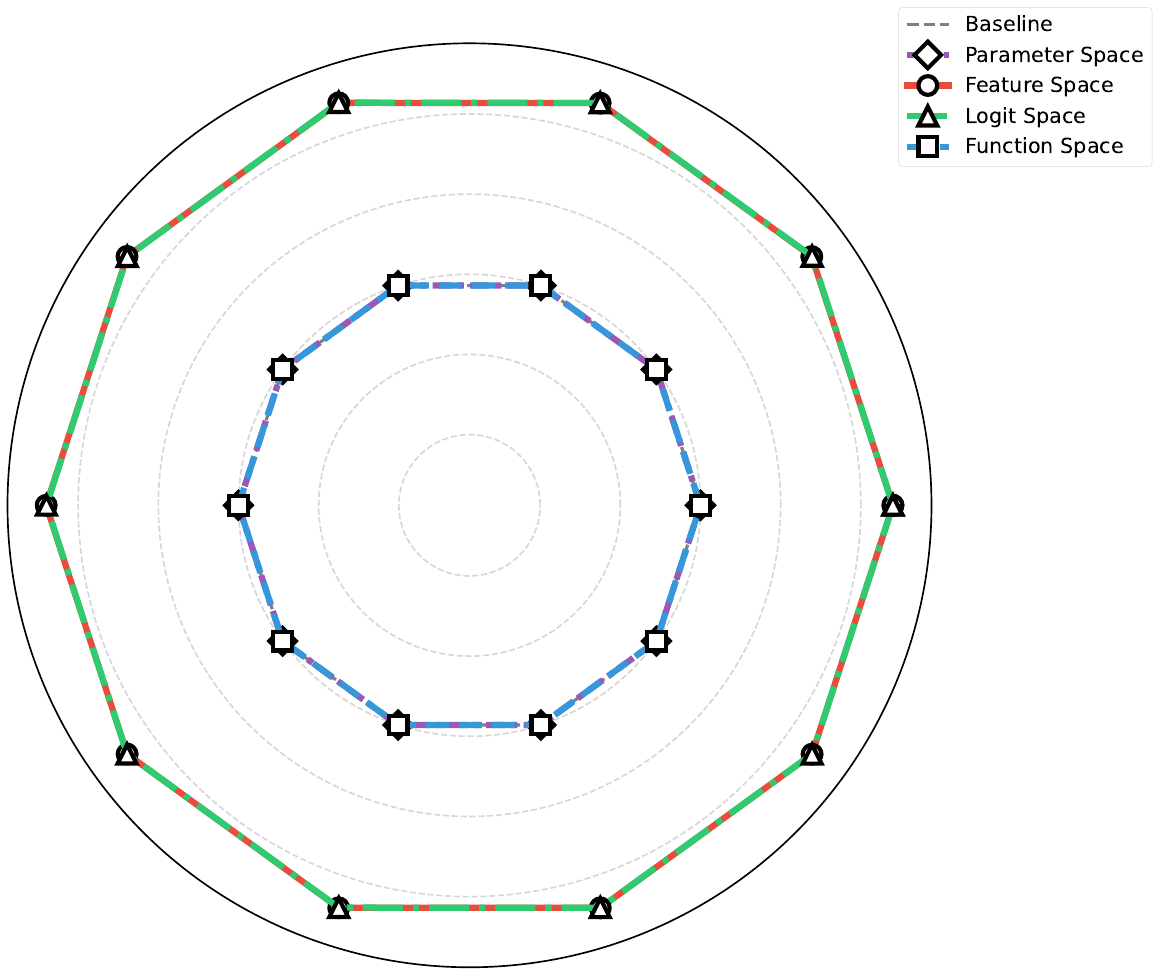}
		\includegraphics[width=0.15\textwidth]{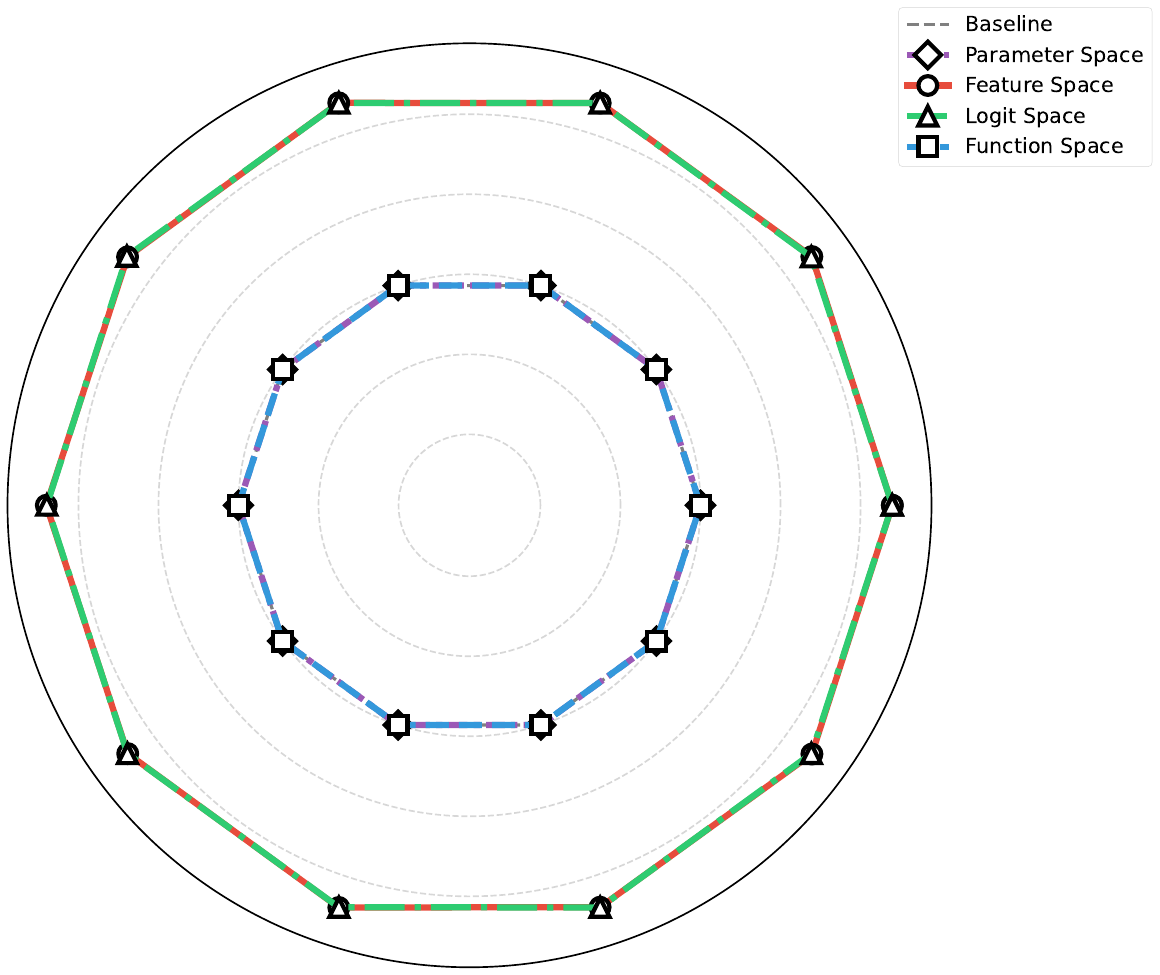}
		\includegraphics[width=0.15\textwidth]{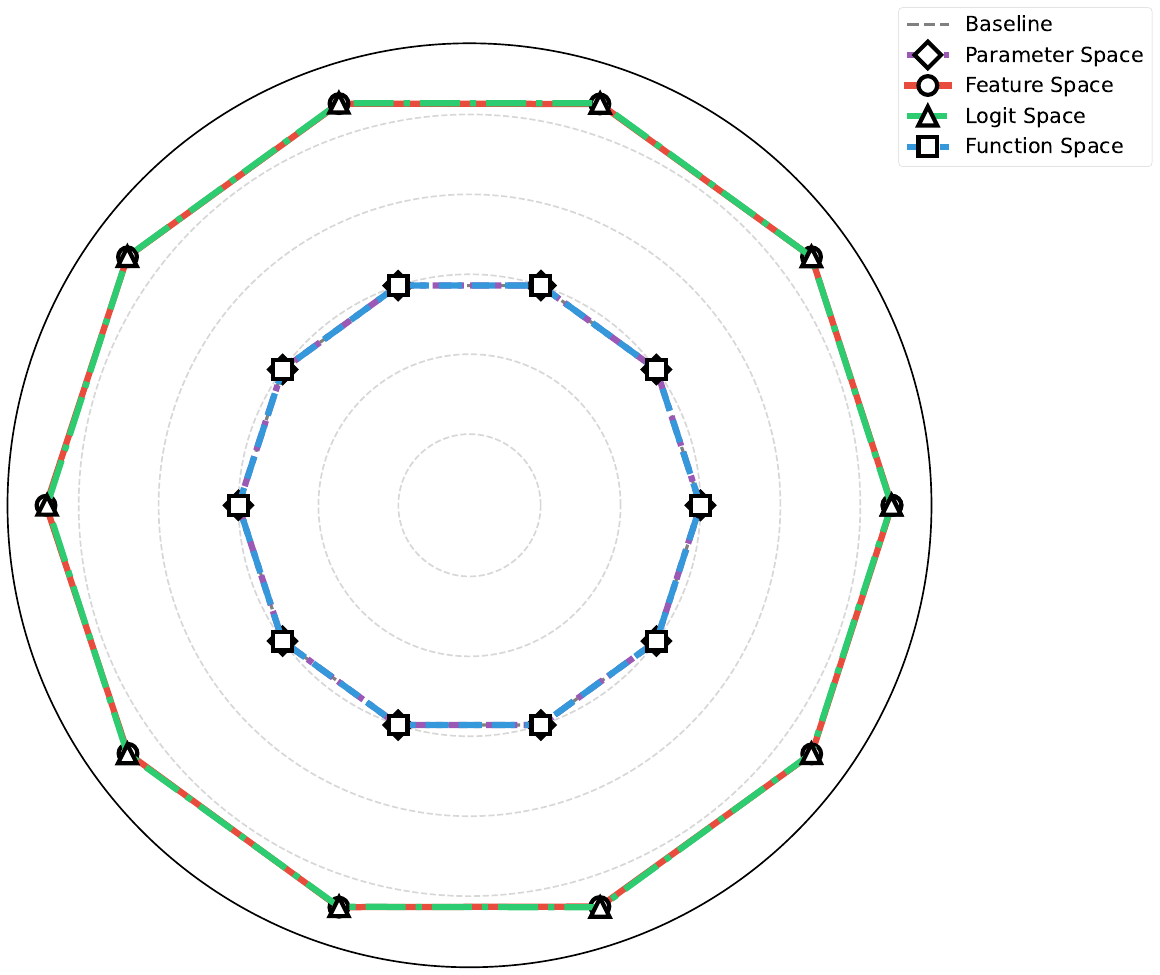}
		\includegraphics[width=0.15\textwidth]{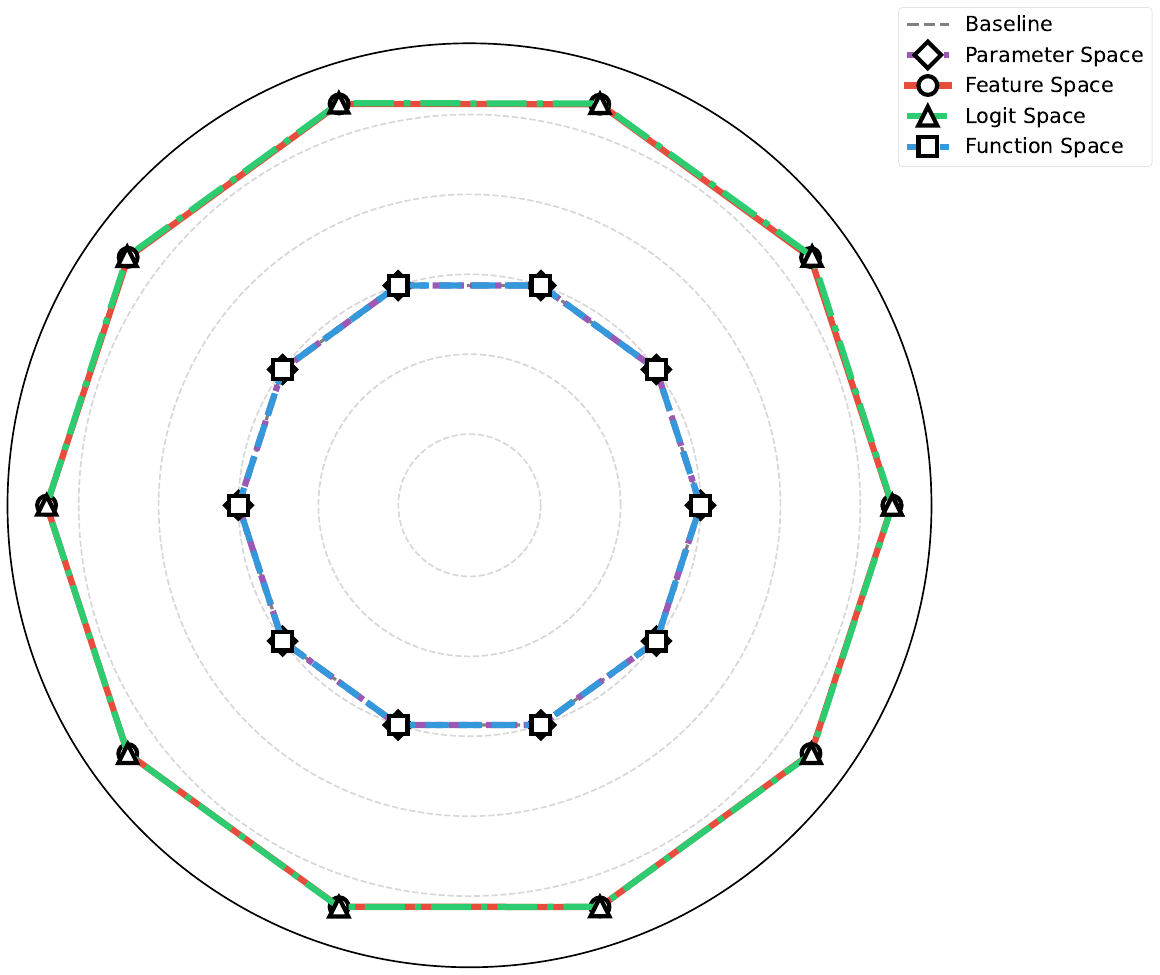}
		\includegraphics[width=0.15\textwidth]{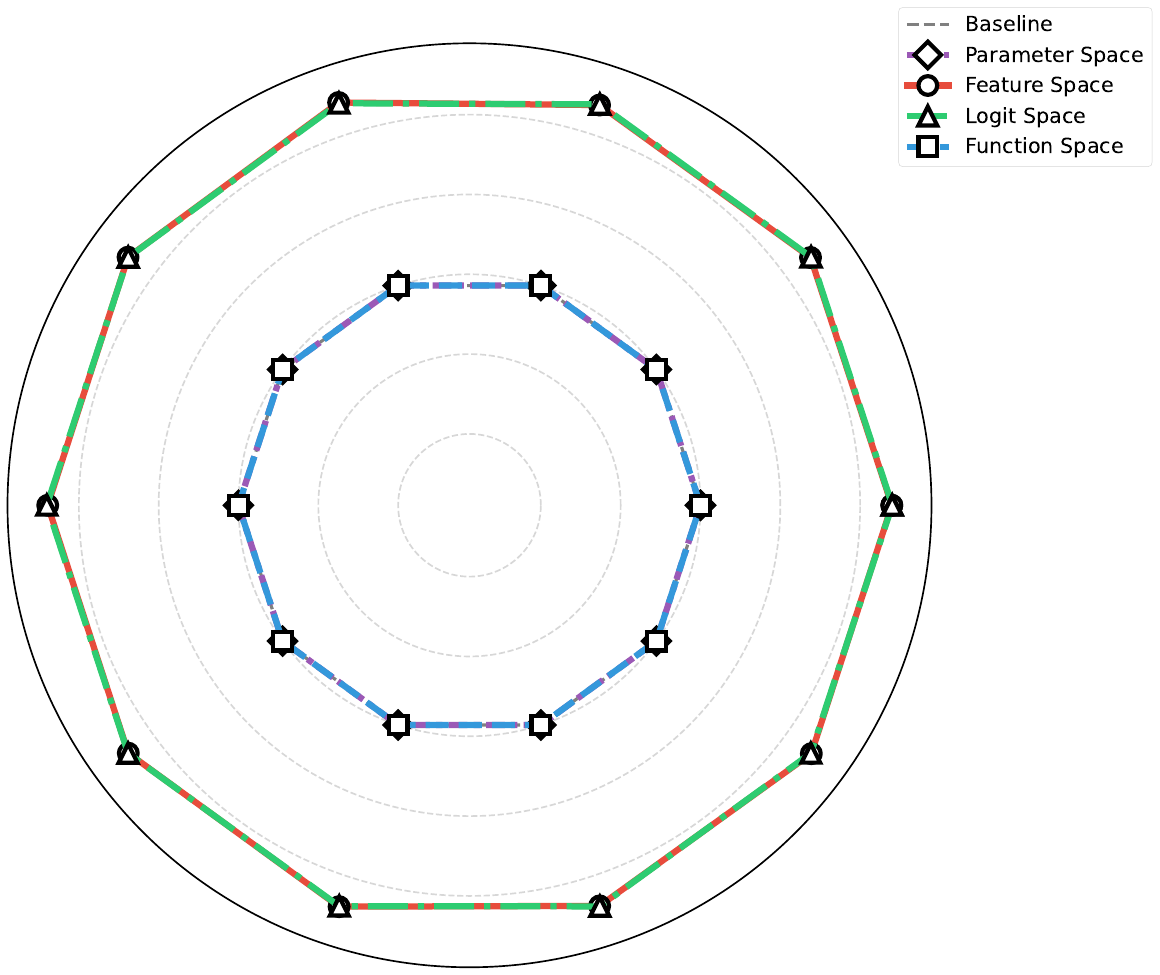}
		\includegraphics[width=0.15\textwidth]{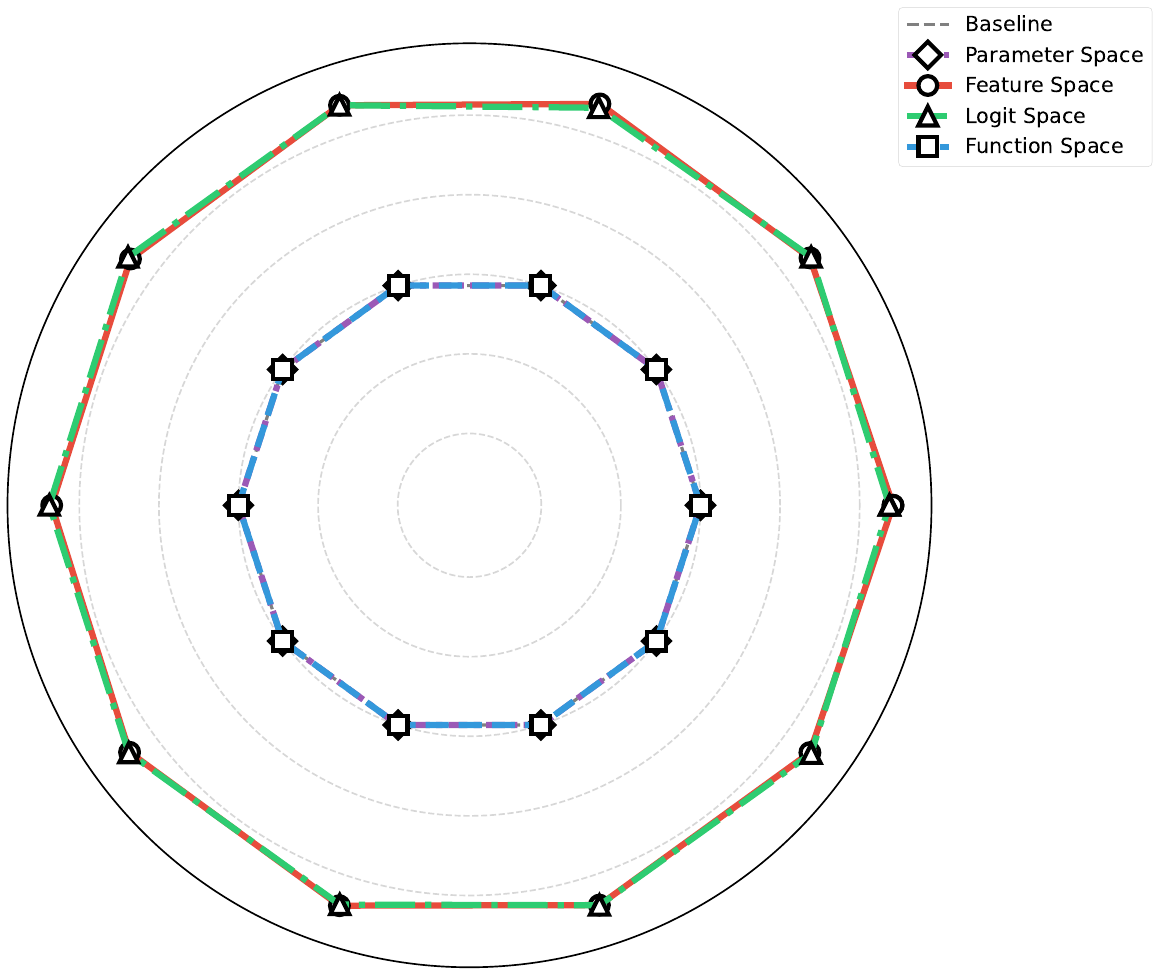}
		\subcaption{Effect of different perturbation magnitude on loss in ImageNetR.}
		\label{fig:row1}
	\end{subfigure}
	\vspace{1ex}
	
	\begin{subfigure}[b]{\textwidth}
		\centering
		\includegraphics[width=0.15\textwidth]{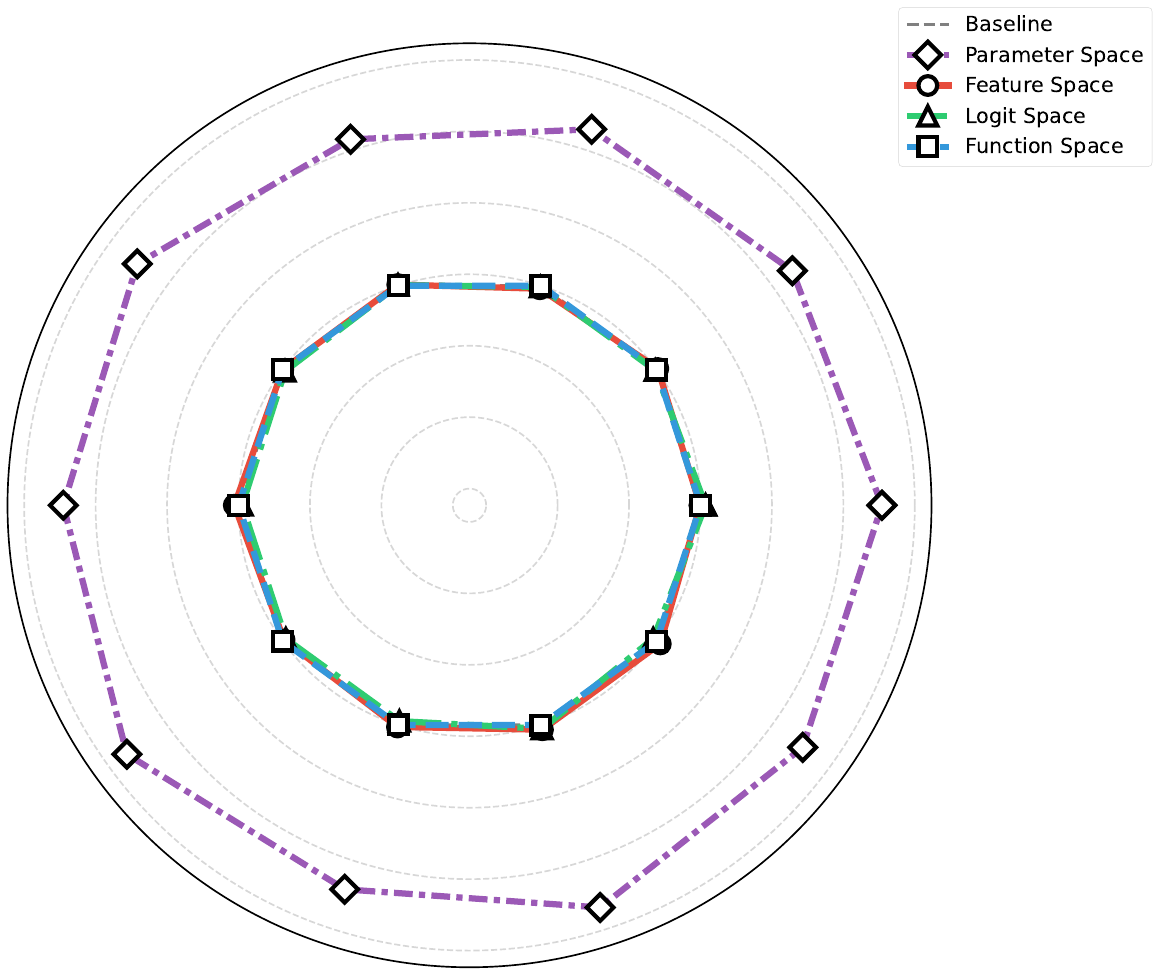}
		\includegraphics[width=0.15\textwidth]{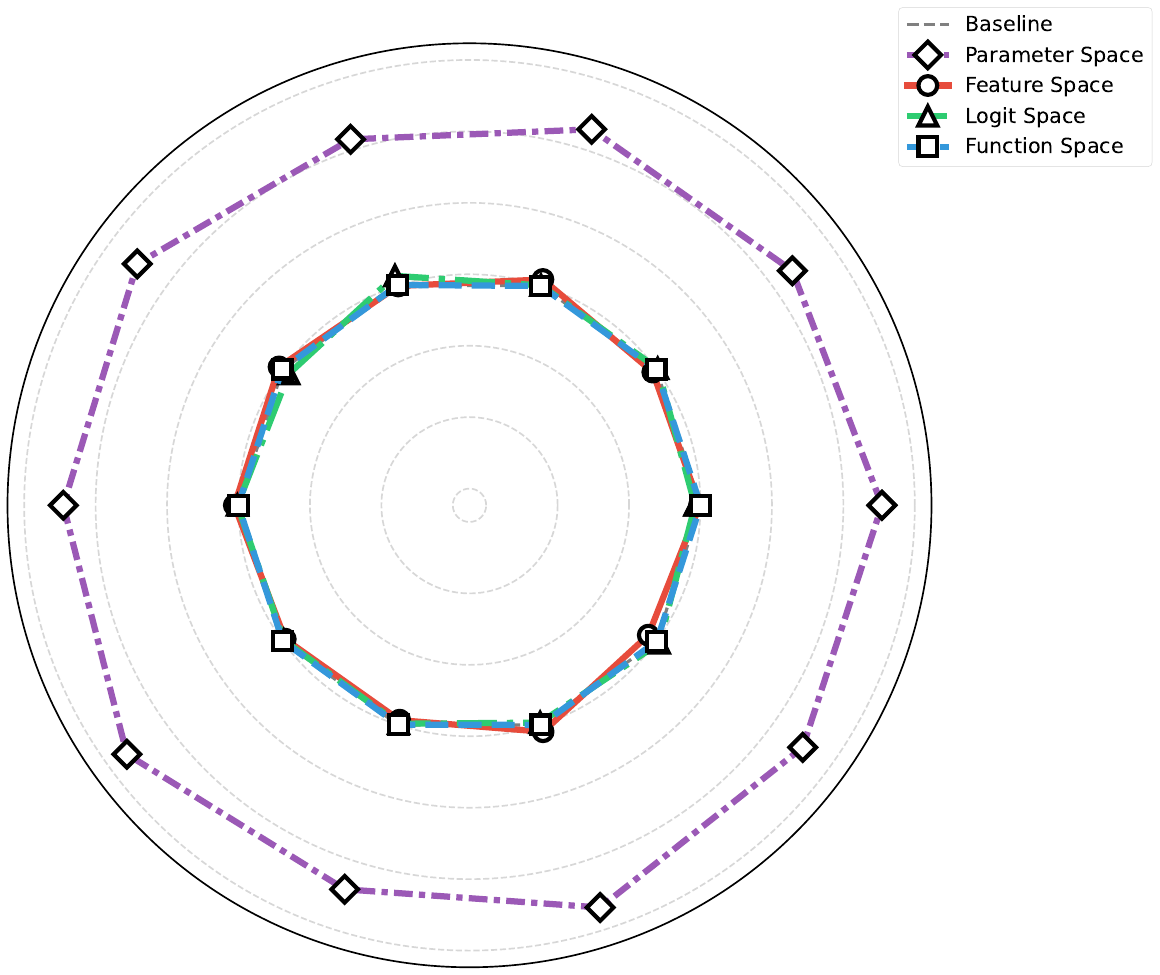}
		\includegraphics[width=0.15\textwidth]{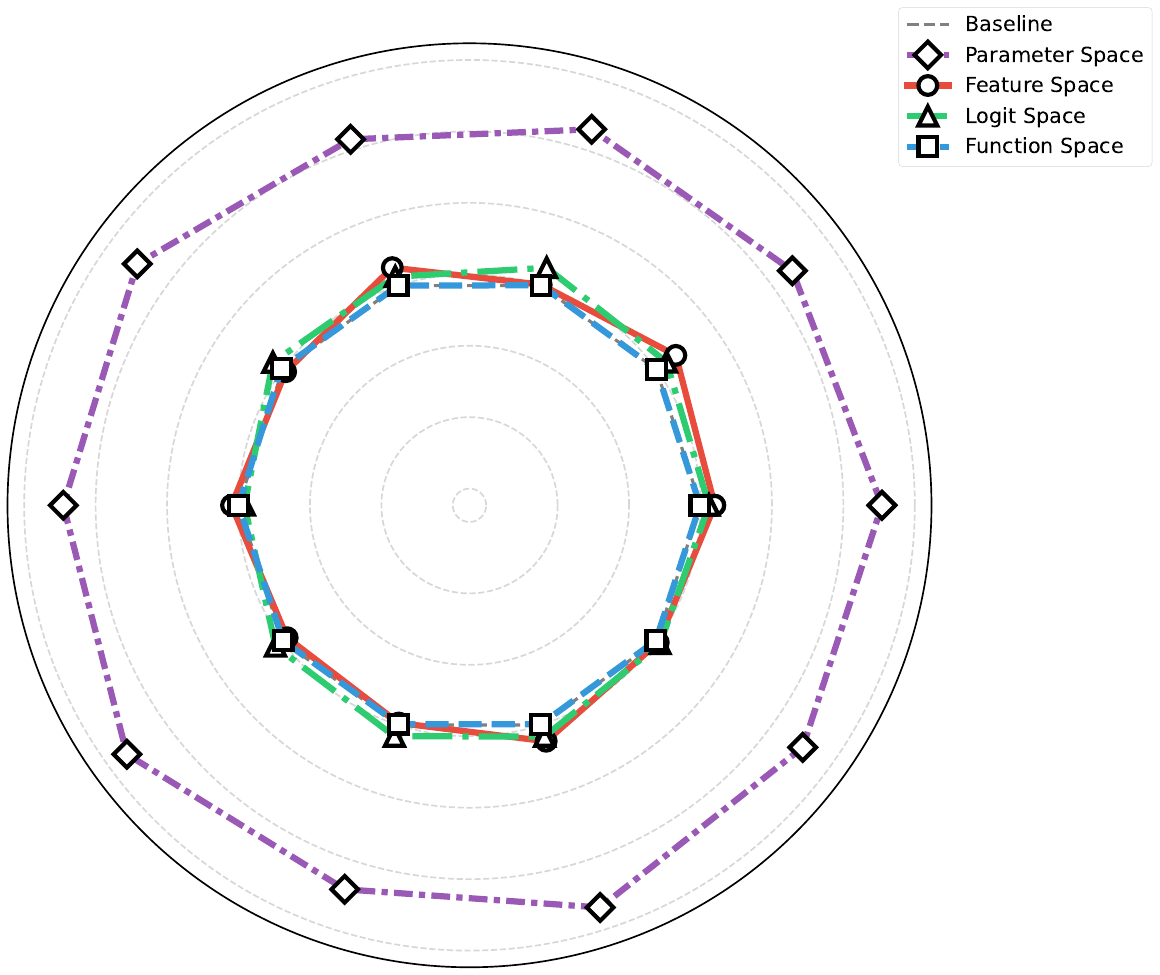}
		\includegraphics[width=0.15\textwidth]{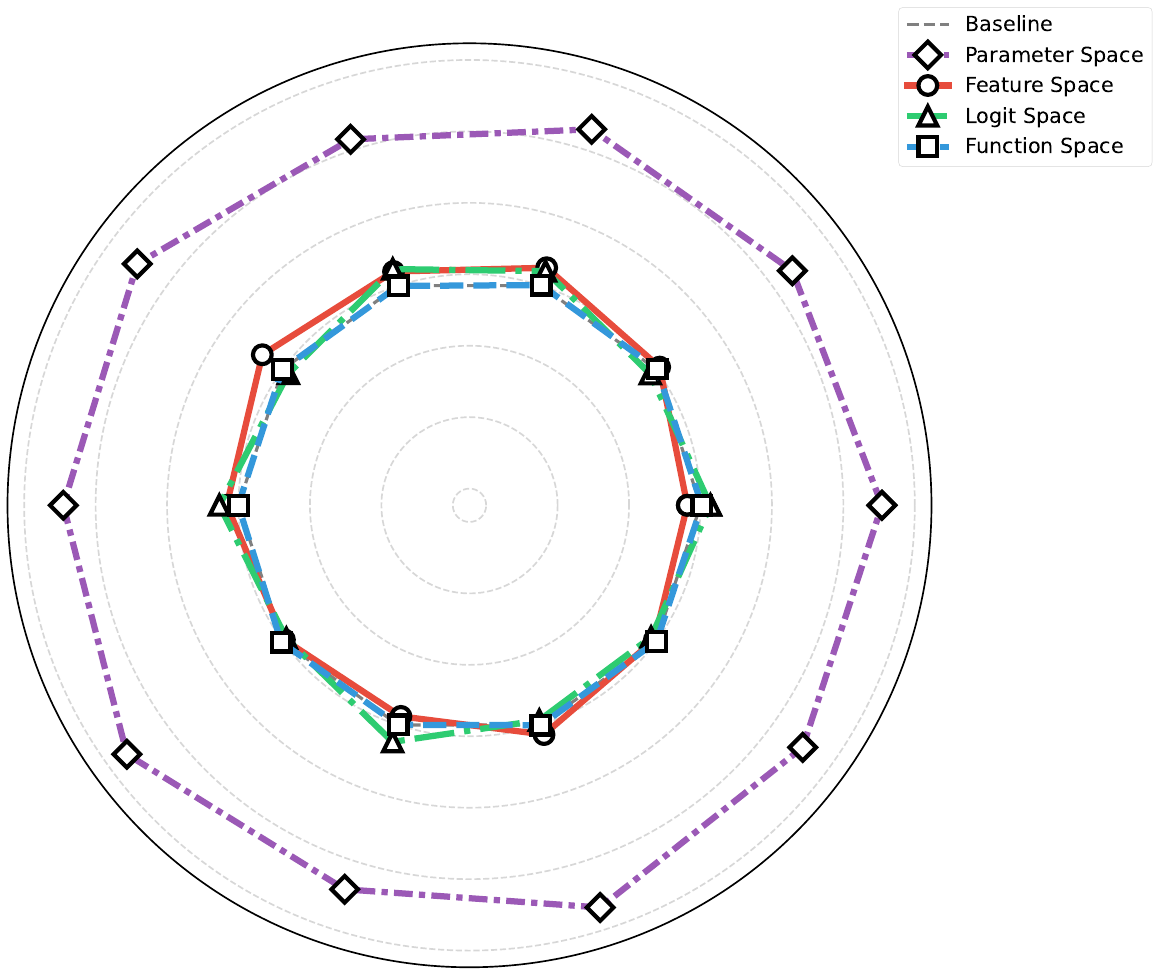}
		\includegraphics[width=0.15\textwidth]{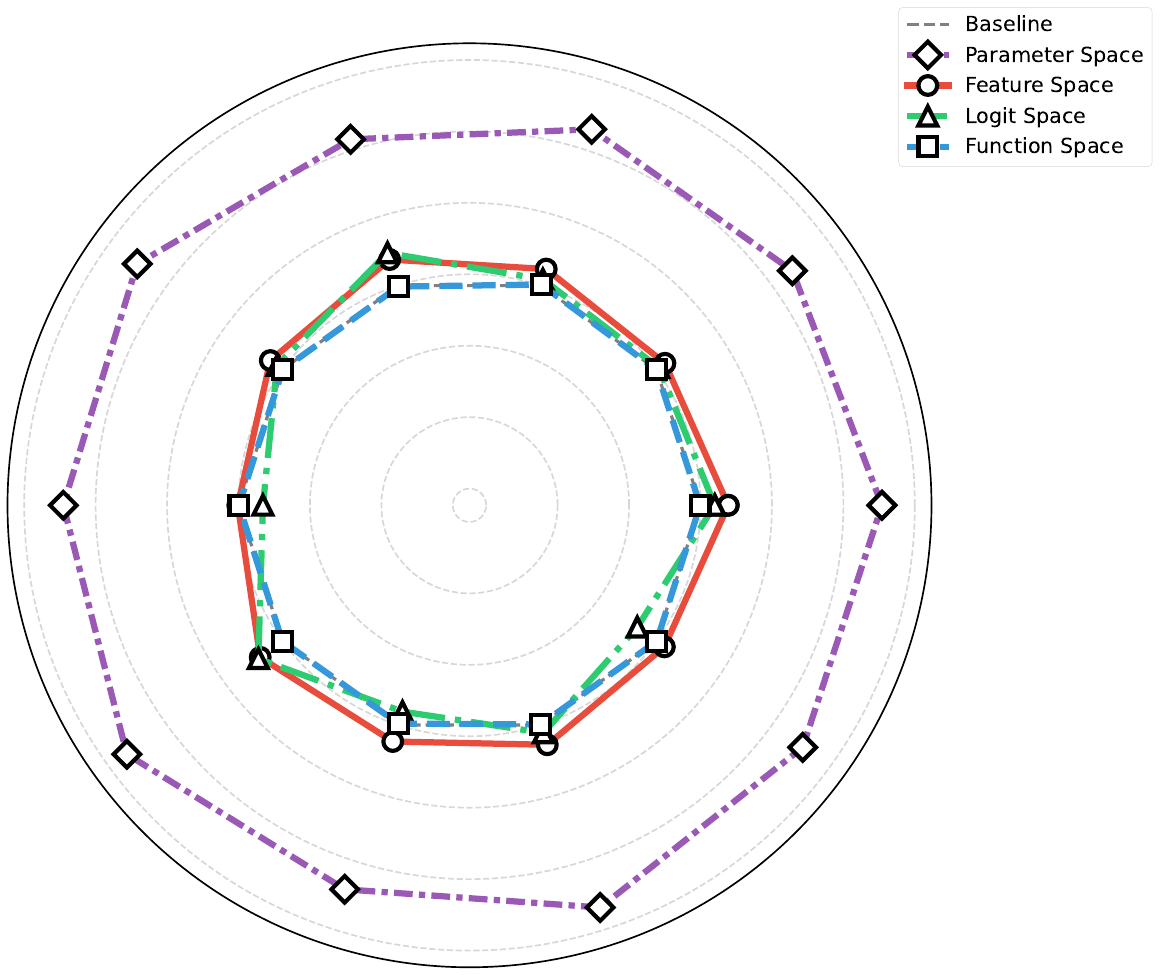}
		\includegraphics[width=0.15\textwidth]{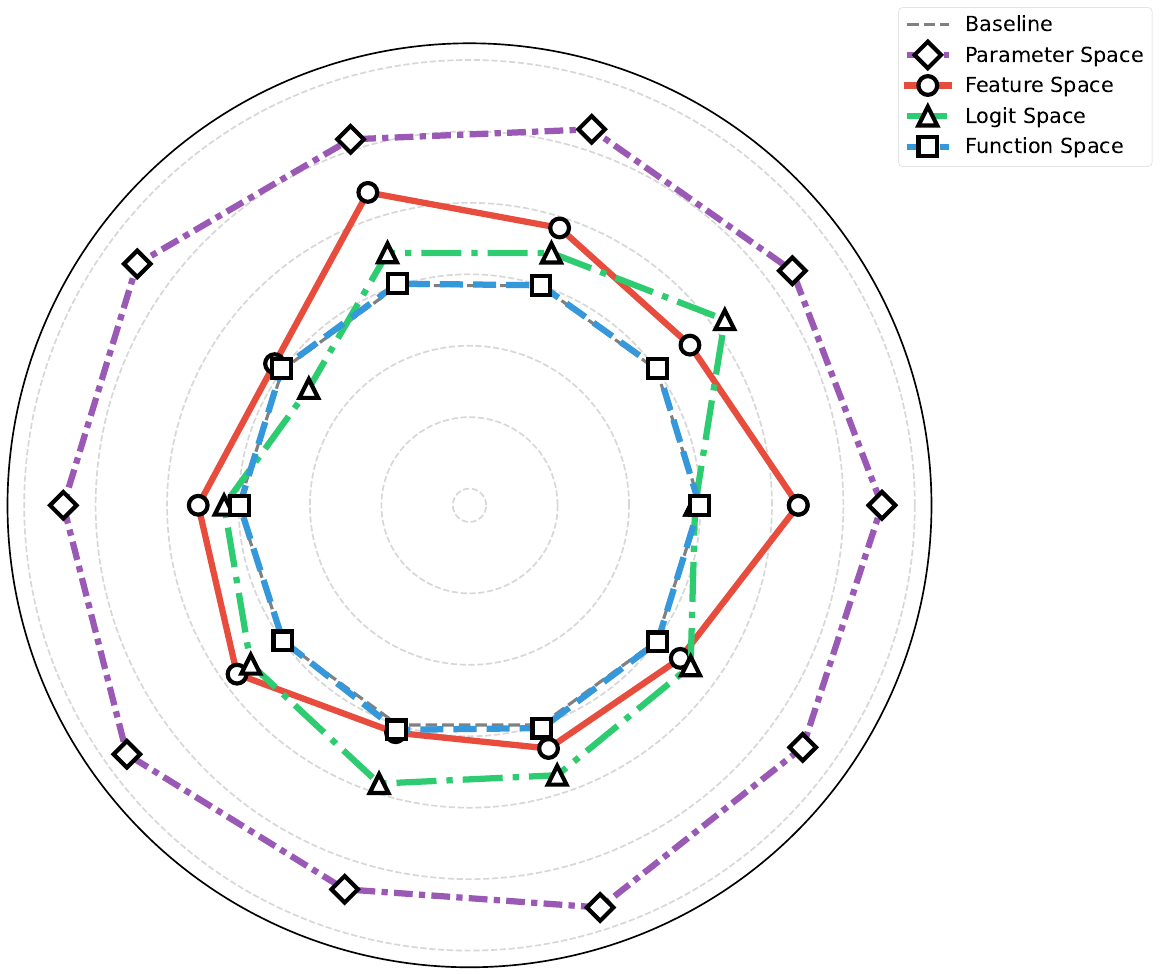}
		\subcaption{Effect of different perturbation magnitude on loss in ImageNetSketch.}
		\label{fig:row1}
	\end{subfigure}
	\vspace{1ex}

	\caption{Variation in data loss across different perturbation magnitudes in parameter, feature, logit, and function spaces, evaluated on six benchmarks: ImageNet, ImageNetV2, ImageNet-A, ImageNet-R, ImageNet-Sketch. Function space perturbations induce the smallest loss across different cases, demonstrating superior robustness across datasets and perturbation magnitudes.(perturbations magnitude is set to $m=0.0004$ in parameter space and \(m\in\{0.1,\,0.2,\,0.3,\,0.4,\,0.5,\,1.0\} \), as illustrated from left to right in the figure, for feature, logit, and function spaces.)}
	\label{fig:loss_space_comparison}
\end{figure}

\begin{figure}[htbp]
	\centering
	\begin{subfigure}[b]{\textwidth}
		\centering
		\includegraphics[width=0.15\textwidth]{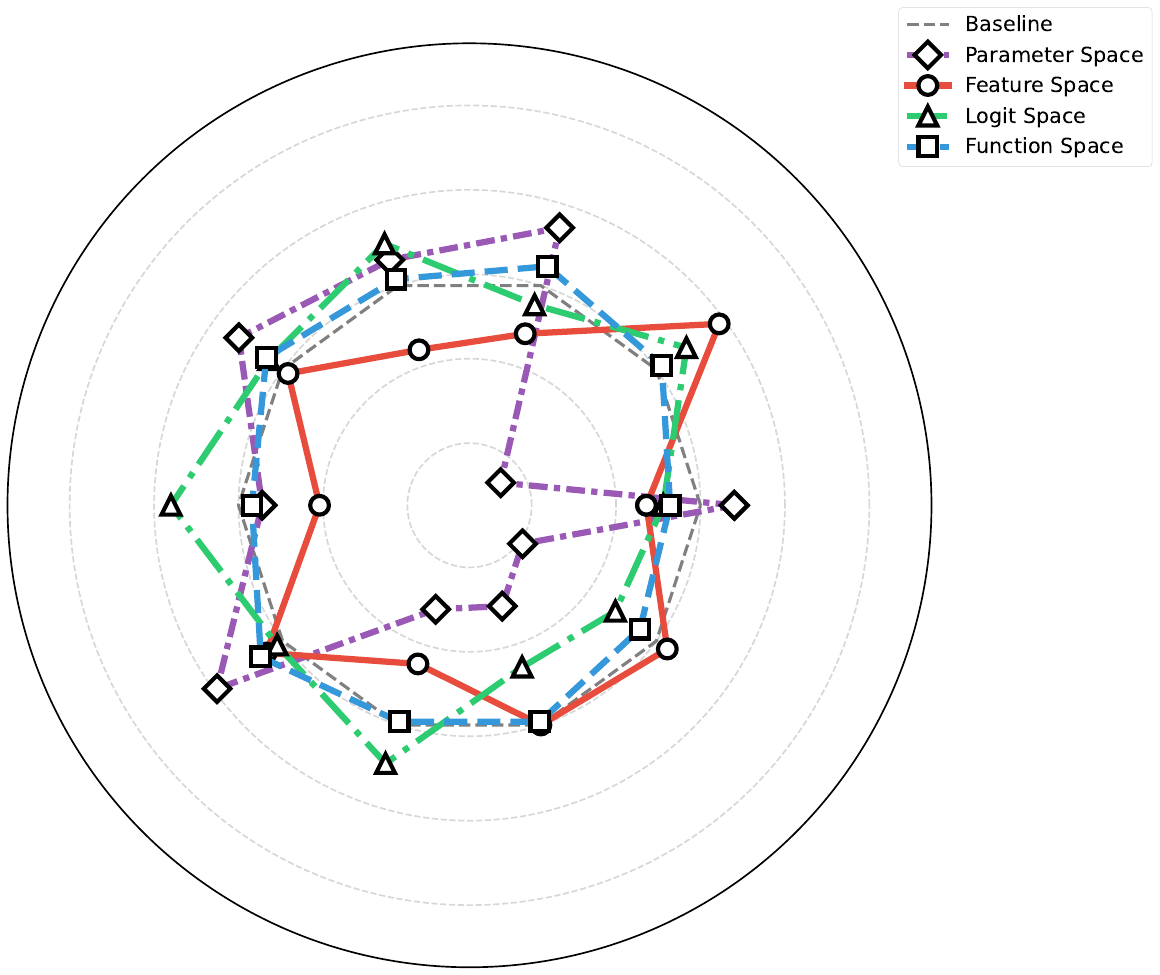}
		\includegraphics[width=0.15\textwidth]{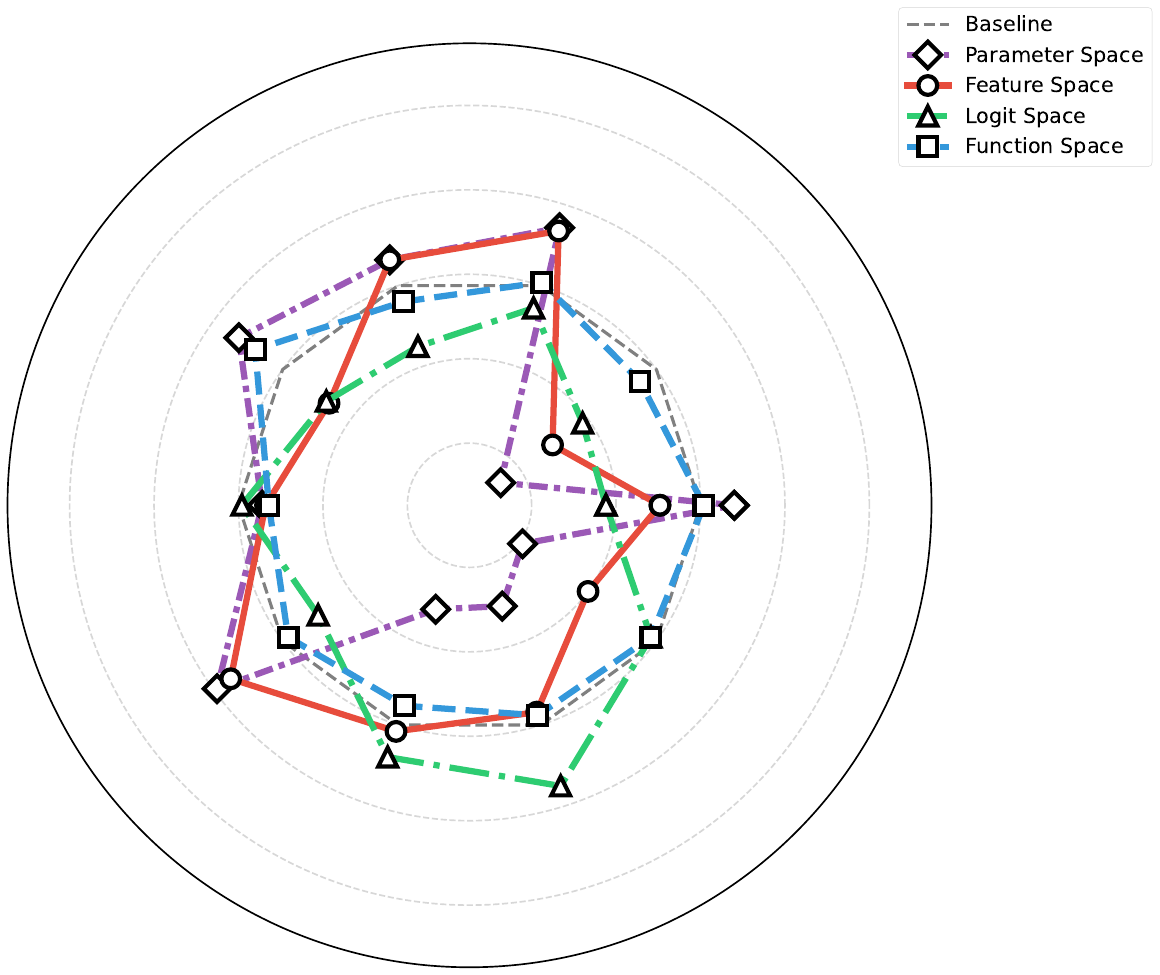}
		\includegraphics[width=0.15\textwidth]{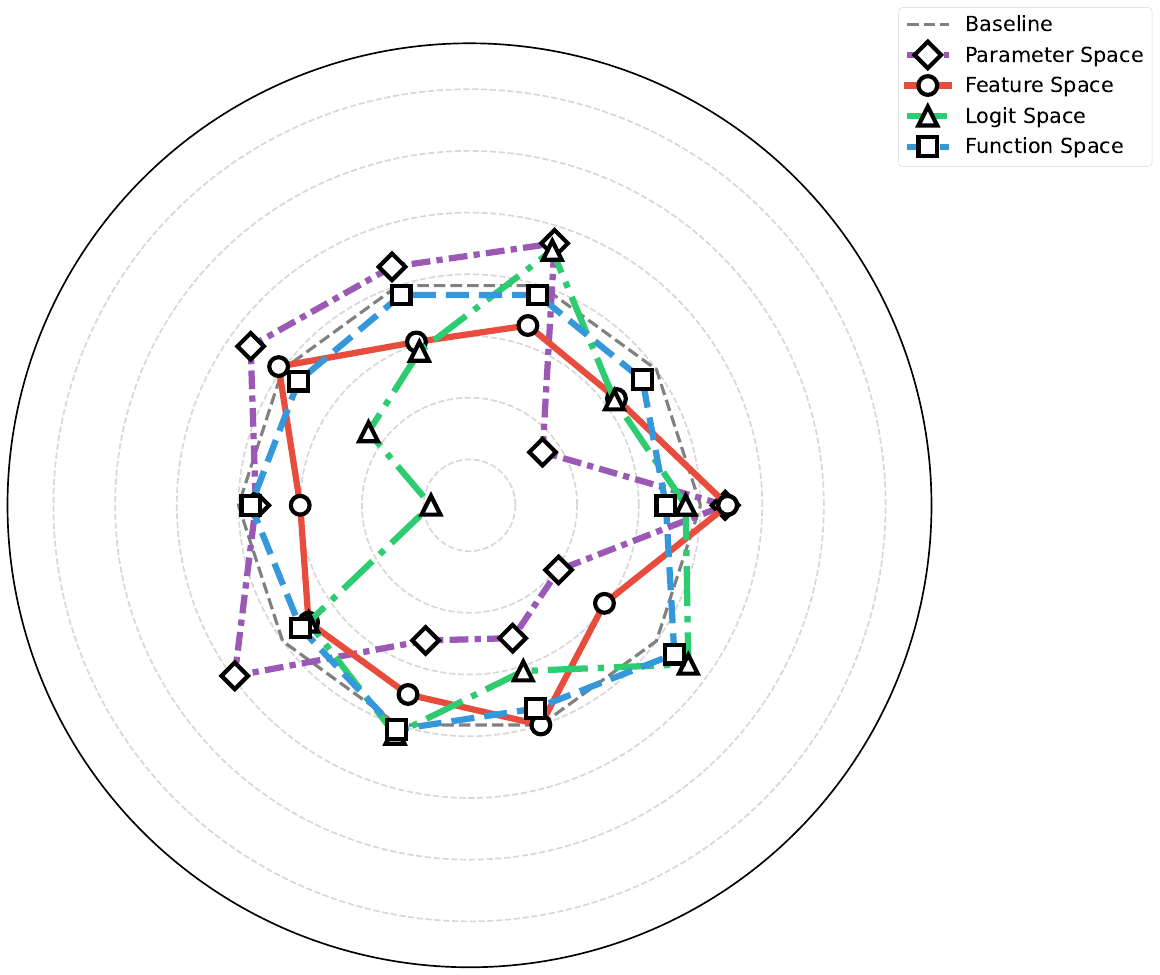}
		\includegraphics[width=0.15\textwidth]{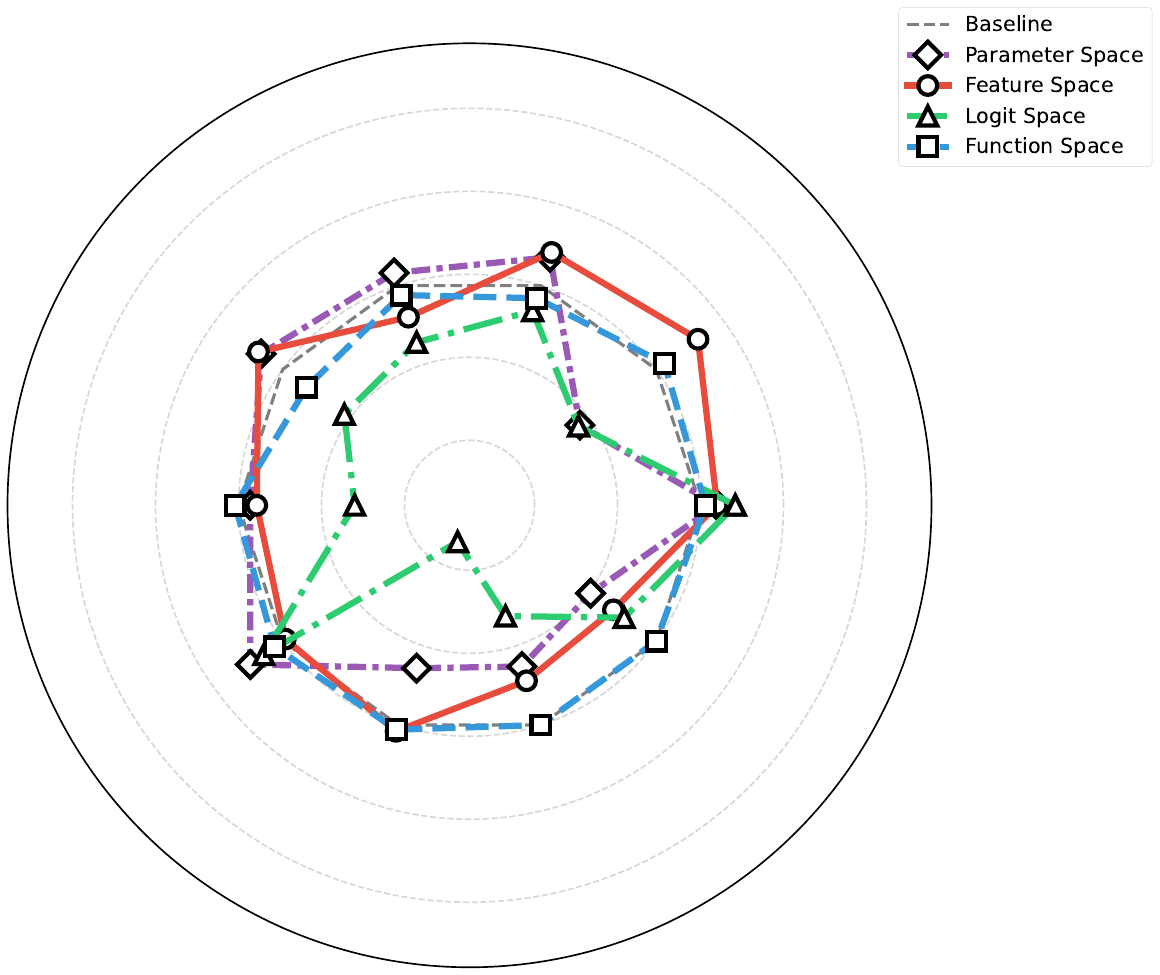}
		\includegraphics[width=0.15\textwidth]{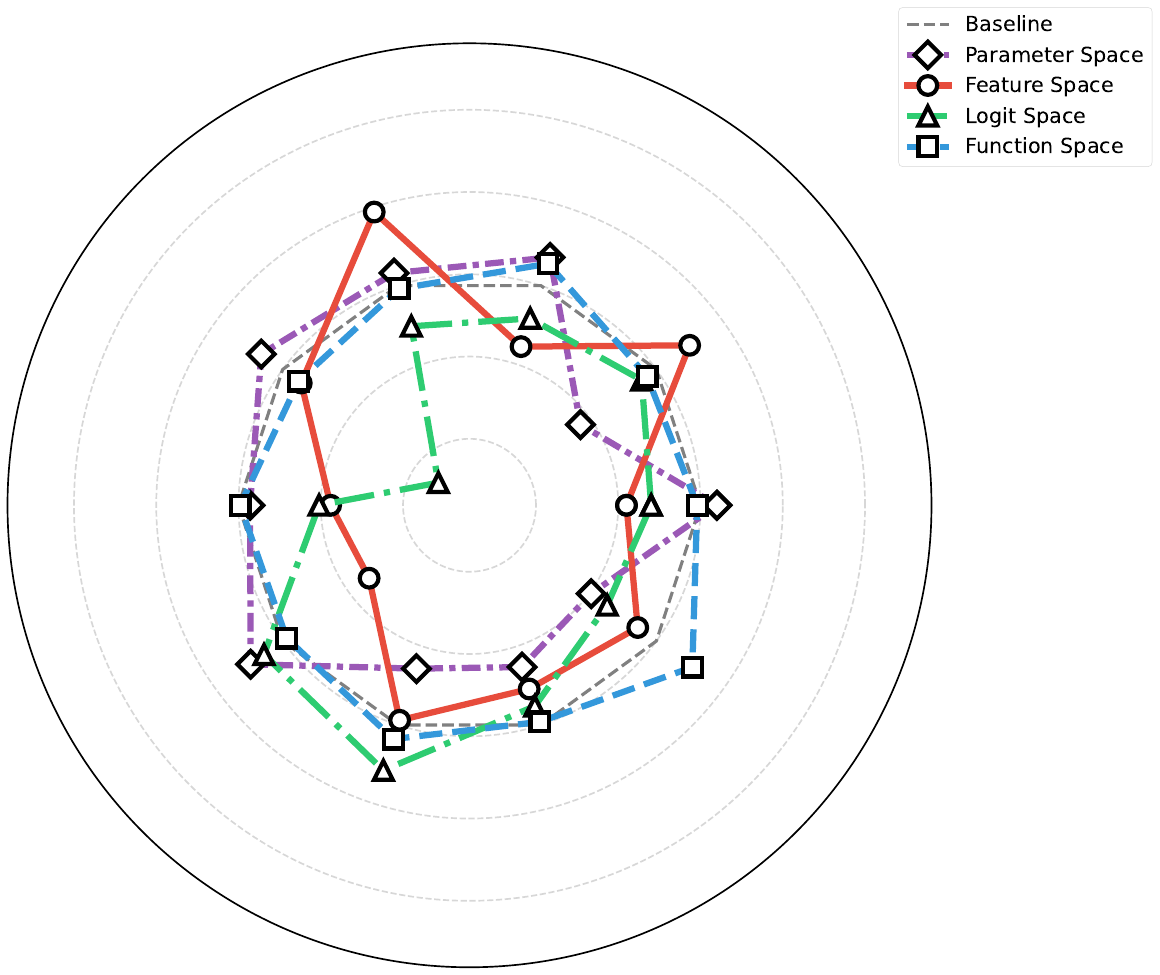}
		\includegraphics[width=0.15\textwidth]{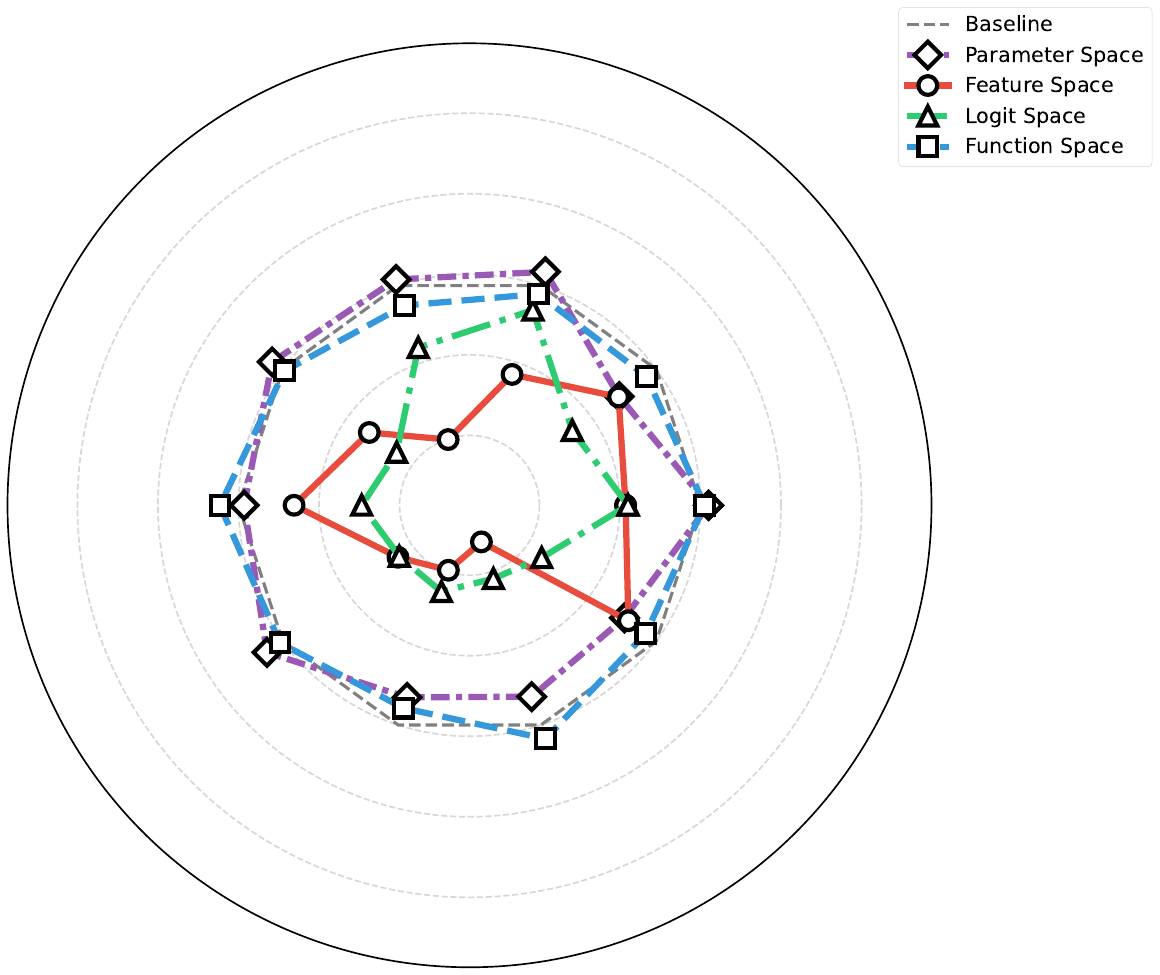}
		\subcaption{Effect of different perturbation magnitude on accuracy in ImageNet.}
		\label{fig:row1}
	\end{subfigure}
	\vspace{1ex}
	
	\begin{subfigure}[b]{\textwidth}
		\centering
		\includegraphics[width=0.15\textwidth]{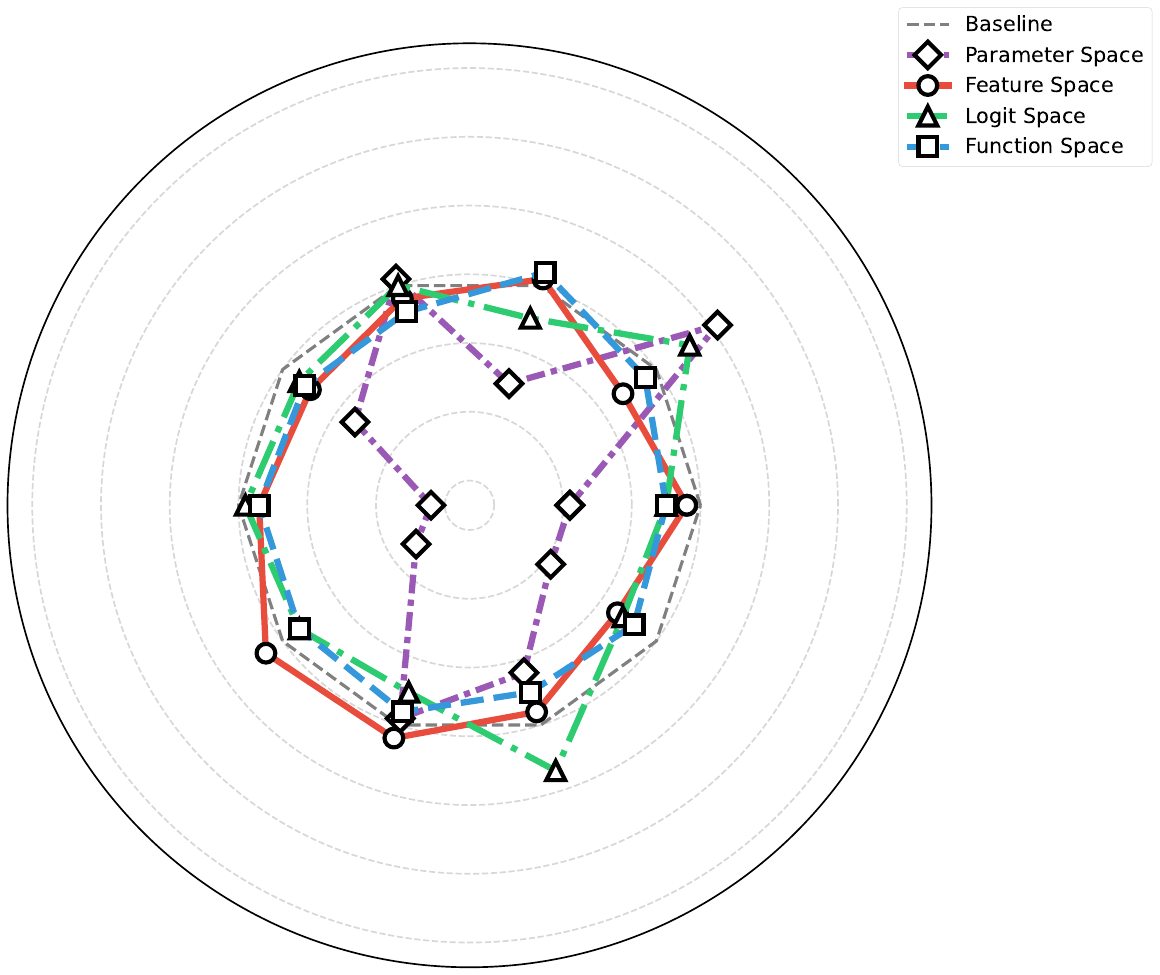}
		\includegraphics[width=0.15\textwidth]{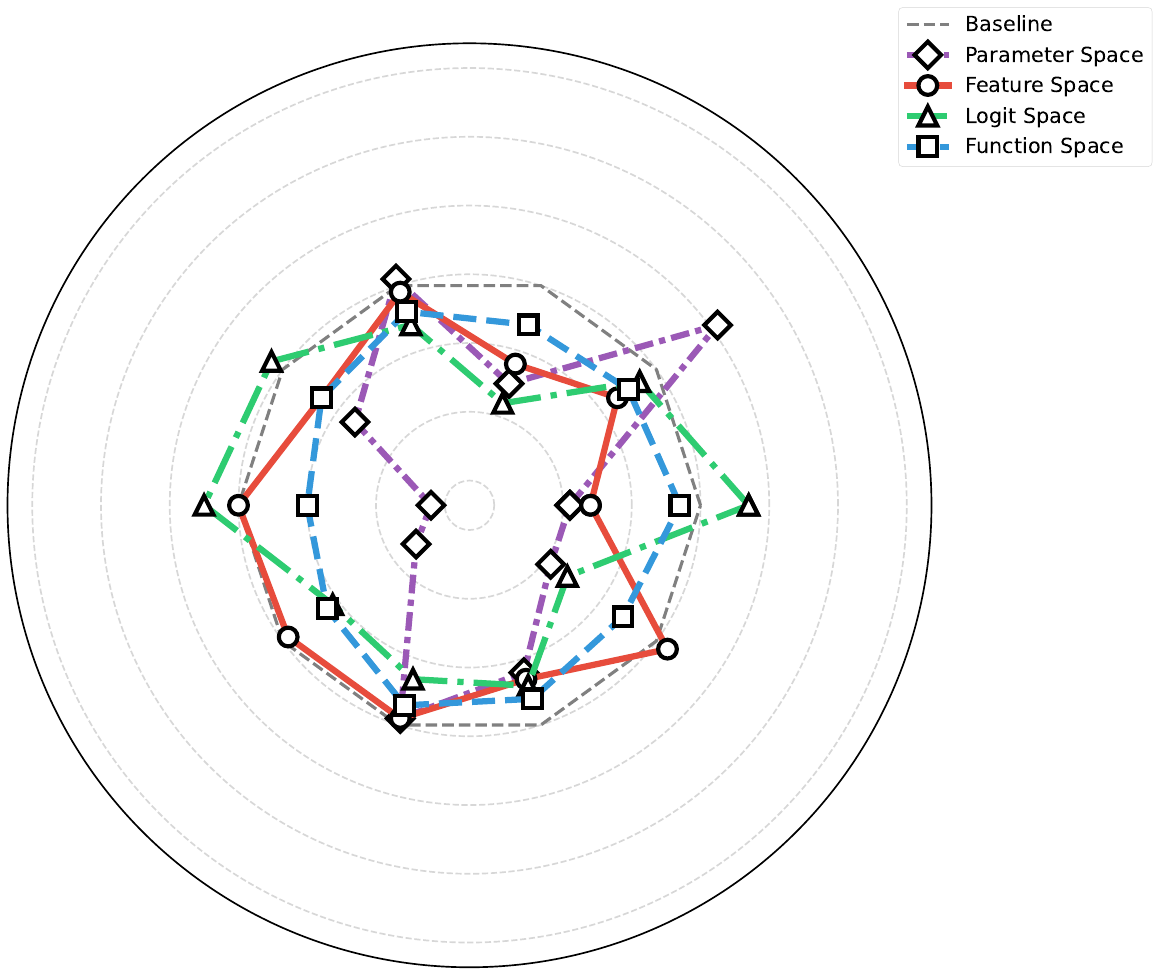}
		\includegraphics[width=0.15\textwidth]{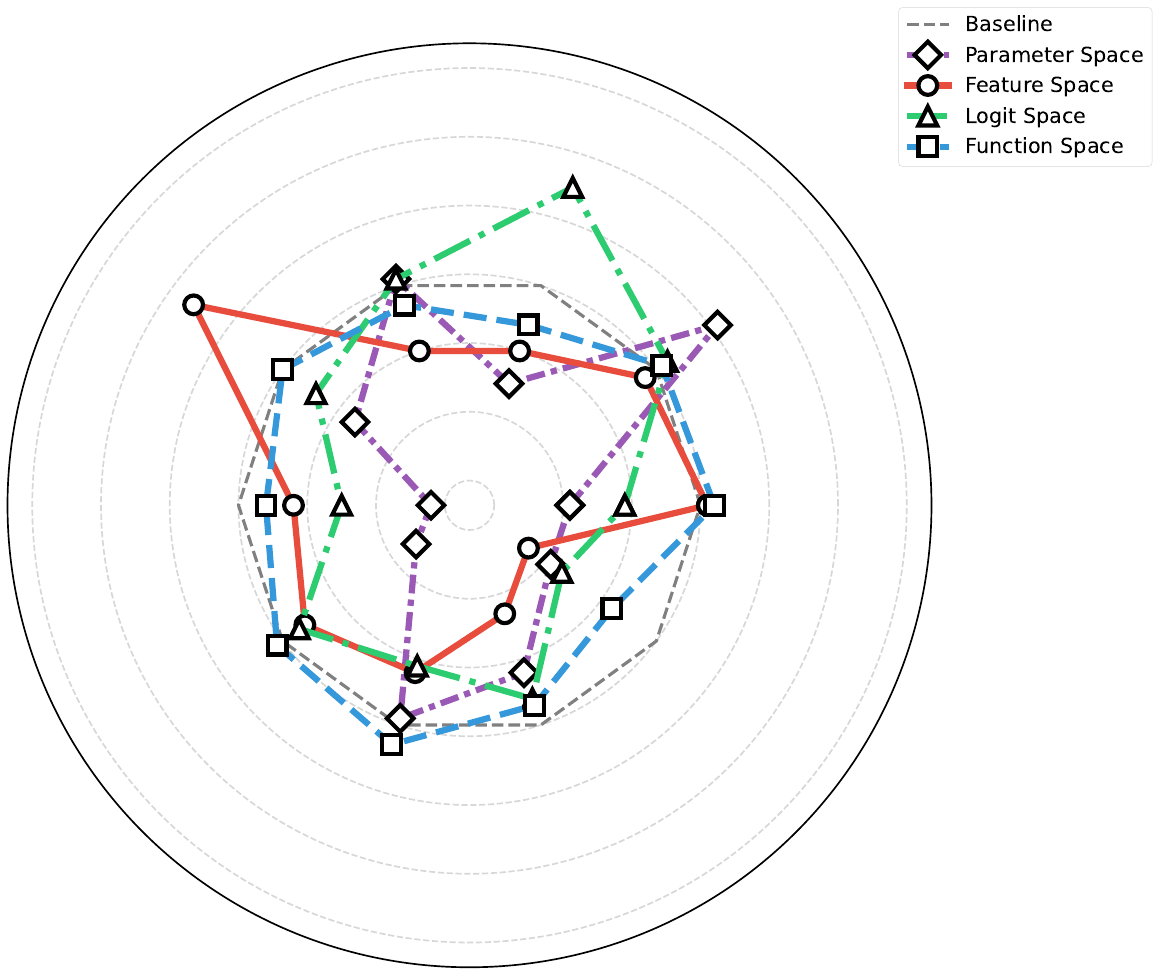}
		\includegraphics[width=0.15\textwidth]{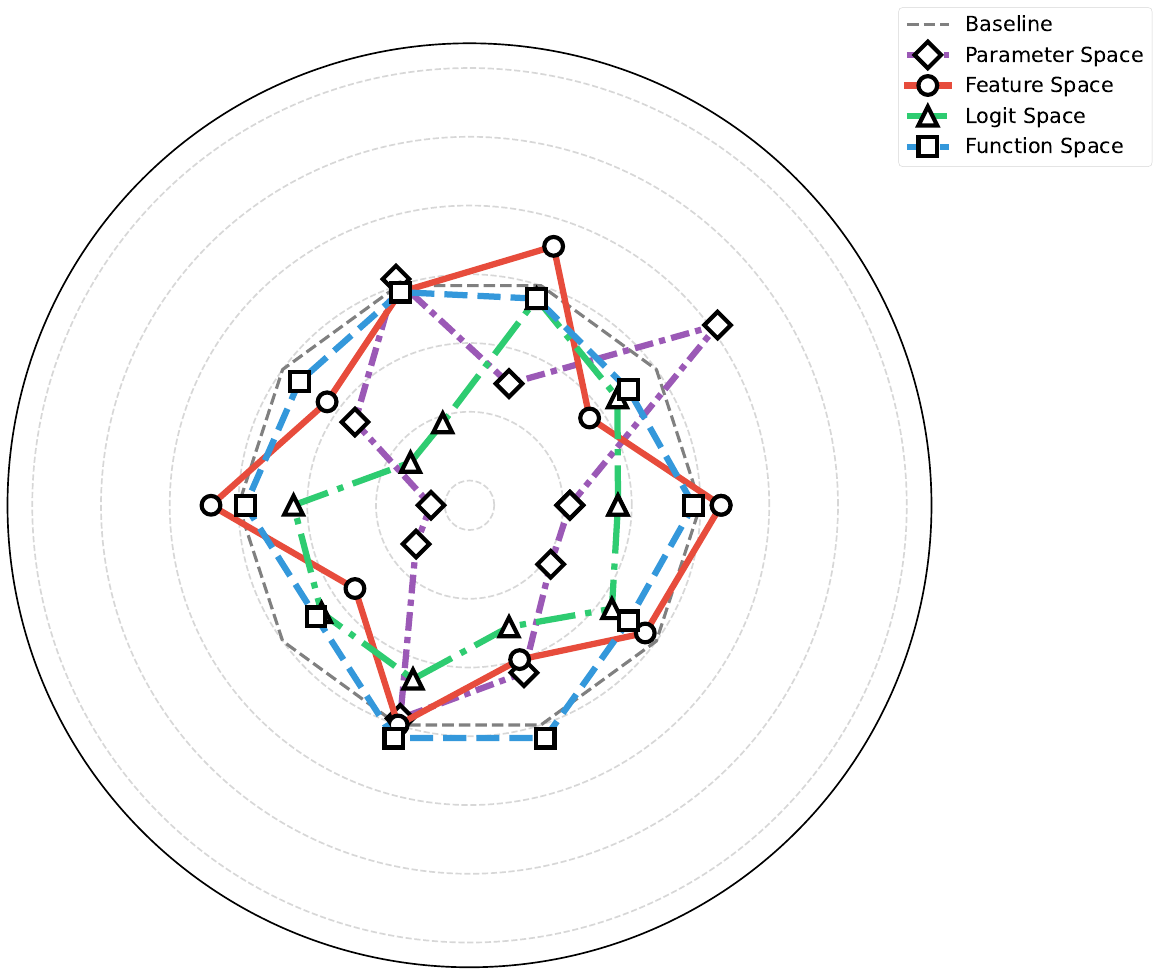}
		\includegraphics[width=0.15\textwidth]{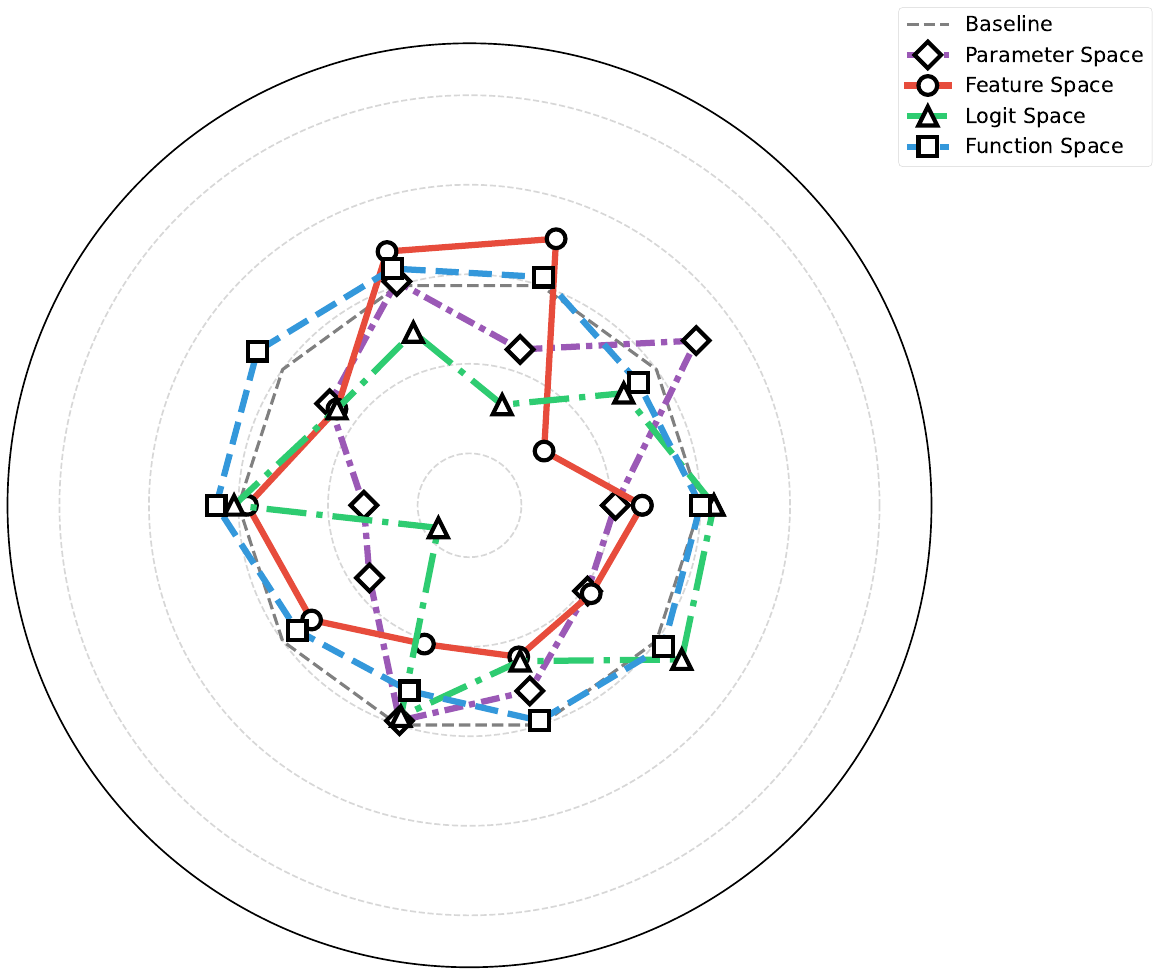}
		\includegraphics[width=0.15\textwidth]{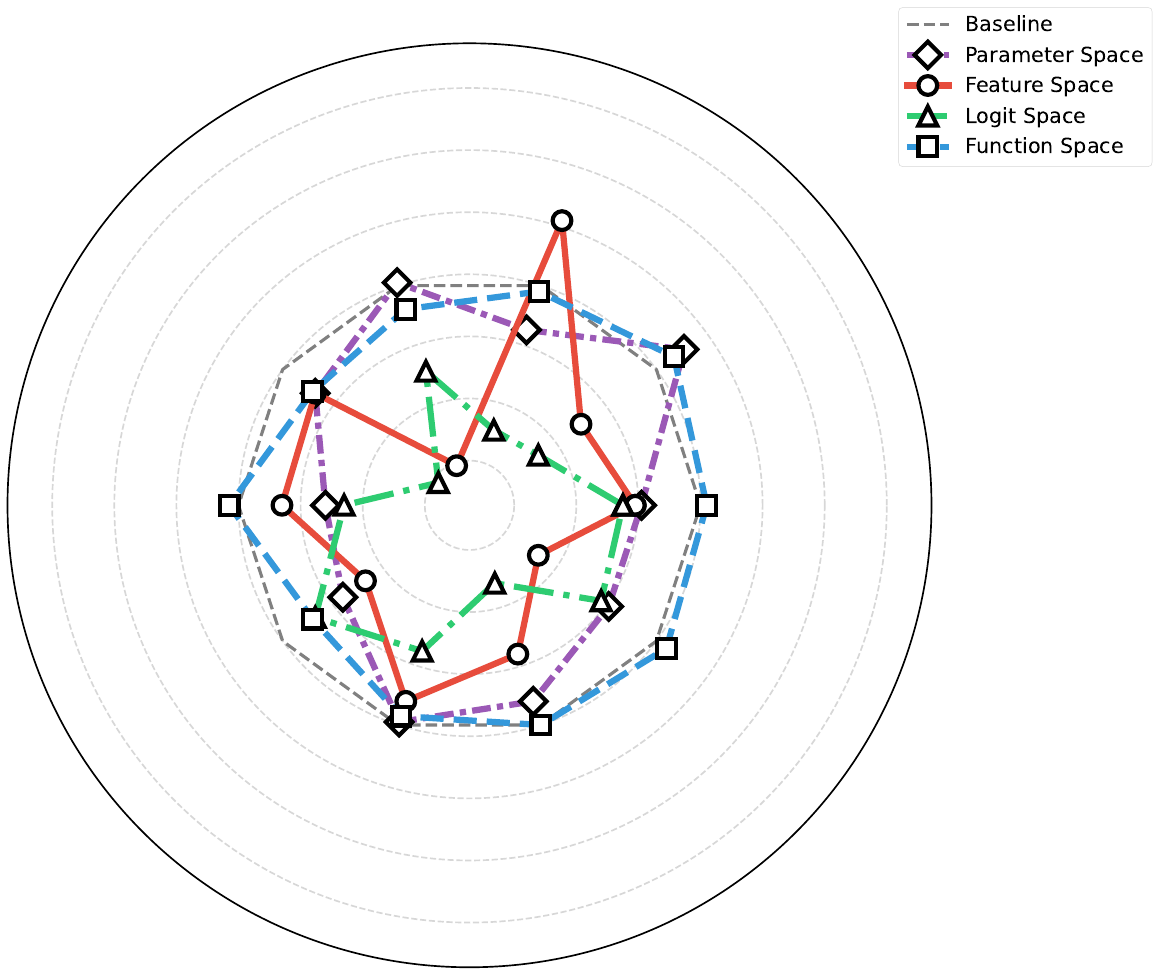}
		\subcaption{Effect of different perturbation magnitude on accuracy in ImageNetV2.}
		\label{fig:row1}
	\end{subfigure}
	\vspace{1ex}
	
	\begin{subfigure}[b]{\textwidth}
		\centering
		\includegraphics[width=0.15\textwidth]{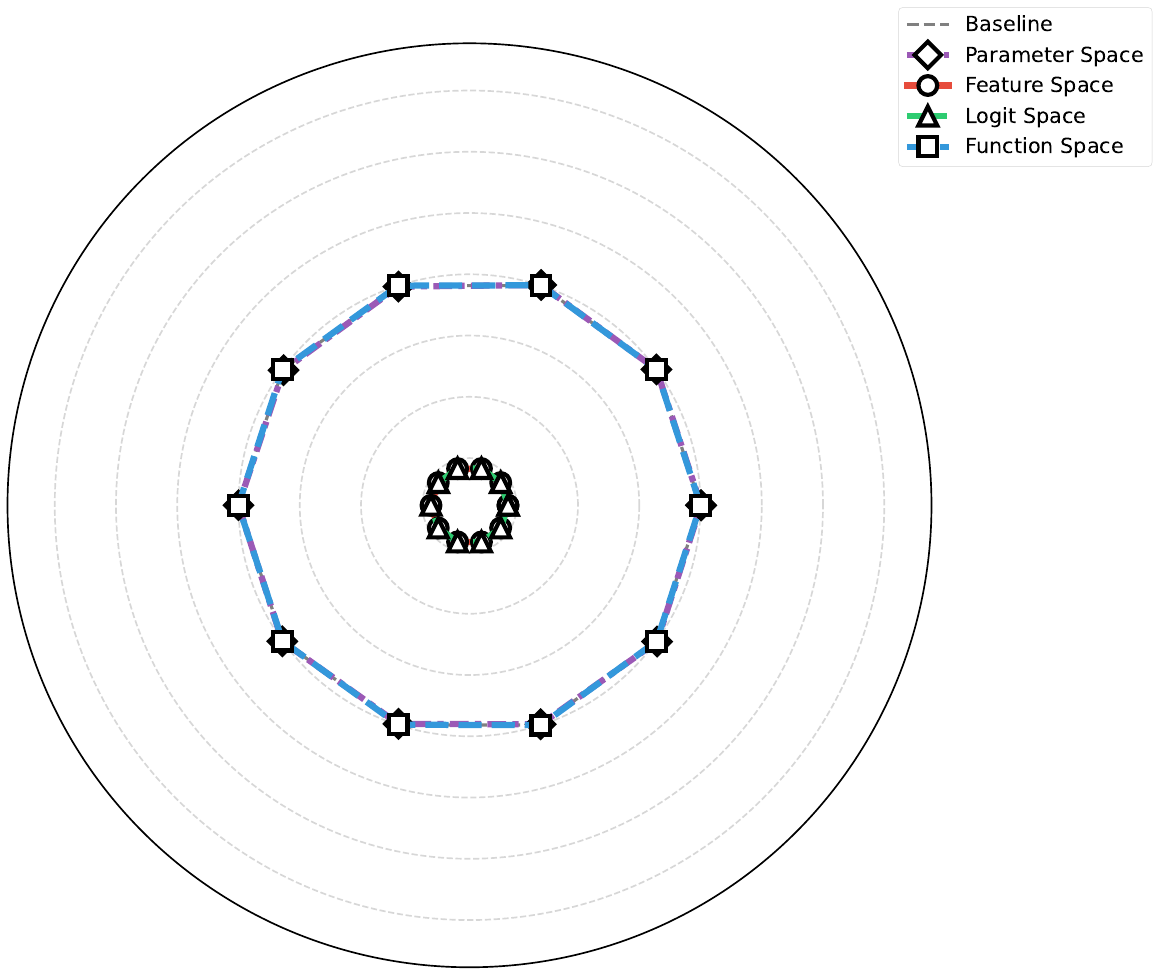}
		\includegraphics[width=0.15\textwidth]{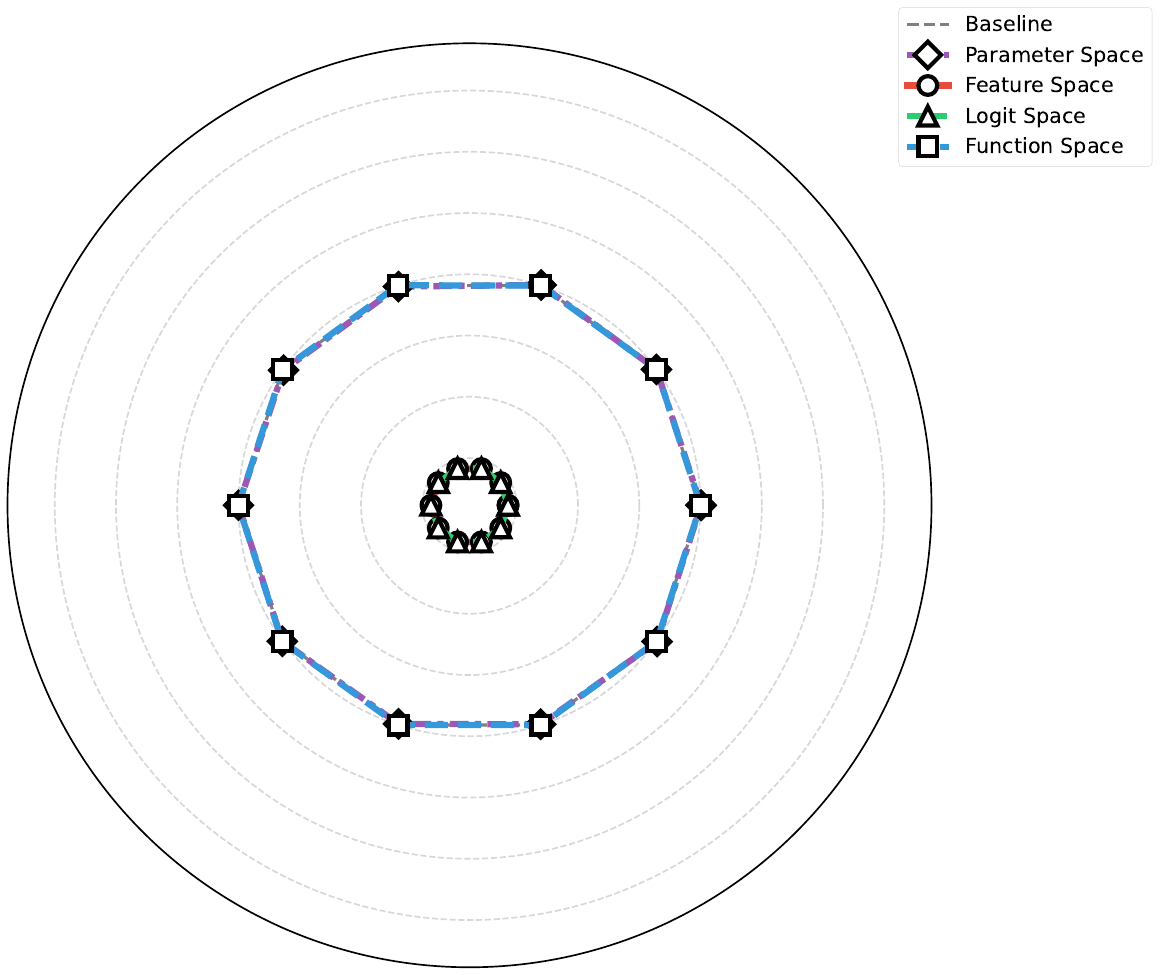}
		\includegraphics[width=0.15\textwidth]{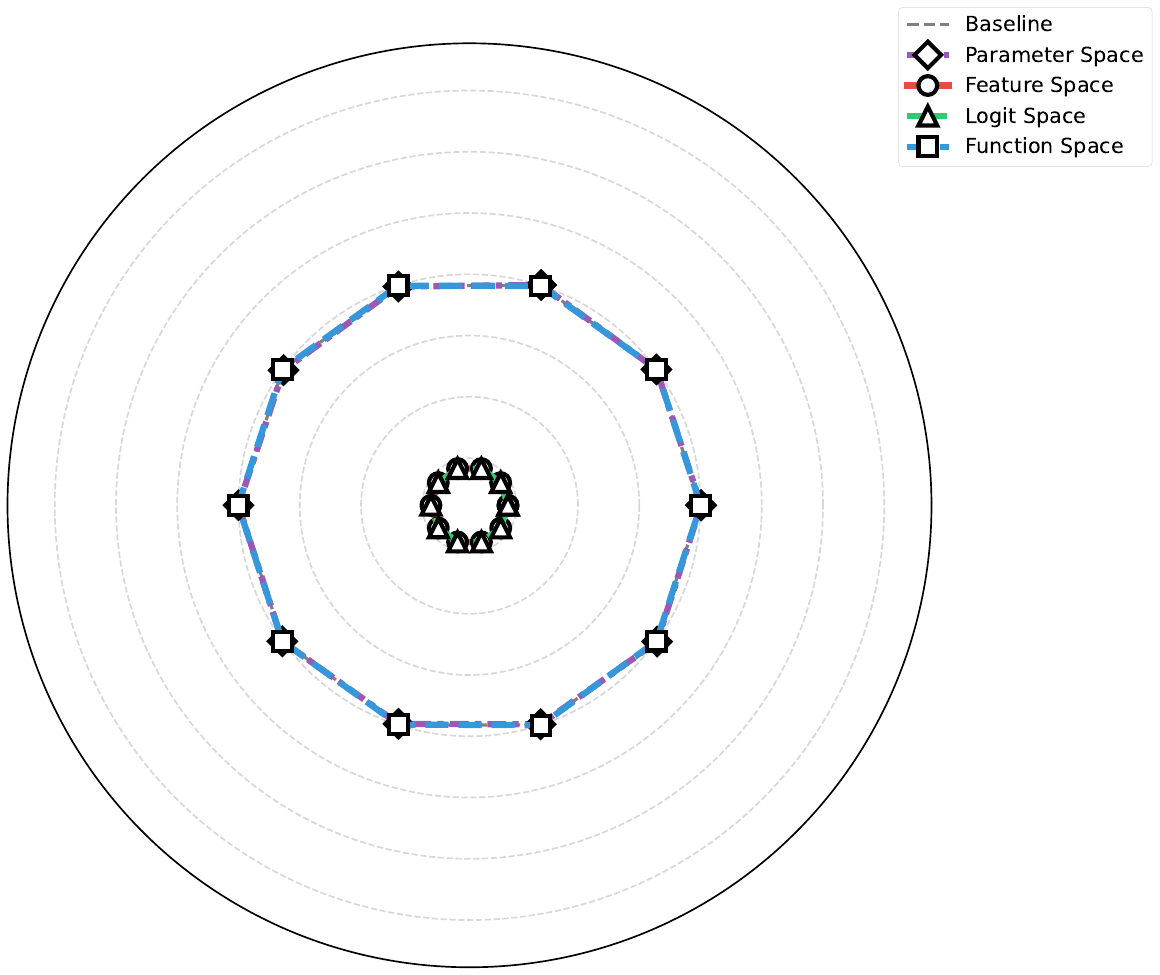}
		\includegraphics[width=0.15\textwidth]{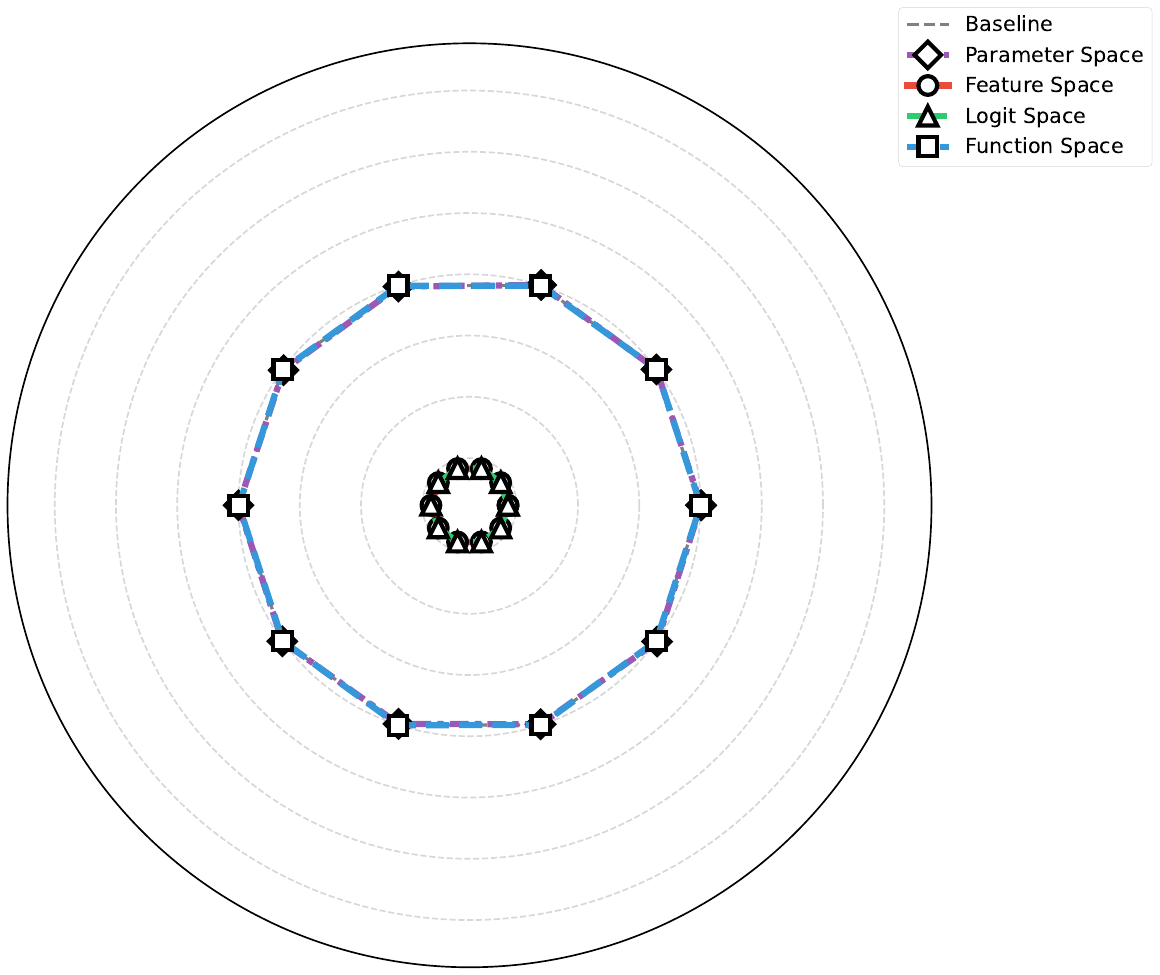}
		\includegraphics[width=0.15\textwidth]{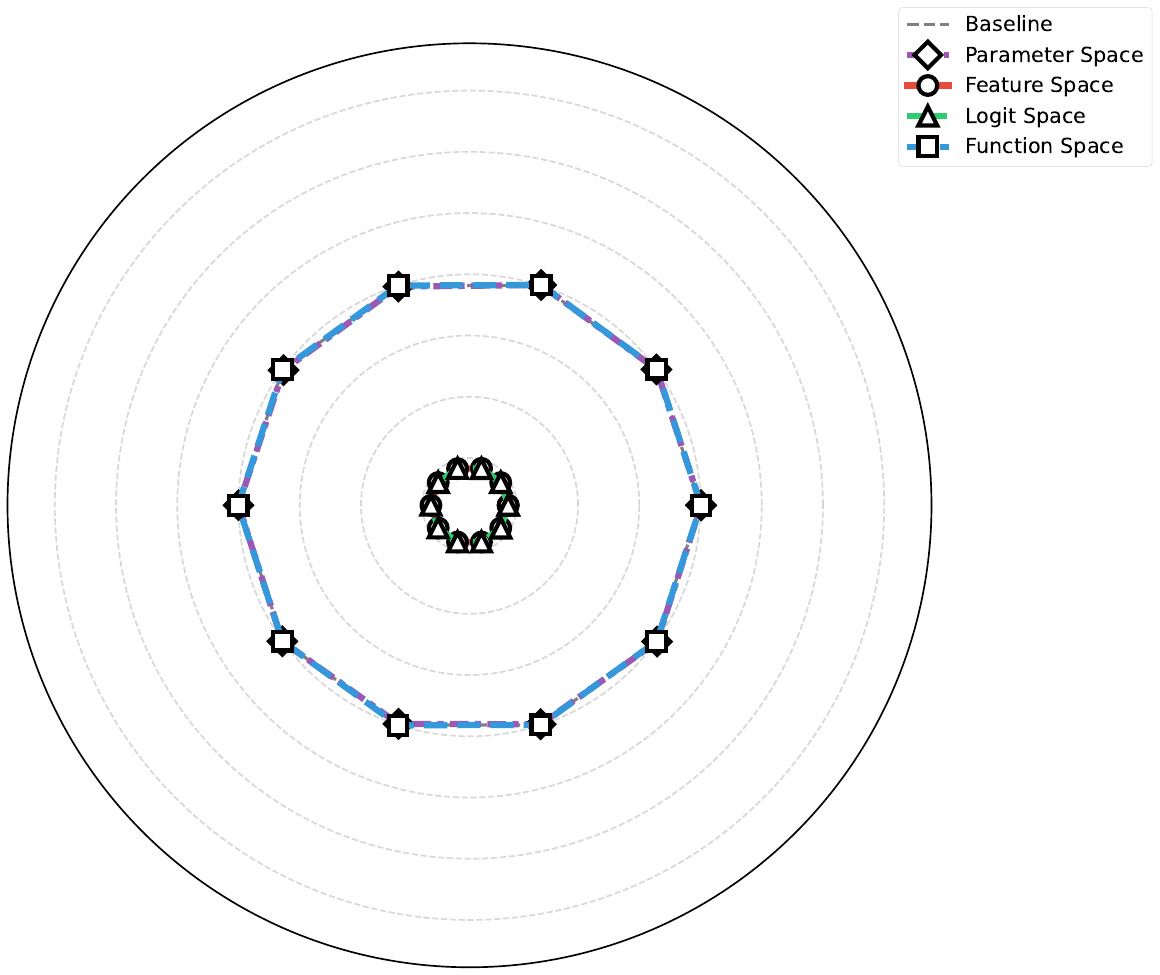}
		\includegraphics[width=0.15\textwidth]{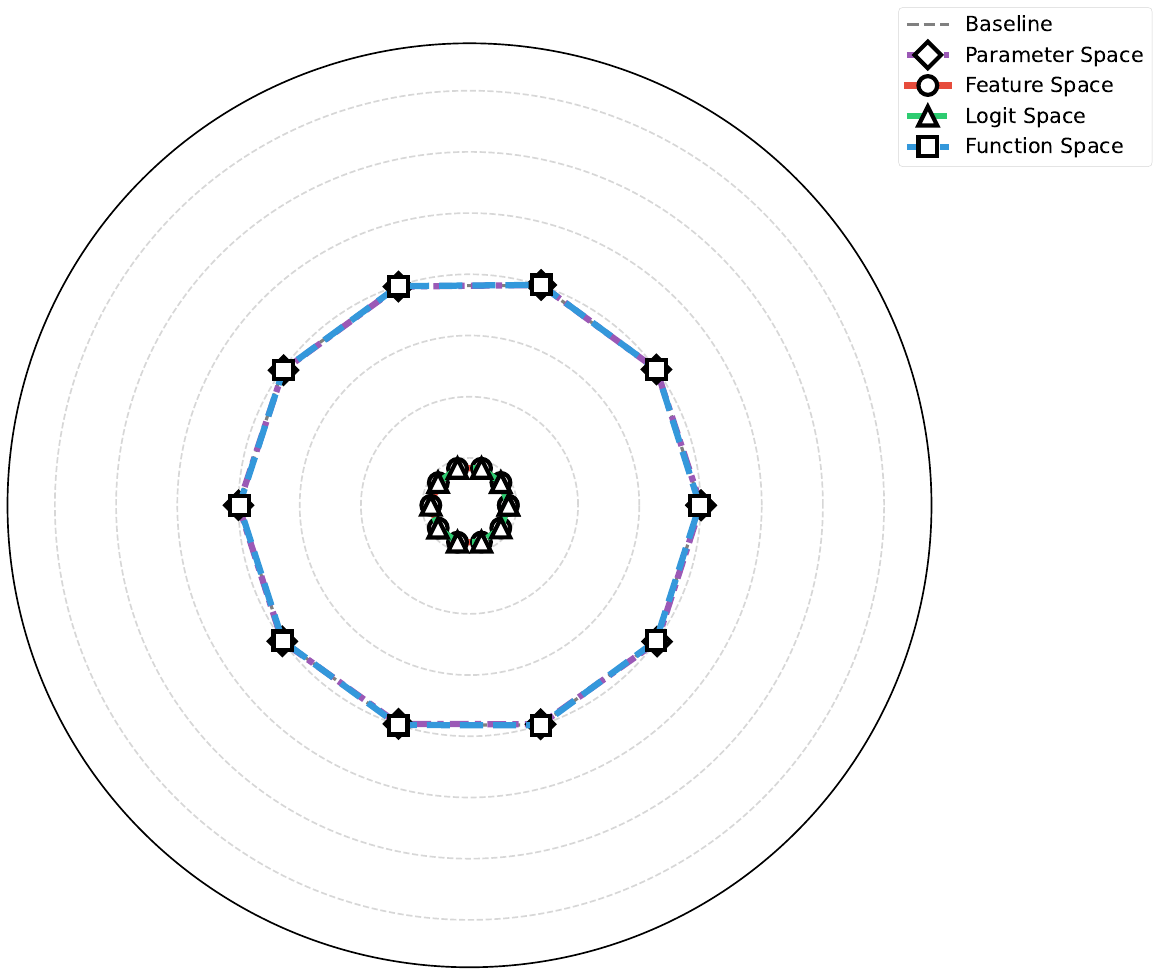}
		\subcaption{Effect of different perturbation magnitude on accuracy in ImageNetA.}
		\label{fig:row1}
	\end{subfigure}
	\vspace{1ex}
	
	\begin{subfigure}[b]{\textwidth}
		\centering
		\includegraphics[width=0.15\textwidth]{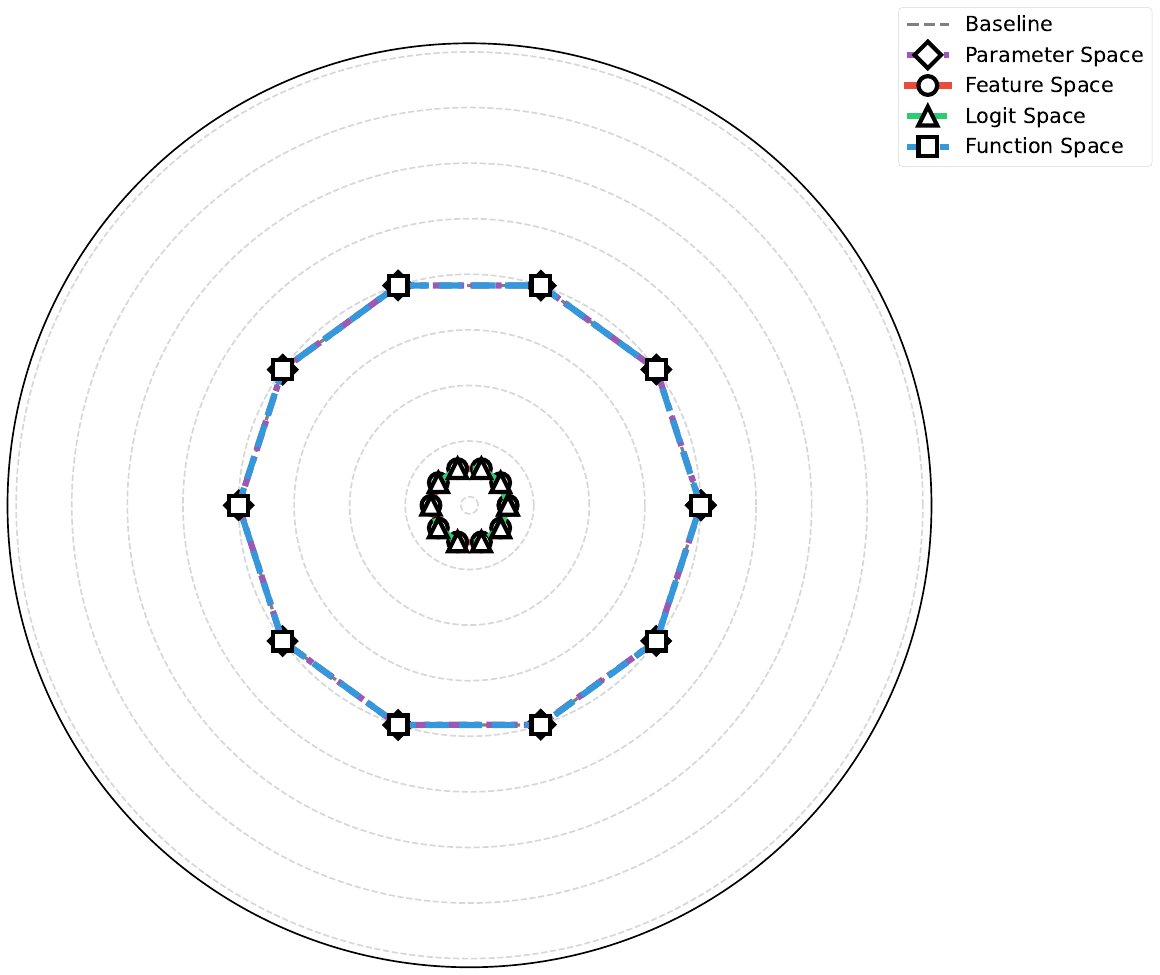}
		\includegraphics[width=0.15\textwidth]{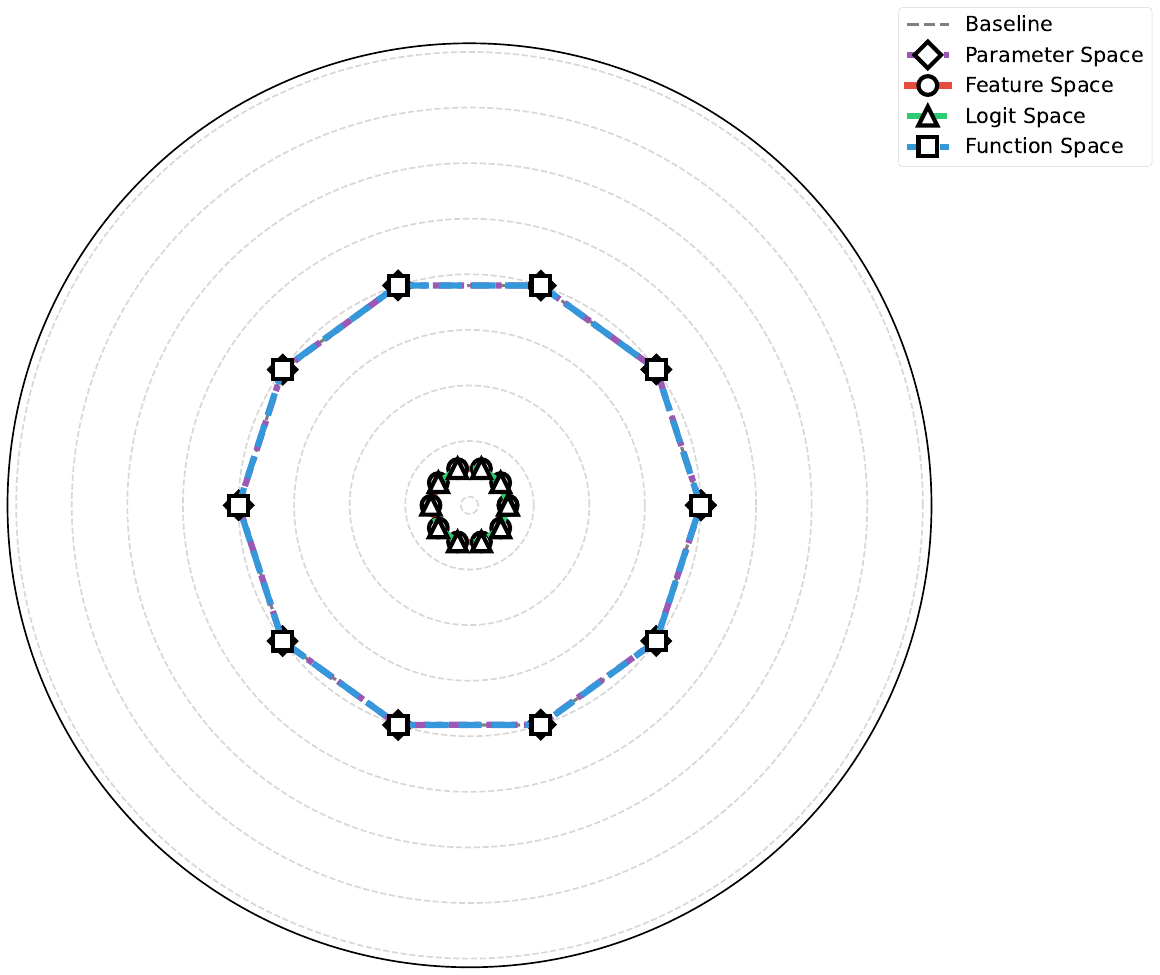}
		\includegraphics[width=0.15\textwidth]{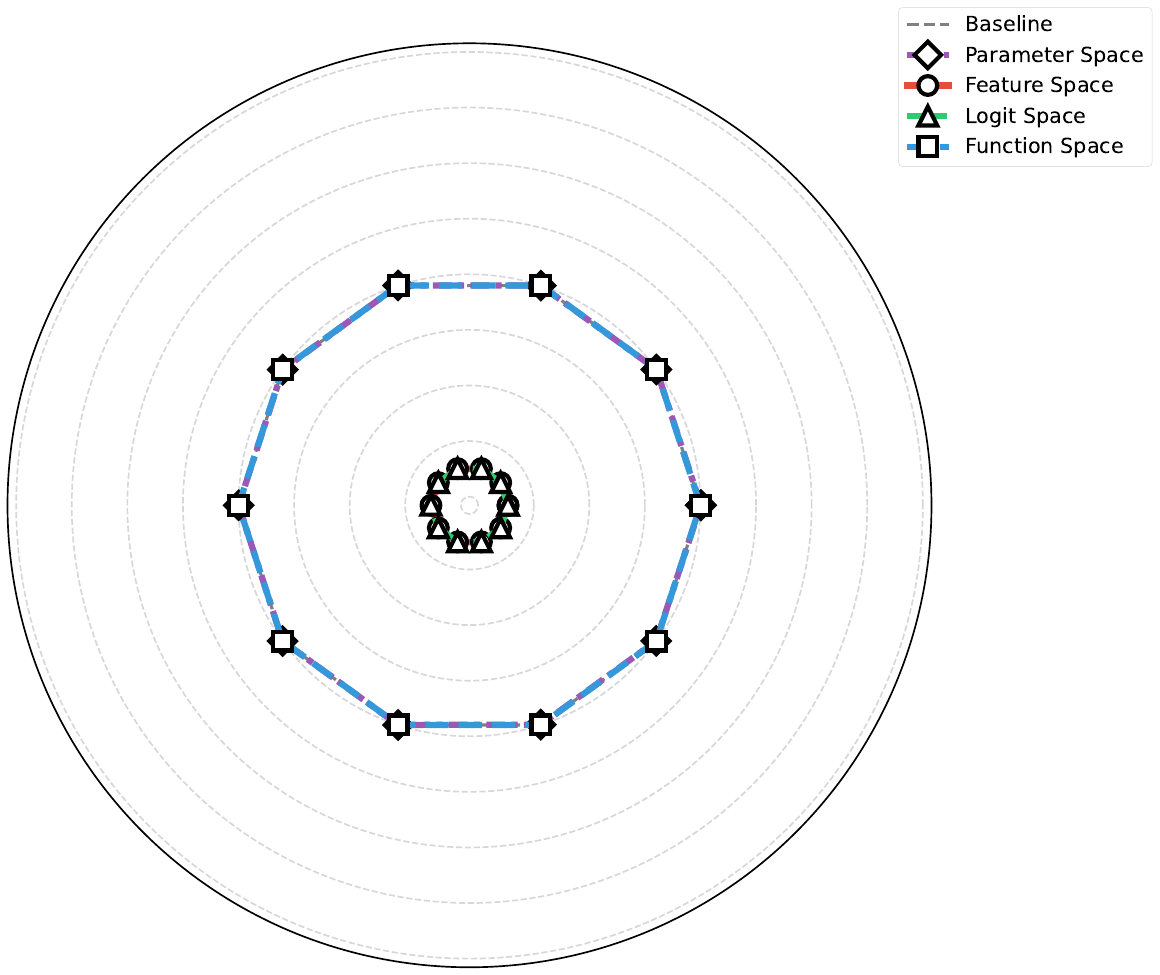}
		\includegraphics[width=0.15\textwidth]{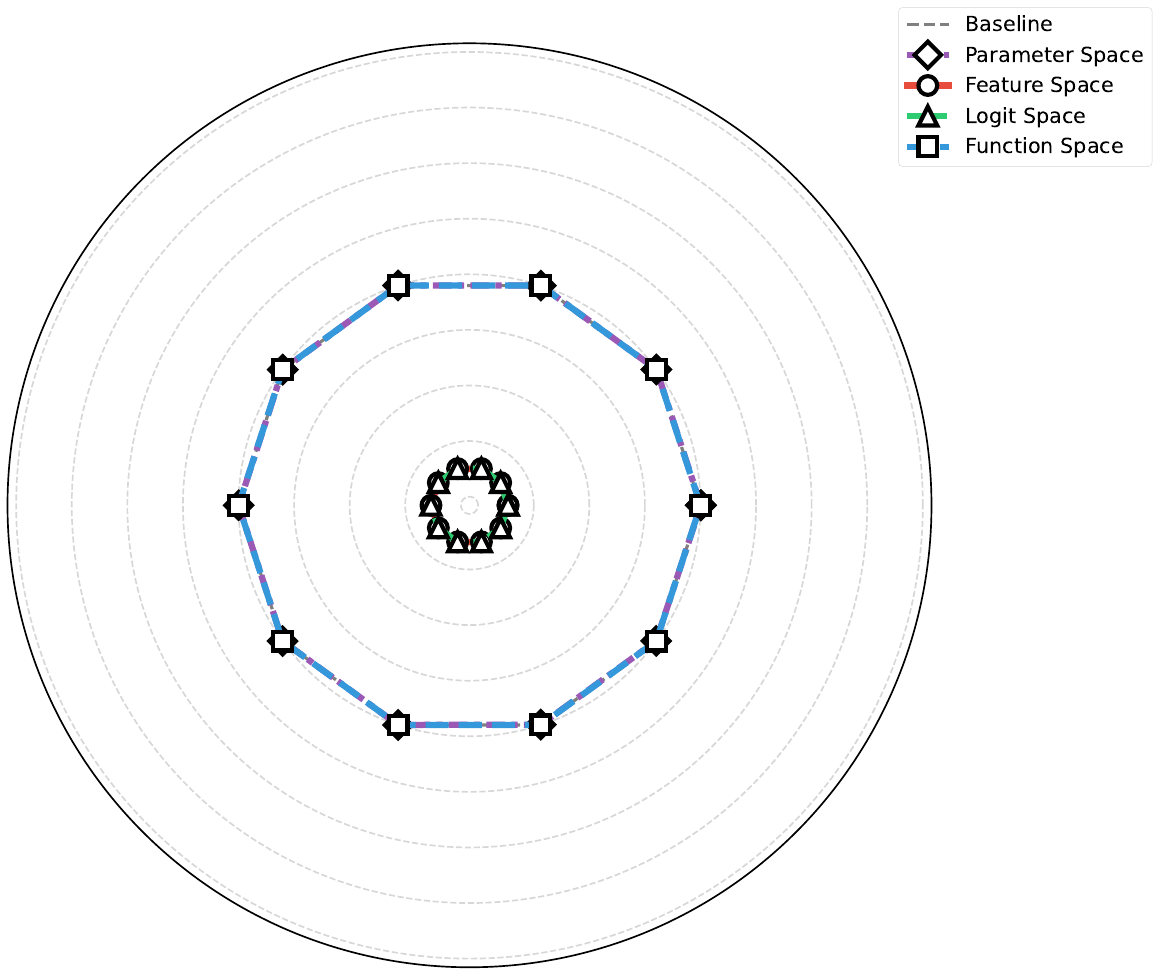}
		\includegraphics[width=0.15\textwidth]{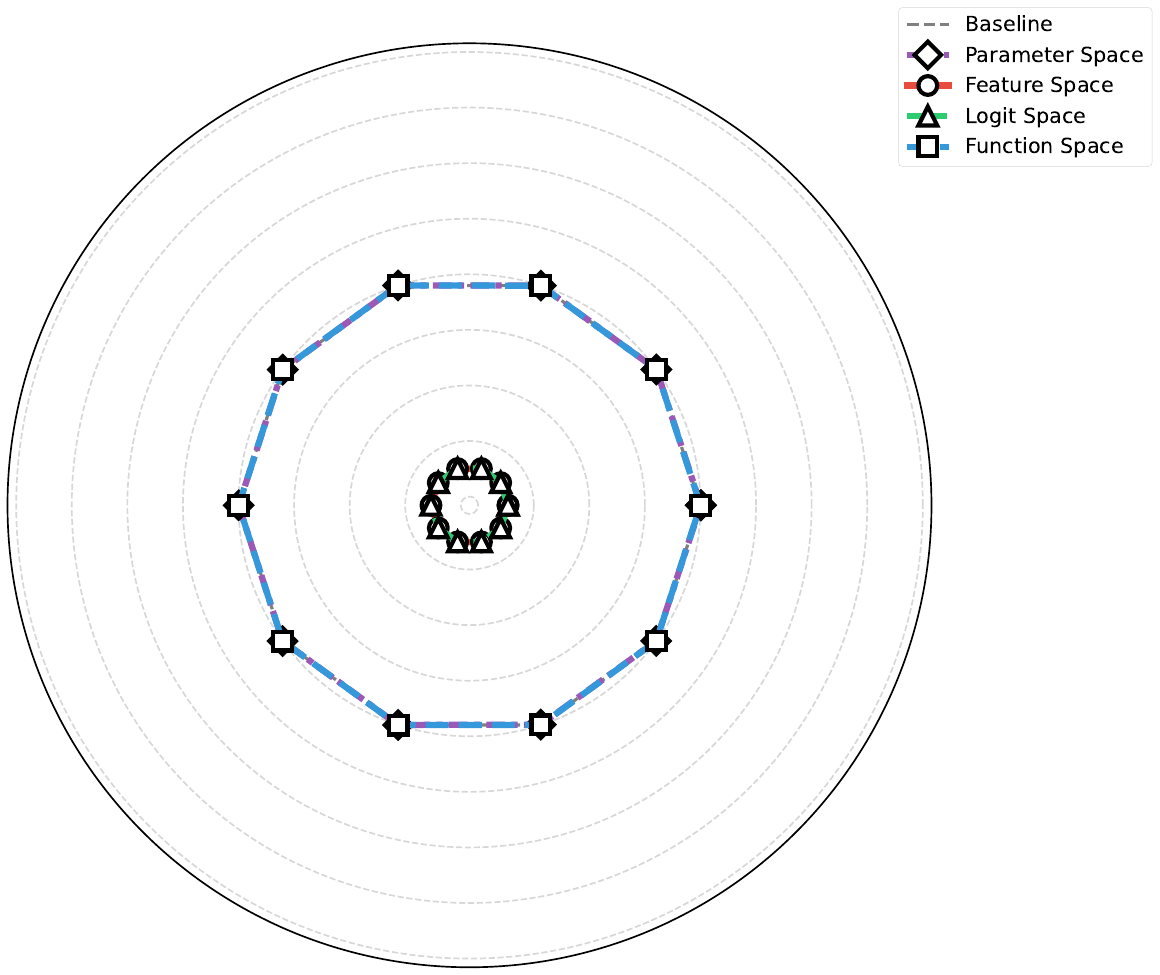}
		\includegraphics[width=0.15\textwidth]{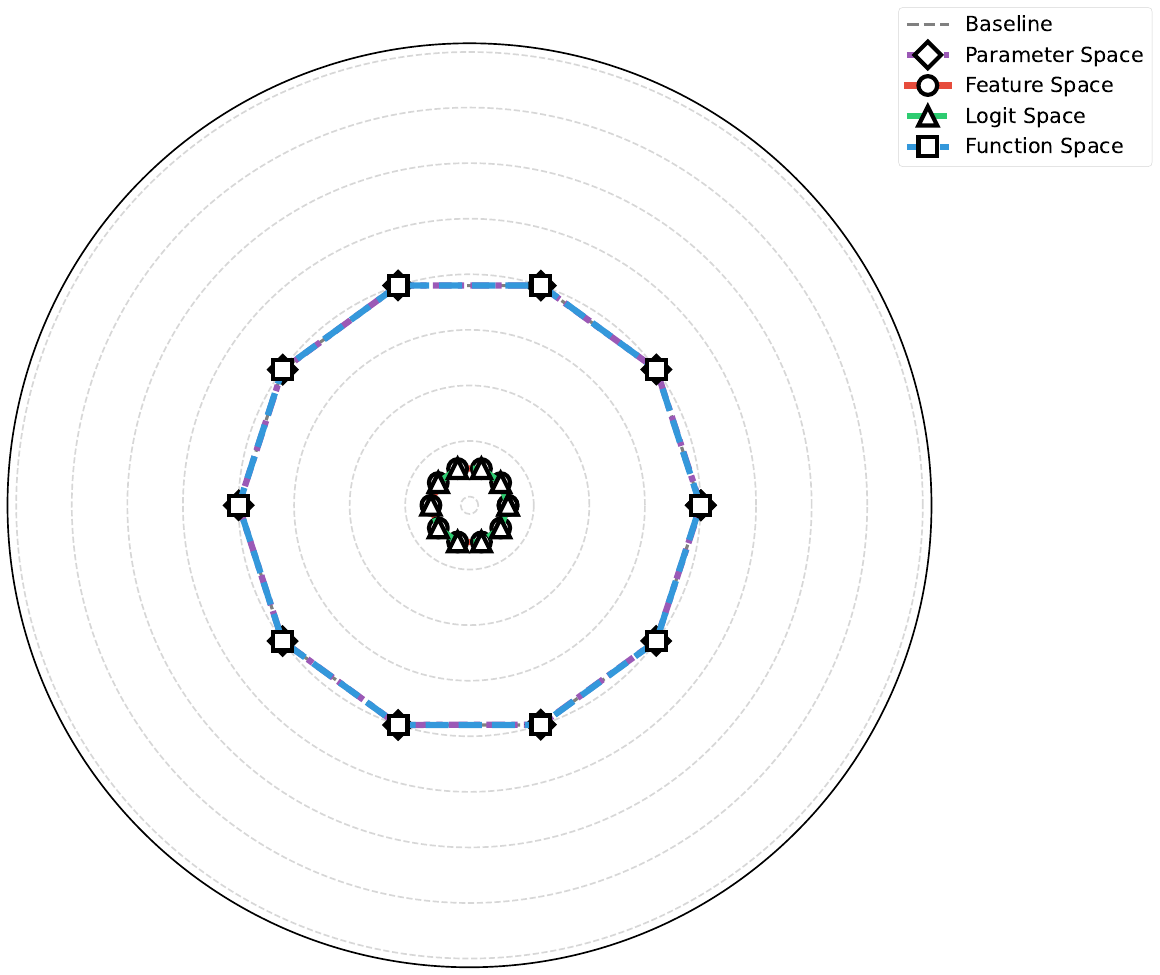}
		\subcaption{Effect of different perturbation magnitude on accuracy in ImageNetR.}
		\label{fig:row1}
	\end{subfigure}
	\vspace{1ex}
	
	\begin{subfigure}[b]{\textwidth}
		\centering
		\includegraphics[width=0.15\textwidth]{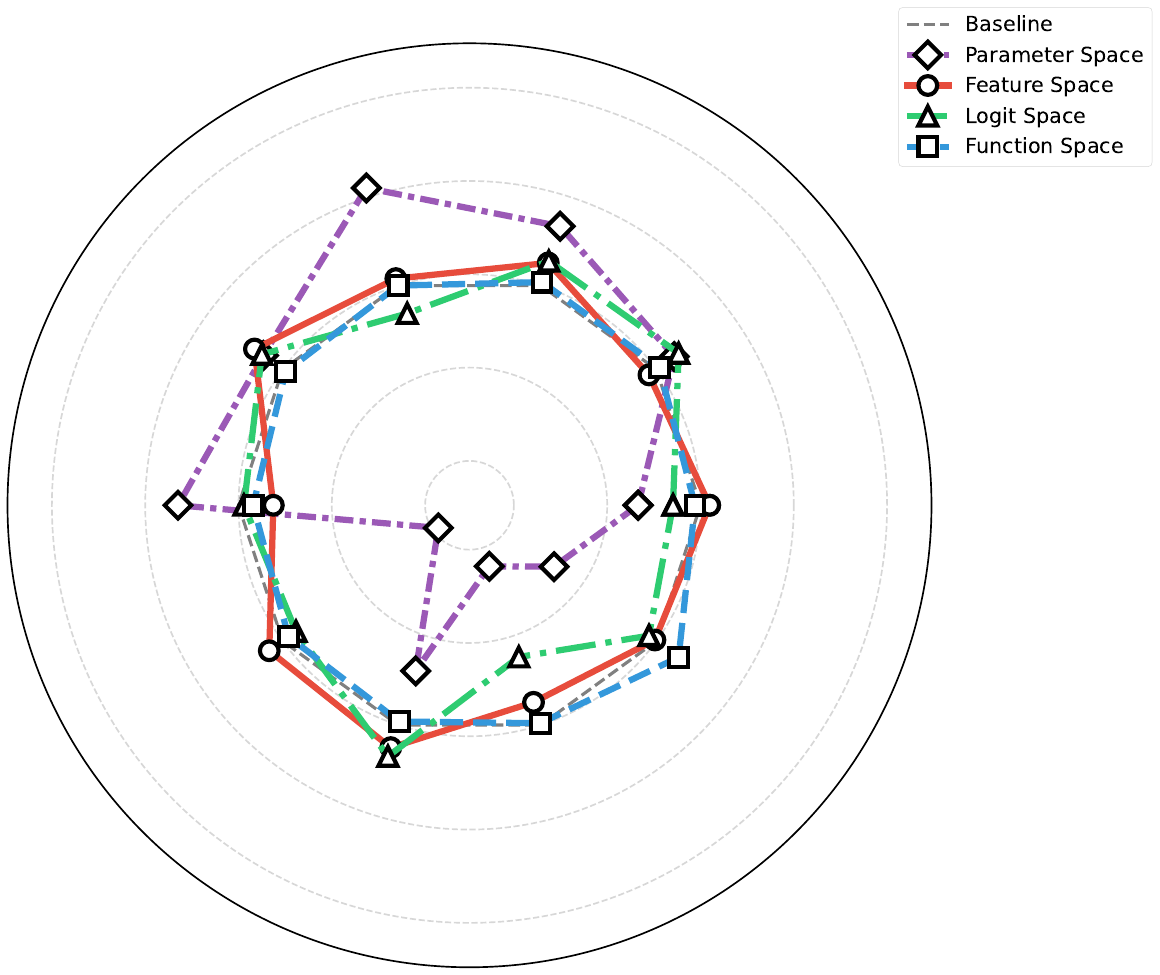}
		\includegraphics[width=0.15\textwidth]{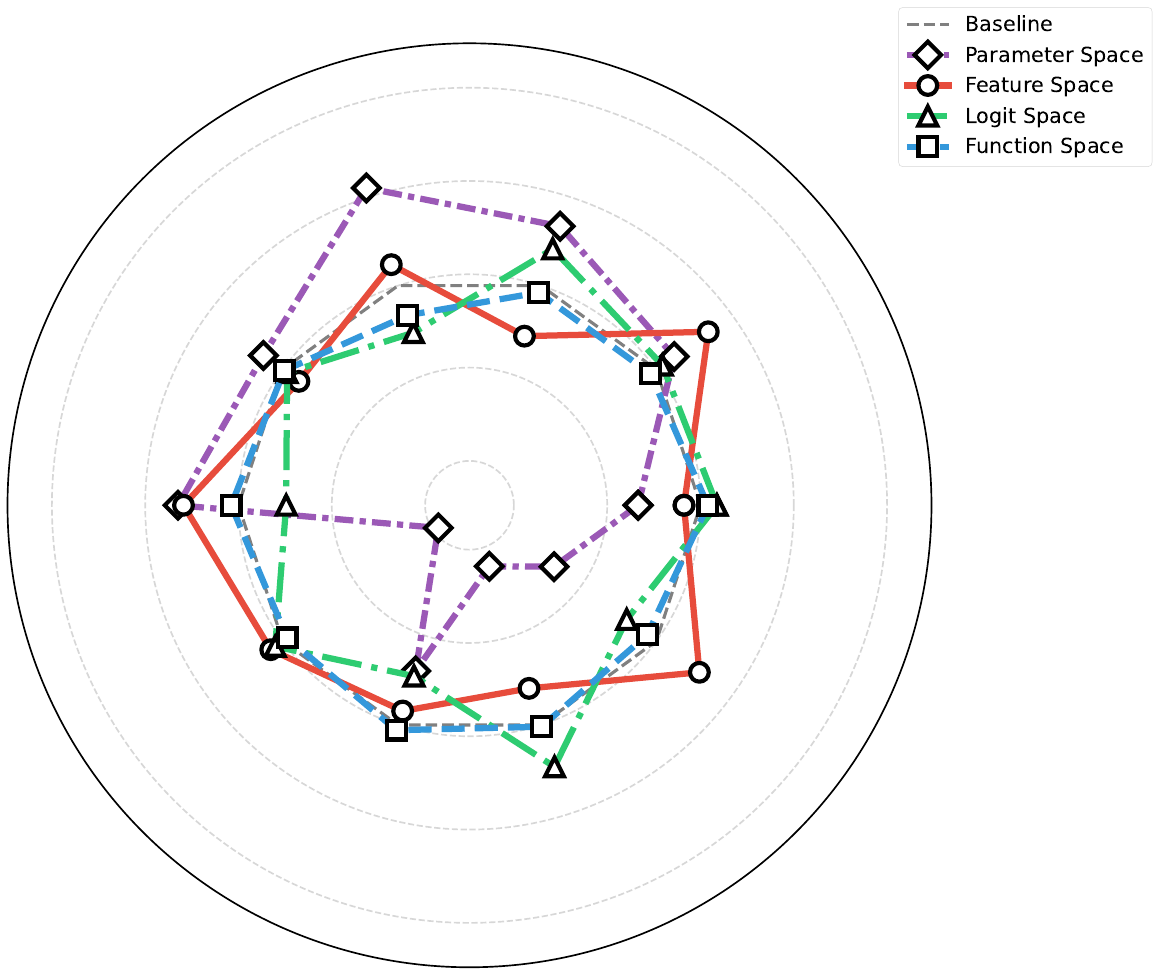}
		\includegraphics[width=0.15\textwidth]{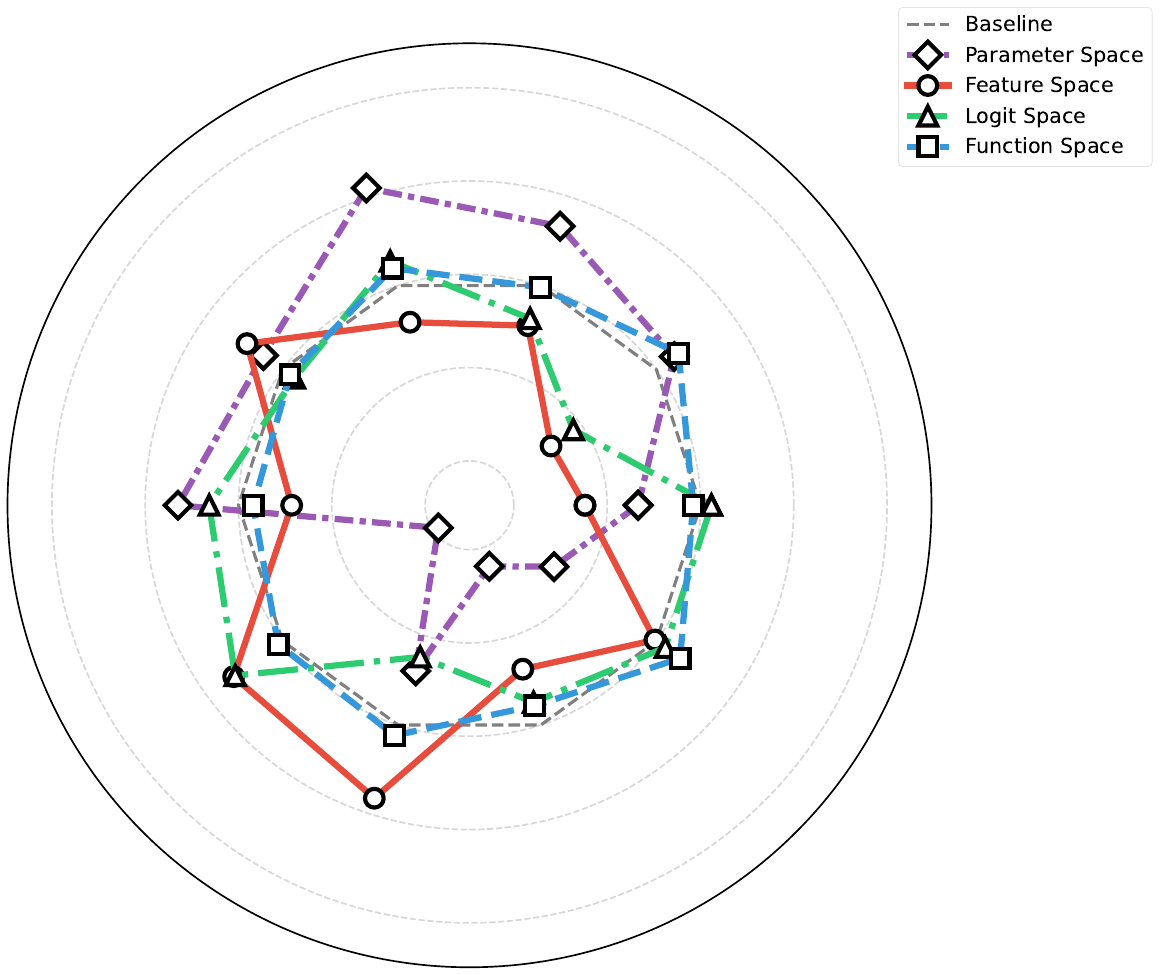}
		\includegraphics[width=0.15\textwidth]{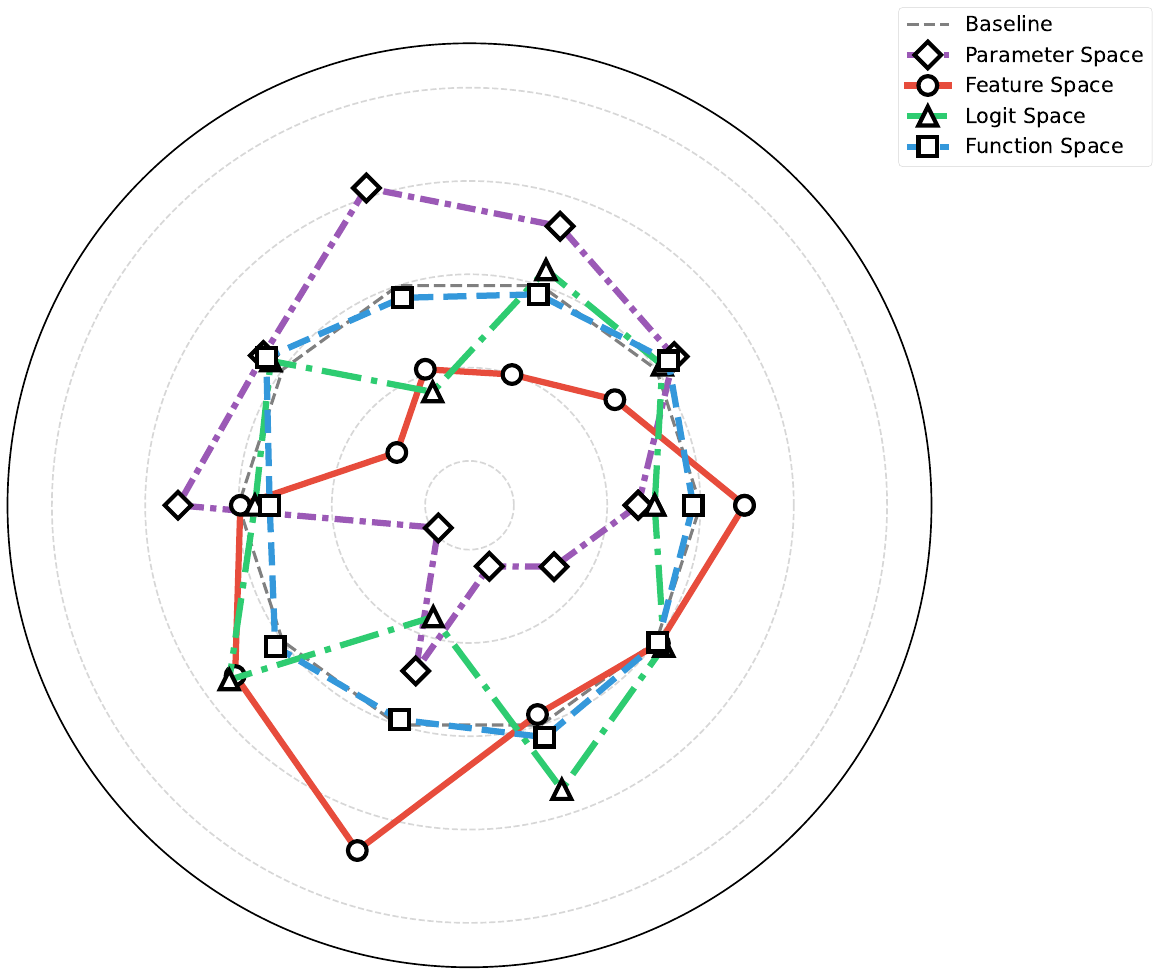}
		\includegraphics[width=0.15\textwidth]{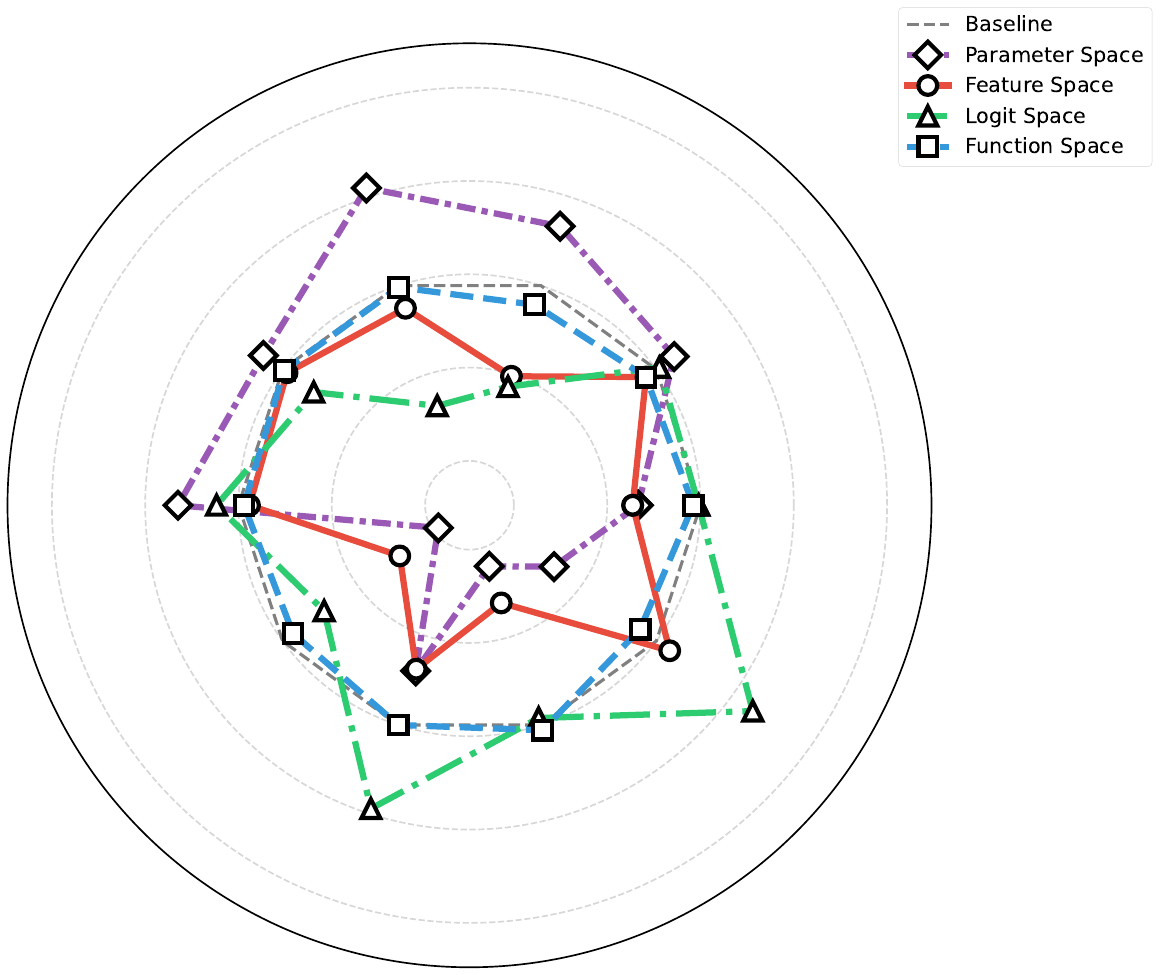}
		\includegraphics[width=0.15\textwidth]{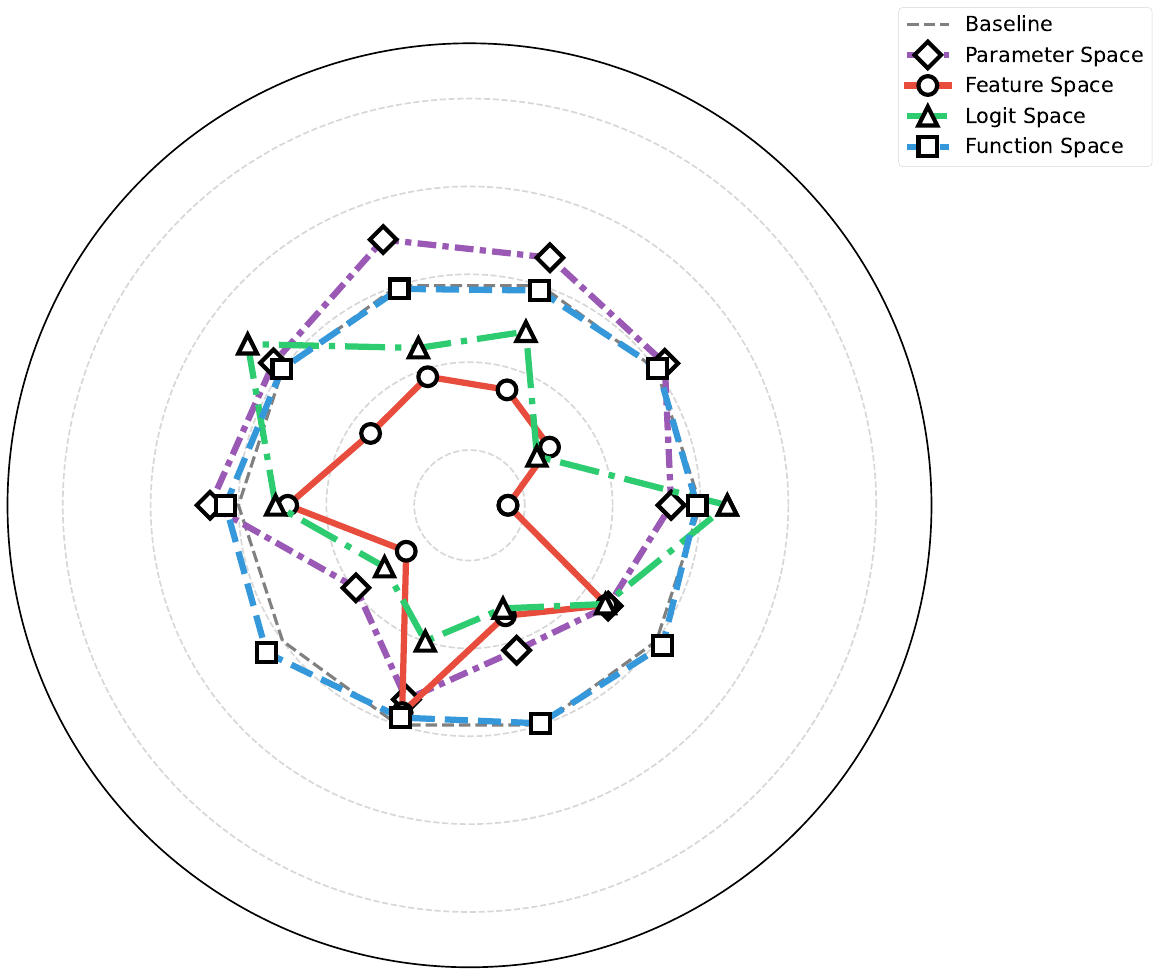}
		\subcaption{Effect of different perturbation magnitude on accuracy in ImageNetSketch.}
		\label{fig:row1}
	\end{subfigure}
	\vspace{1ex}
	
	\caption{Variation in accuracy across different perturbation magnitudes in parameter, feature, logit, and function spaces, evaluated on six benchmarks: ImageNet, ImageNetV2, ImageNet-A, ImageNet-R, ImageNet-Sketch. Function space perturbations induce the best accuracy across different cases, demonstrating superior robustness across datasets and perturbation magnitudes.(perturbations magnitude is set to $m=0.0004$ in parameter space and \(m\in\{0.1,\,0.2,\,0.3,\,0.4,\,0.5,\,1.0\} \), as illustrated from left to right in the figure, for feature, logit, and function spaces.)}
	\label{fig:accuracy_space_comparison}
\end{figure}

\begin{figure}[htbp]
	\centering
	\begin{subfigure}[b]{\textwidth}
		\centering
		\includegraphics[width=0.3\textwidth]{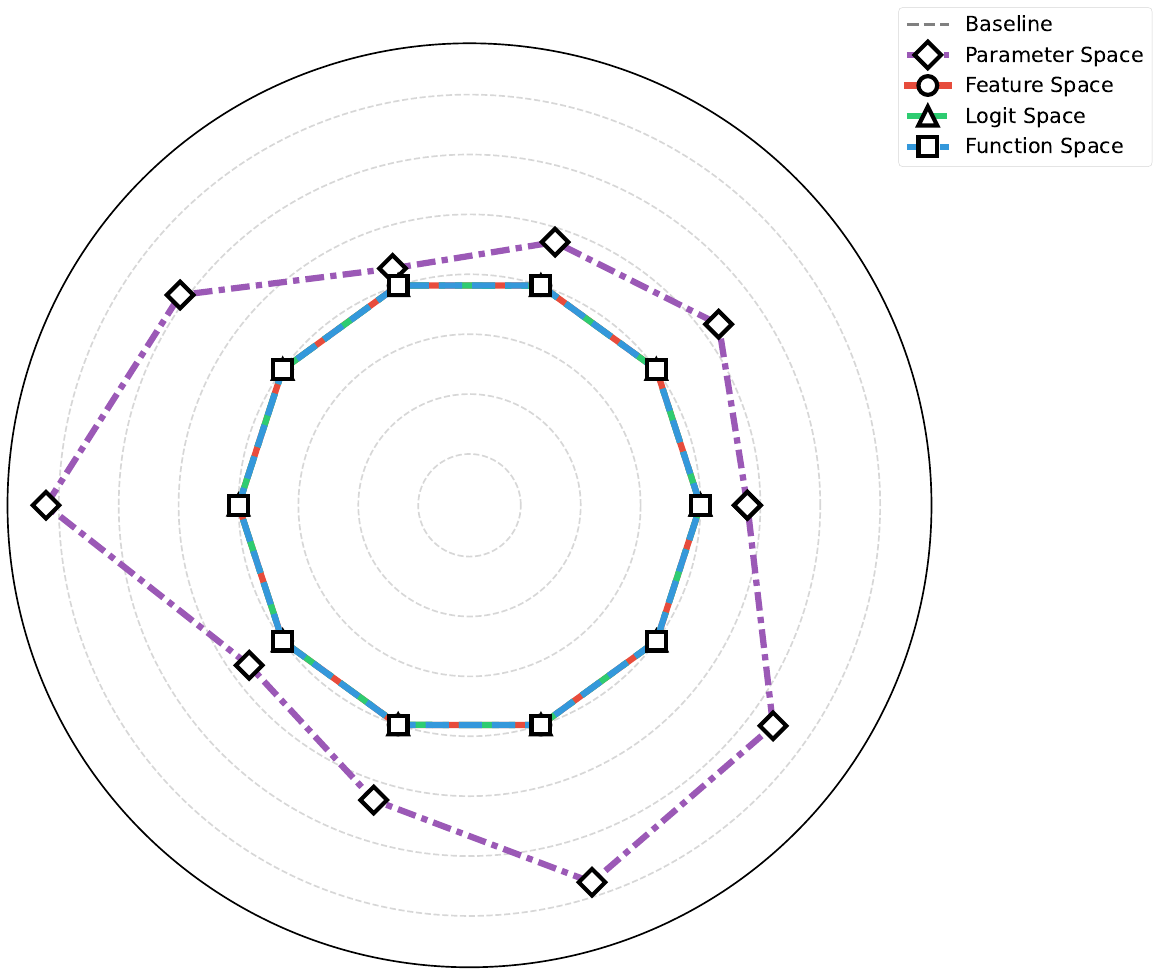}
		\includegraphics[width=0.3\textwidth]{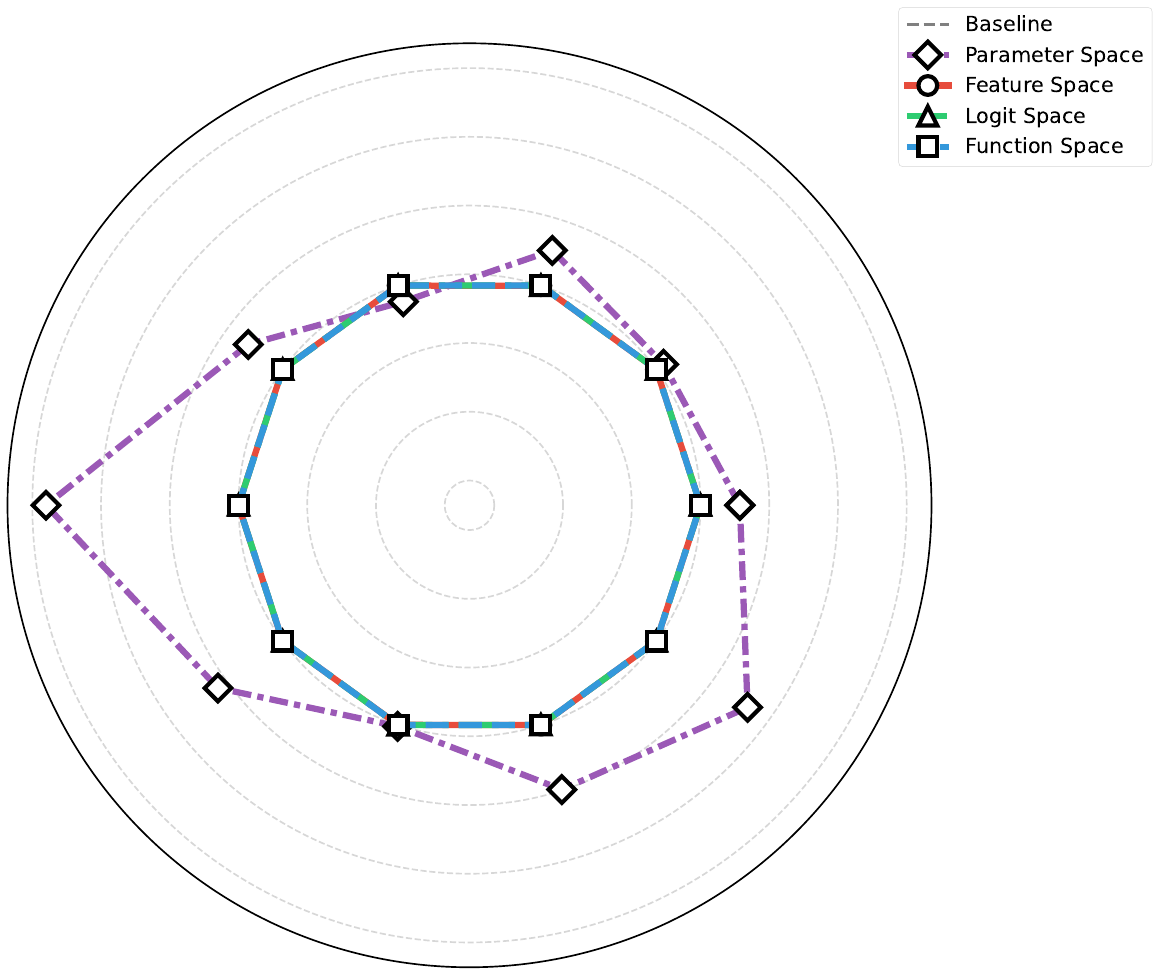}
		\includegraphics[width=0.3\textwidth]{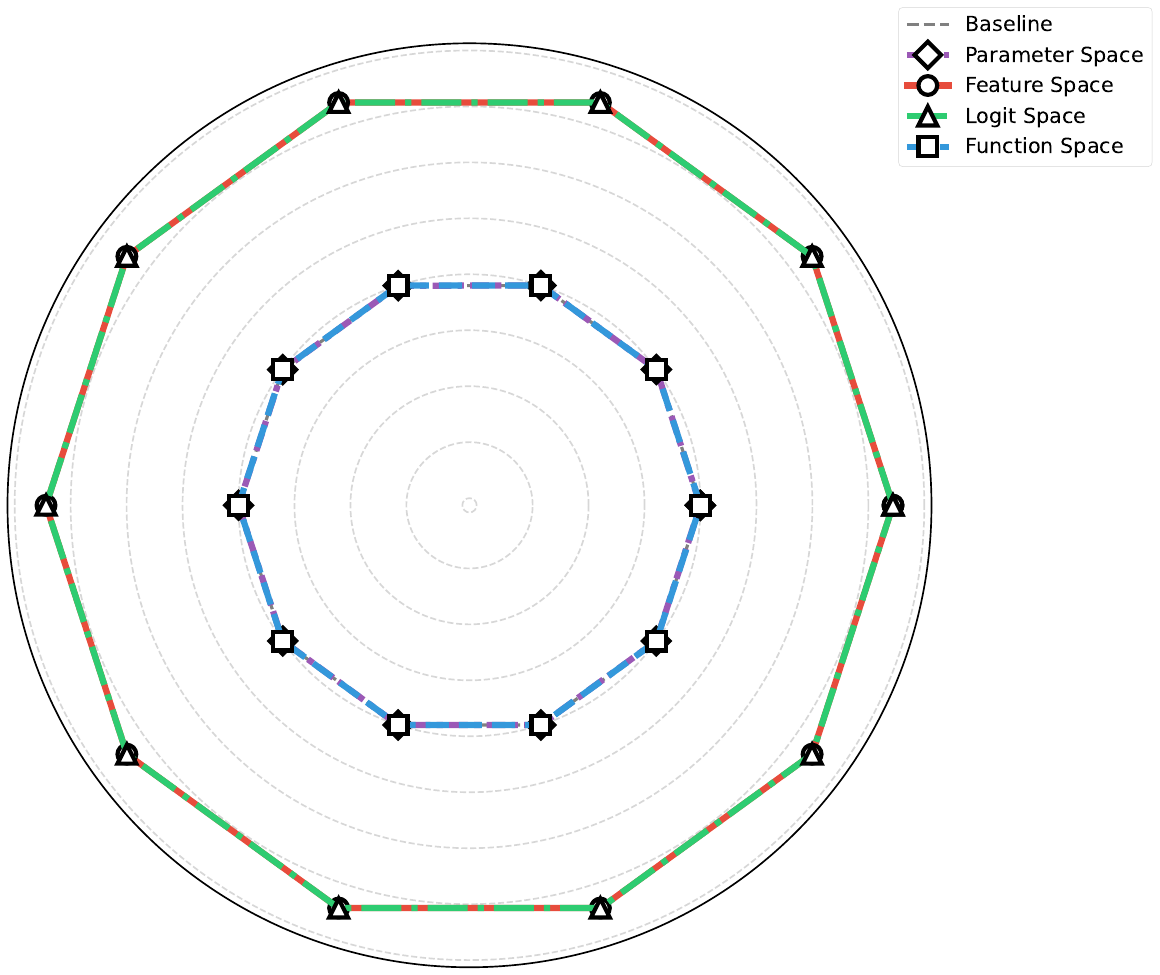}
	\end{subfigure}
	\vspace{1ex}	
	
	\begin{subfigure}[b]{\textwidth}
		\centering
		\includegraphics[width=0.3\textwidth]{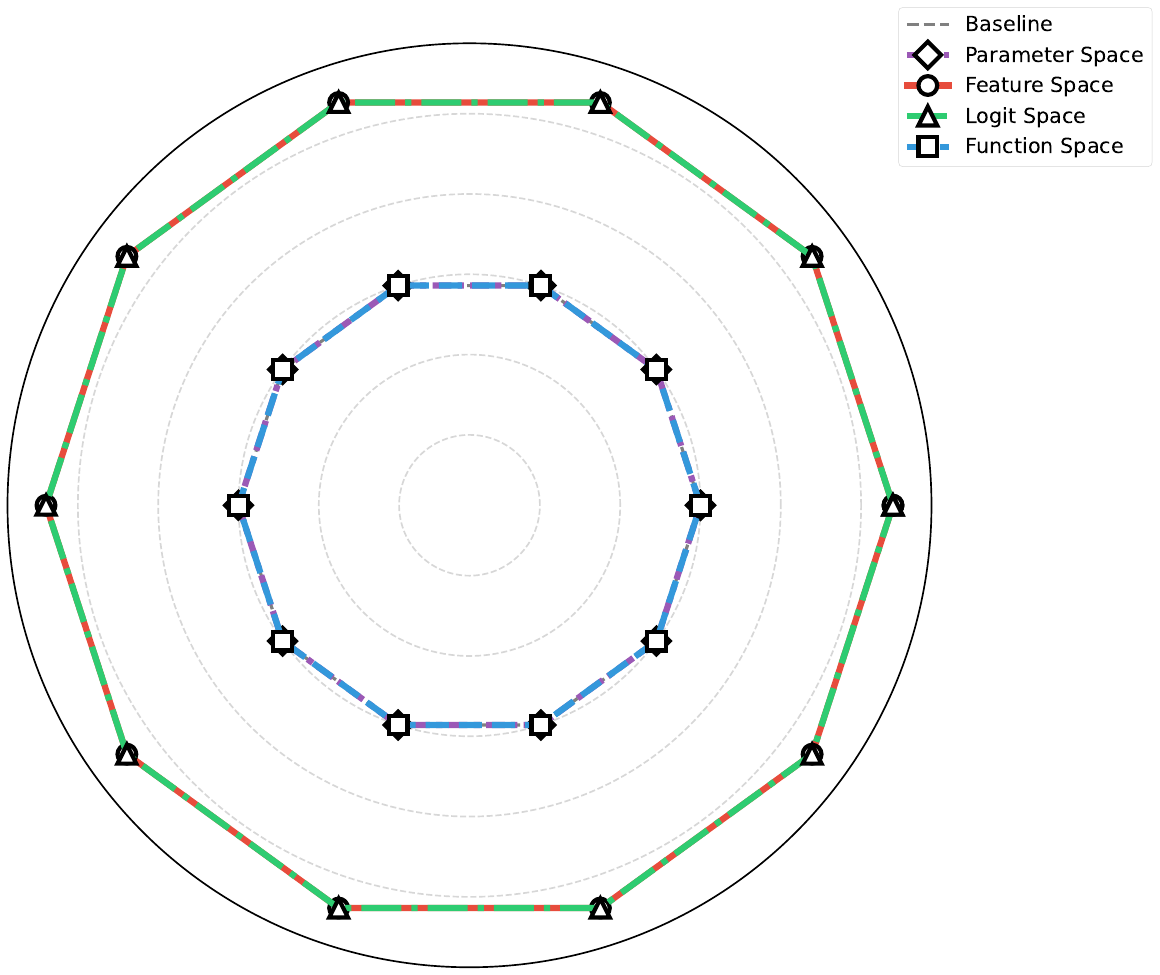}
		\includegraphics[width=0.3\textwidth]{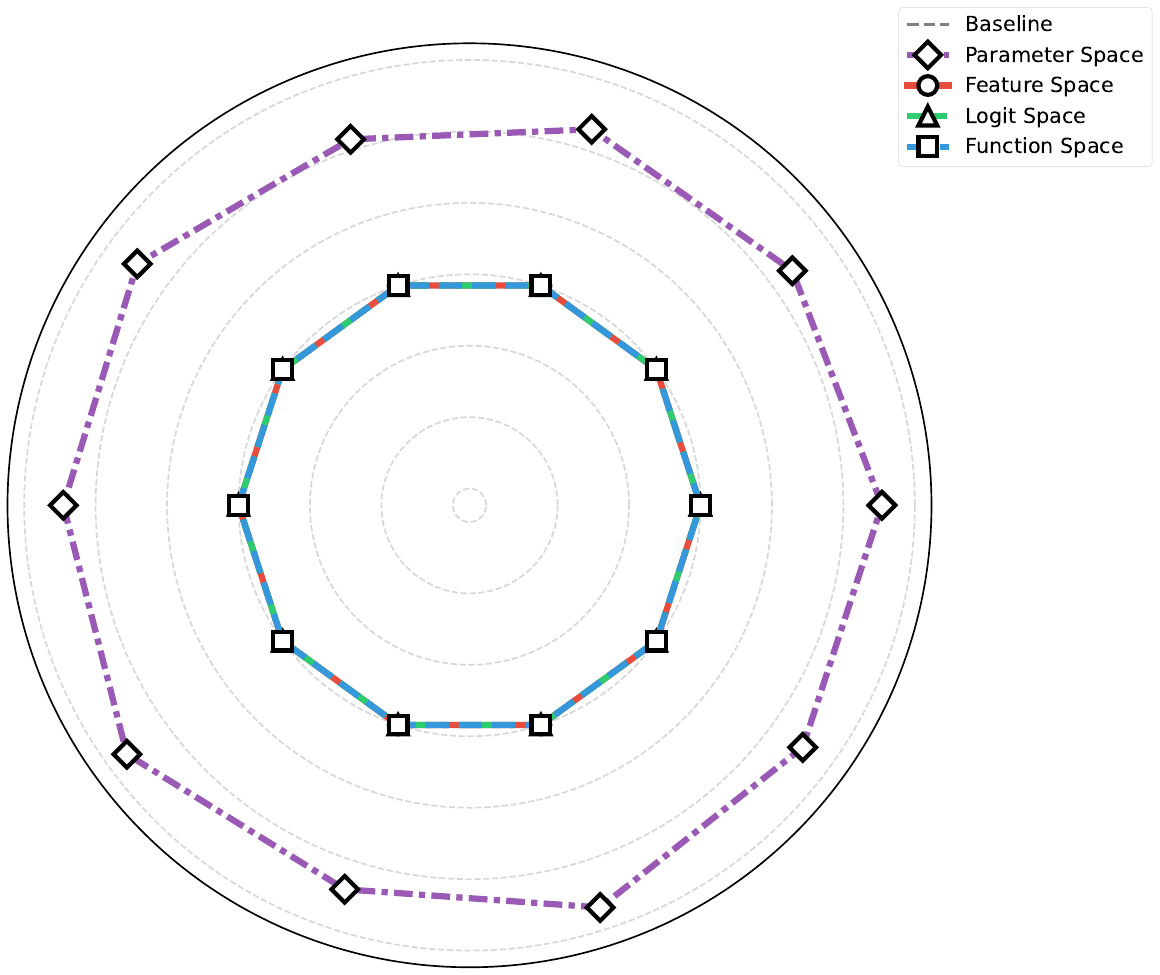}
		\label{fig:row1}
	\end{subfigure}
	\vspace{1ex}	
	
	\caption{Impact of fixed perturbations magnitude, $m=0.0004$ in parameter space, feature, logit, and function spaces, on the loss across five benchmarks (ImageNet, ImageNetV2, ImageNet-A, ImageNet-R, and ImageNet-Sketch). Function space perturbations induce the smallest loss across different cases, demonstrating superior robustness across datasets.}
	\label{fig:normalized_perturbation_loss}
\end{figure}

\begin{figure}[htbp]
	\centering
	\begin{subfigure}[b]{\textwidth}
		\centering
		\includegraphics[width=0.3\textwidth]{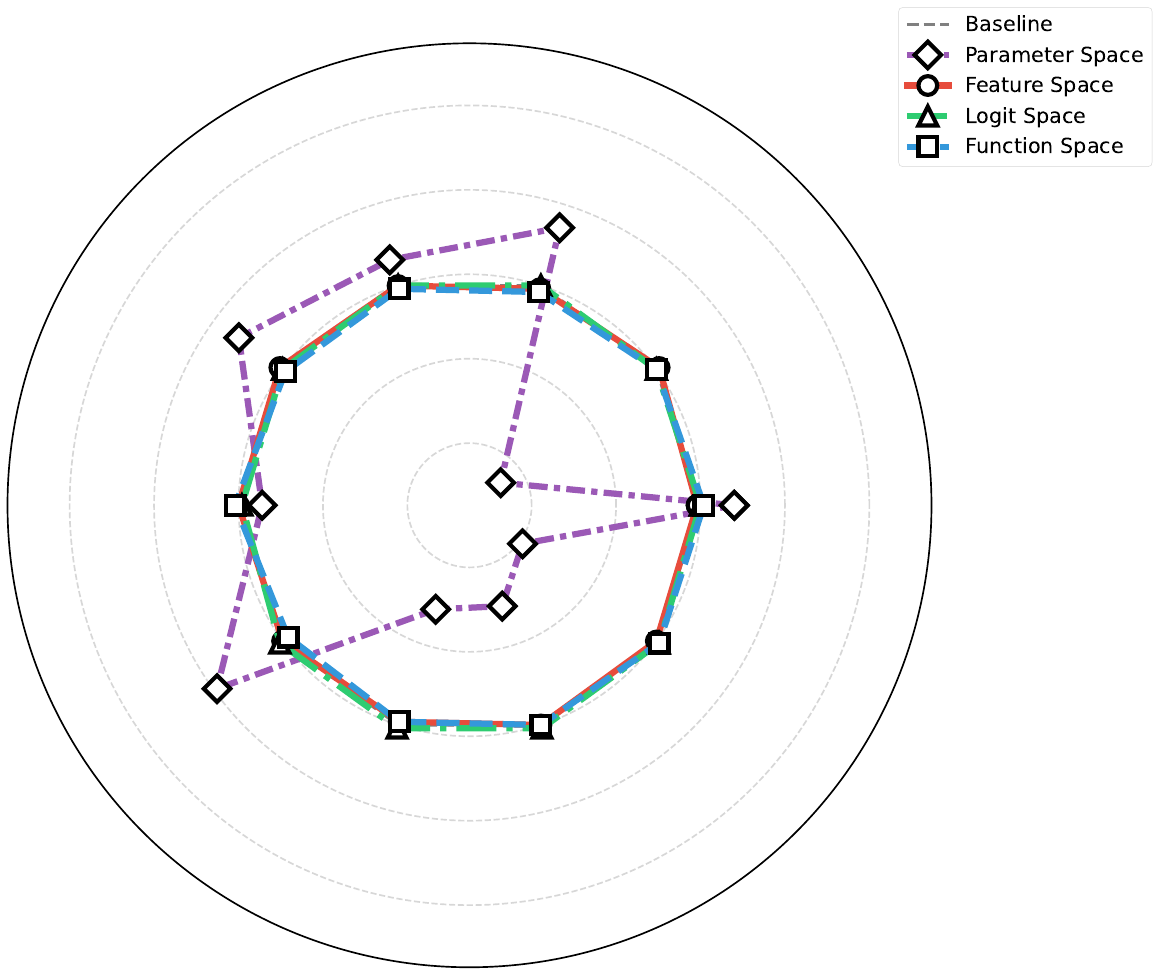}
		\includegraphics[width=0.3\textwidth]{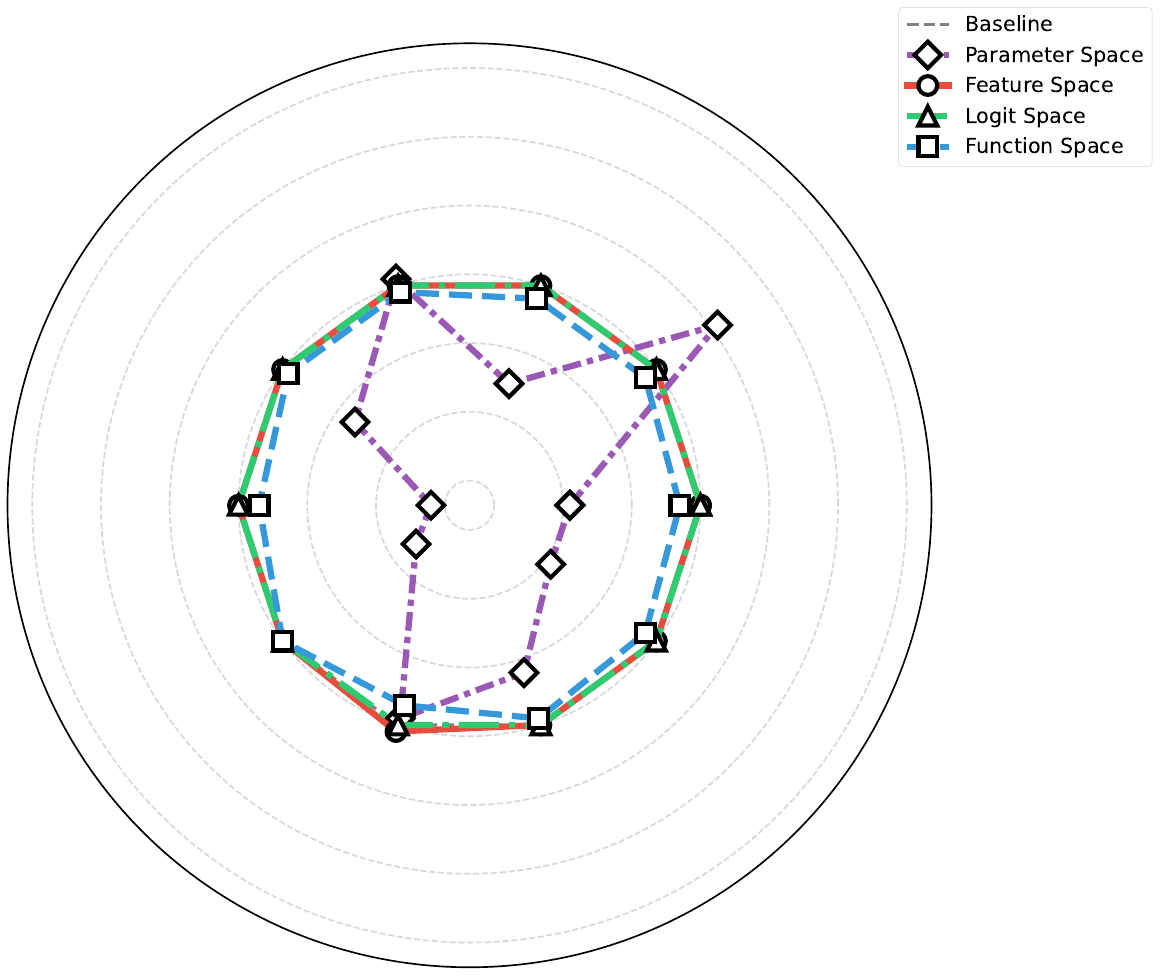}
		\includegraphics[width=0.3\textwidth]{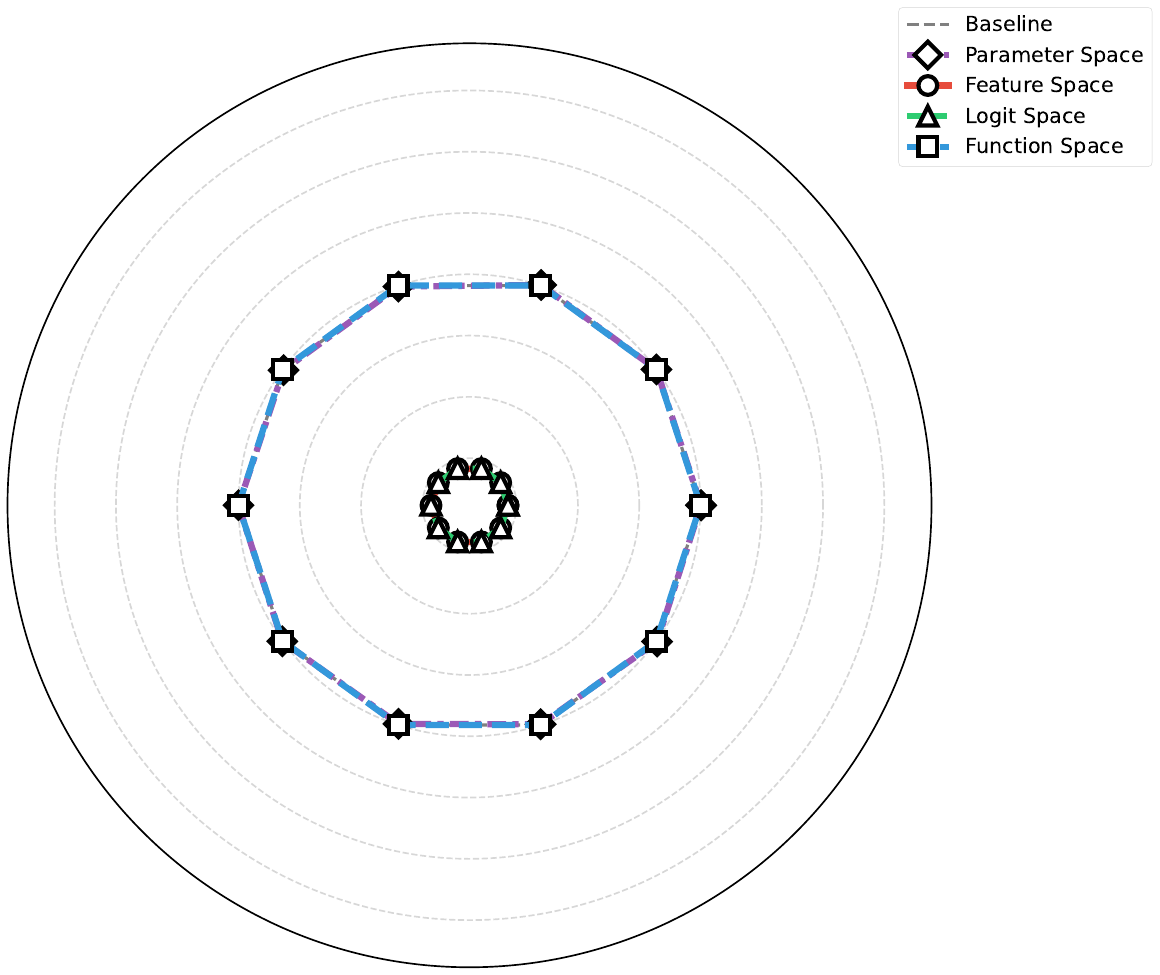}
	\end{subfigure}
	\vspace{1ex}	
	
	\begin{subfigure}[b]{\textwidth}
		\centering
		\includegraphics[width=0.3\textwidth]{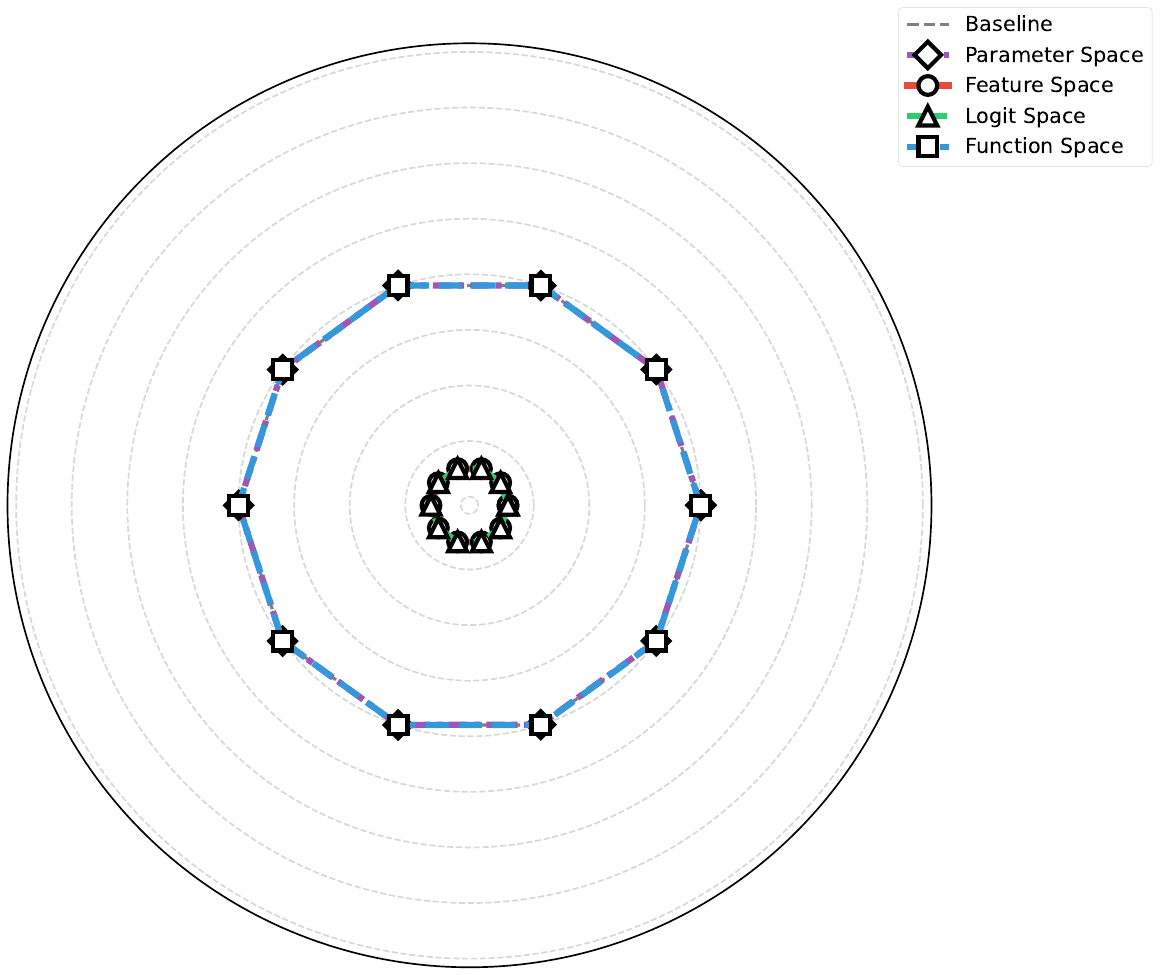}
		\includegraphics[width=0.3\textwidth]{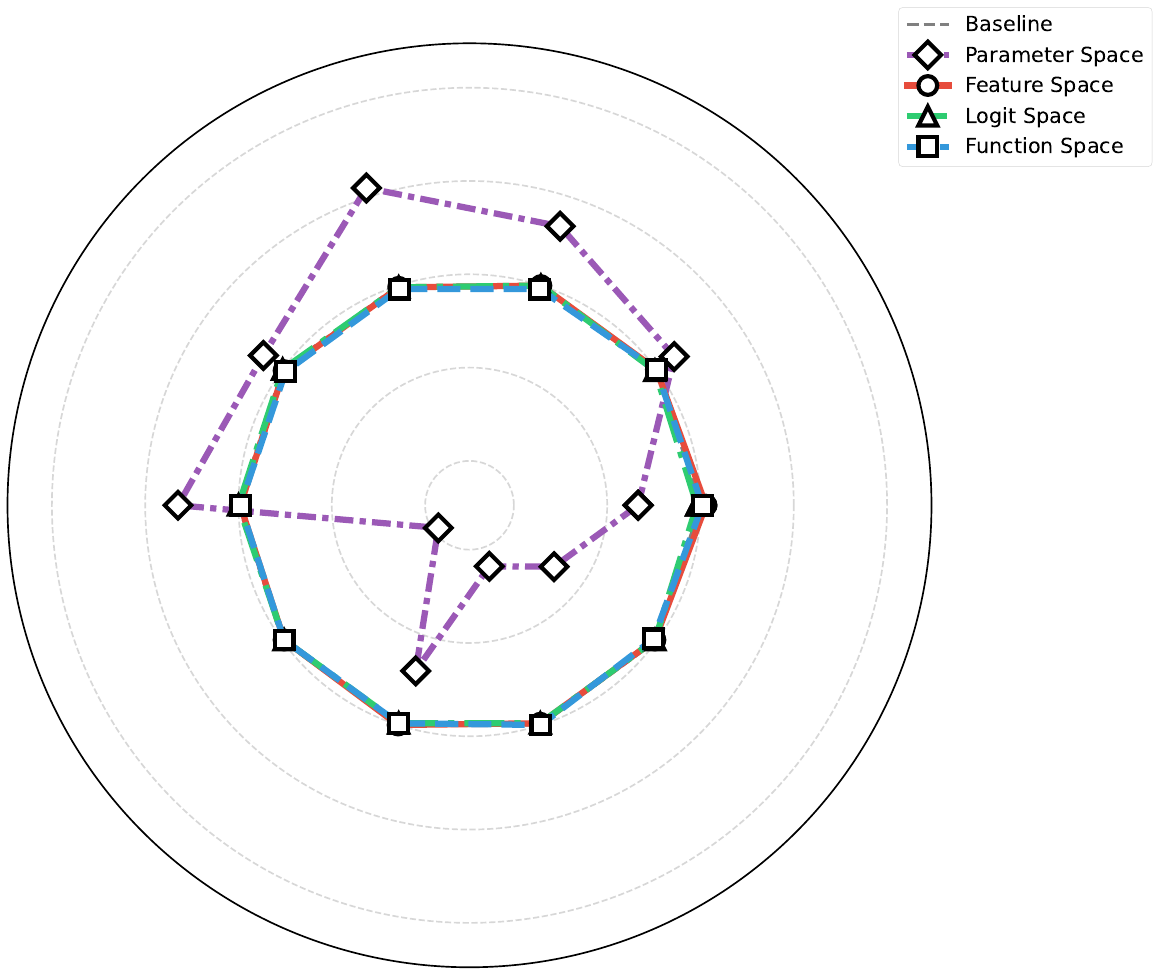}
		\label{fig:row1}
	\end{subfigure}
	\vspace{1ex}	
	
	\caption{Impact of fixed perturbations magnitude $m=0.0004$ in parameter, feature, logit, and function spaces, on the accuracy across five benchmarks (ImageNet, ImageNetV2, ImageNet-A, ImageNet-R, and ImageNet-Sketch). Function‐space perturbations induce the best accuracy across different cases, demonstrating superior robustness across datasets.}
	\label{fig:normalized_perturbation_acc}
\end{figure}

\subsection{For Finding 1 : Regularization Effects across Different Spaces}
To systematically evaluate the regularization effects across different spaces, we conducted a controlled comparative study of model performance variations under equivalent perturbation magnitudes in four distinct spaces: parameter space, feature space, logits space, and our proposed function space. 
Specifically, we generate 10 random unit-norm perturbations in each of four spaces-parameter, feature, logit, and the proposed function space and generated perturbations of fixed magnitude along those directions. After perturbation, each model is evaluated on five benchmarks: ImageNet, ImageNetV2, ImageNet-A, ImageNet-R, and ImageNet-Sketch.

In Figures~\ref{fig:loss_space_comparison} and~\ref{fig:accuracy_space_comparison}, we plot the average accuracy degradation and loss increase, respectively, for perturbation magnitudes
\[
m \,\in\, \{0.1,\,0.2,\,0.3,\,0.4,\,0.5,\,1.0\}
\]
in the feature, logit, and function spaces. Because the parameter space is far more sensitive, we rescale its magnitudes to
\(m' \;=\; 0.0004\)
to produce comparable performance changes. To enable a direct head-to-head comparison, Figure~\ref{fig:normalized_perturbation_loss} and Figure~\ref{fig:normalized_perturbation_acc} applied a perturbation of magnitude $0.0004$ in the parameter, feature, logit, and function spaces.

These experiments reveal a dramatic contrast in robustness across representational spaces. A tiny perturbation of just $0.005$ in parameter space, logit space or feature space causes a catastrophic collapse in accuracy and a steep rise in loss, whereas function spaces produce only mild performance change (As seen in the Figure~\ref{fig:normalized_perturbation_loss} and Figure~\ref{fig:normalized_perturbation_acc}). Moreover, as we increase the perturbation magnitude, both accuracy and loss curves in the feature and logit space settings become increasingly erratic (Figures~\ref{fig:loss_space_comparison} and~\ref{fig:accuracy_space_comparison}), while the corresponding curves for function space perturbations remain nearly flat. In particular, at the largest tested magnitude ($m = 1.0$) applied across all ten random directions, function‐space perturbations yield consistently higher accuracy and lower loss than either feature or logit space perturbations. (As the perturbation in the parameter space increases, model performance degrades significantly. Therefore, we fix the perturbation magnitude in the parameter space at $0.0004$.)

\subsection{For Finding 2 : Effect of FAR and FCR Objectives on OOD Robustness}

We evaluate the impact of the Functional Alignment Regularization(FAR) and the Functional Consistency Regularization(FCR)  across three benchmarks( ImageNet, WILDS‐iWildCam, and WILDS‐FMoW) using three backbone architectures (ViT-B/32, ViT-B/16, and ResNet50). For each architecture, we fine‐tune the model under three objective variants:

\begin{itemize}
	\item \textbf{FAR only:} add Functional alignment regularization during fine-tuning;
	\item \textbf{FCR only:} add Functional consistency regularization during fine-tuning;
	\item \textbf{FAR+FCR:} add Functional alignment regularization and Functional consistency regularization during fine-tuning.
\end{itemize}

\begin{figure}[htbp]
	\centering
	\begin{subfigure}[b]{\textwidth}
		\includegraphics[width=\linewidth]{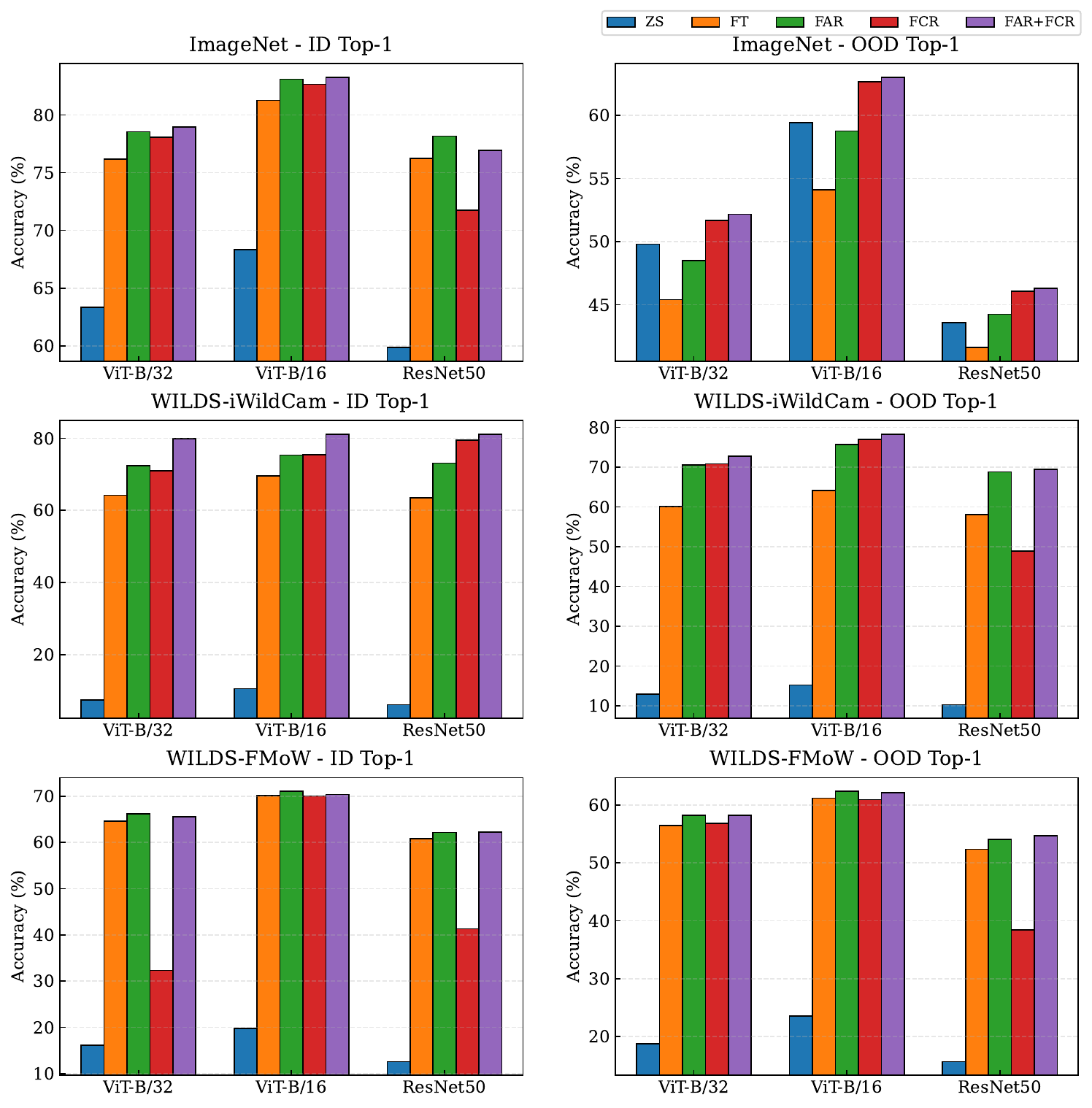}
	\end{subfigure}
	\caption{Relative performance gains over the pretrained model under full-parameter fine-tuning (FT) when incorporating the function-alignment regularizer (FT+FAR), the function-consistency regularizer (FT+FCR), and both regularizers together (FT+FAR+FCR). Both FAR and FCR individually improve upon the FT baseline, and combine FAR with FCR yields further performance gains.}
	\label{fig:far_fcr_results}
\end{figure}

Figure~\ref{fig:far_fcr_results} reports the average OOD accuracy for each variant. We make two key observations:
\begin{enumerate}
	\item \textbf{FAR significantly improves robustness:} applying FAR alone yields substantial gains in both ID and OOD accuracy compared to standard fine‐tuning, demonstrating that FAR effectively enhances its distribution‐shift resilience.
	\item \textbf{FAR and FCR provide complementary benefits:} combining FAR with FCR achieves the highest OOD accuracy, indicating that FCR further enhances robustness combined with FAR, leading to the strongest overall robustness.
\end{enumerate}

These results demonstrate that while FAR and FCR individually contribute to OOD stability, their combination provides complementary benefits that maximize the model’s ability to generalize under distribution shifts.

\subsection{Combined with Model-Ensemble based Methods}

\begin{figure}[htbp]
	\centering
	\begin{subfigure}[b]{\textwidth}
		\centering
		\includegraphics[width=0.3\textwidth]{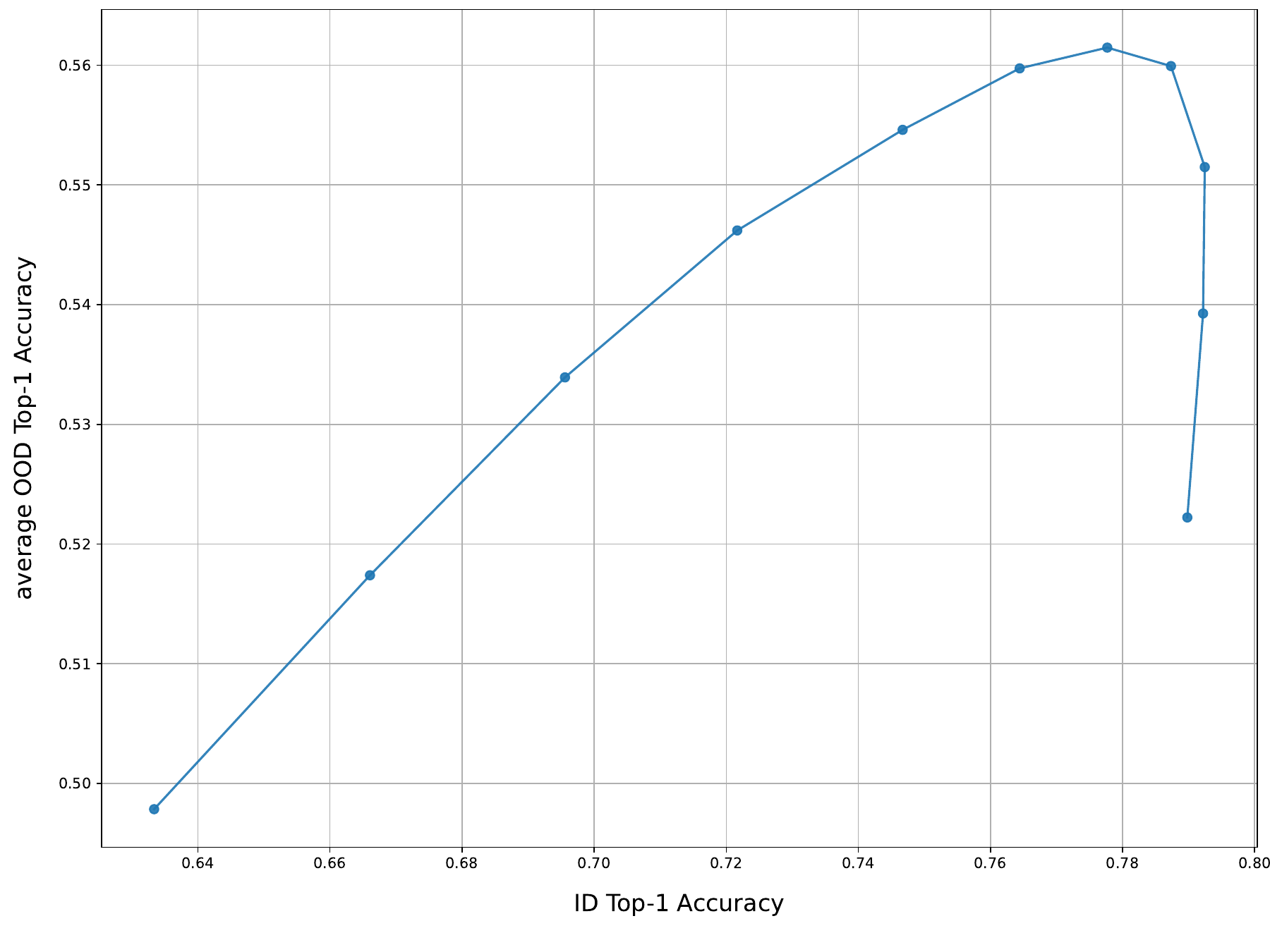}
		\includegraphics[width=0.3\textwidth]{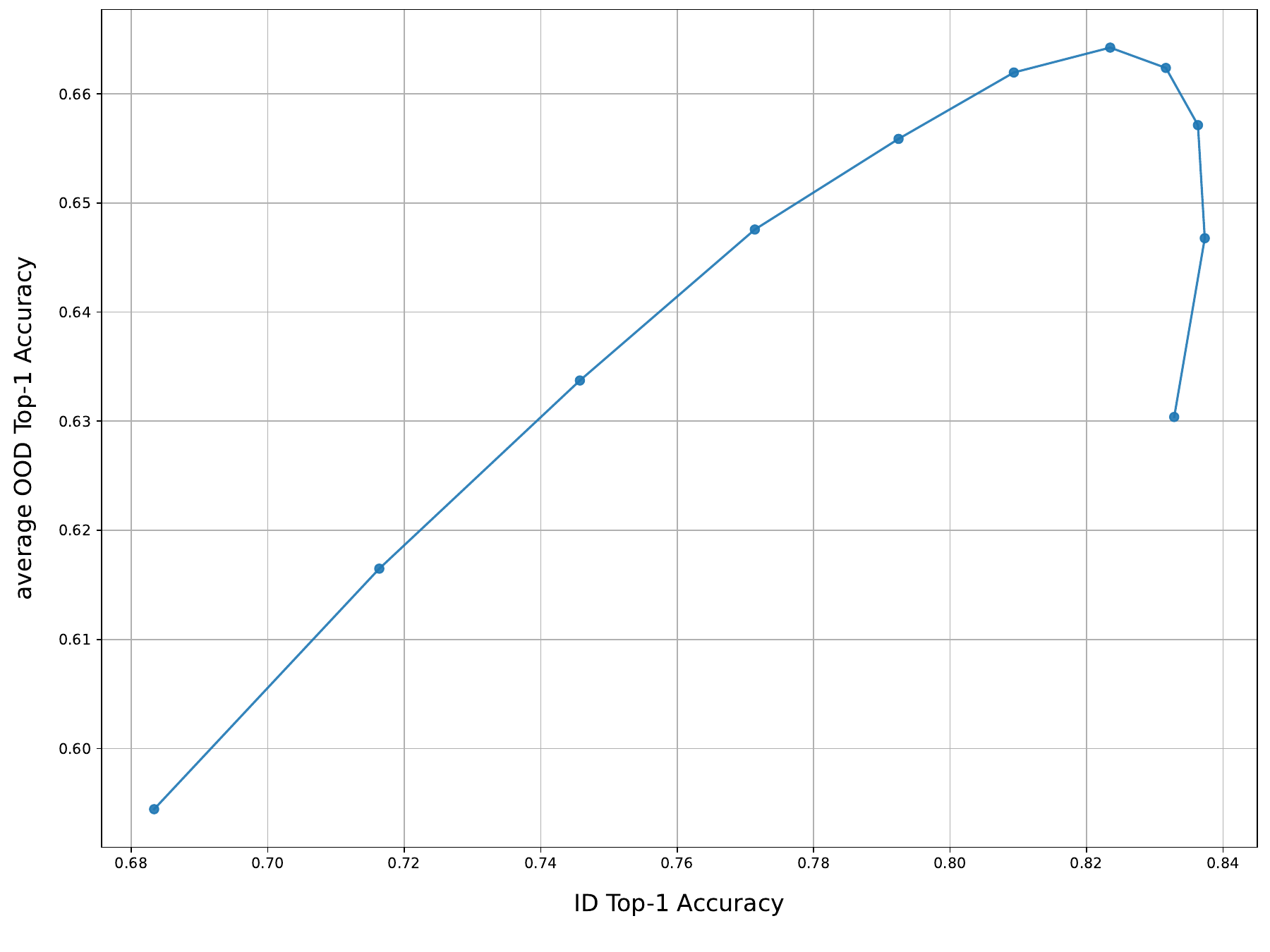}
		\includegraphics[width=0.3\textwidth]{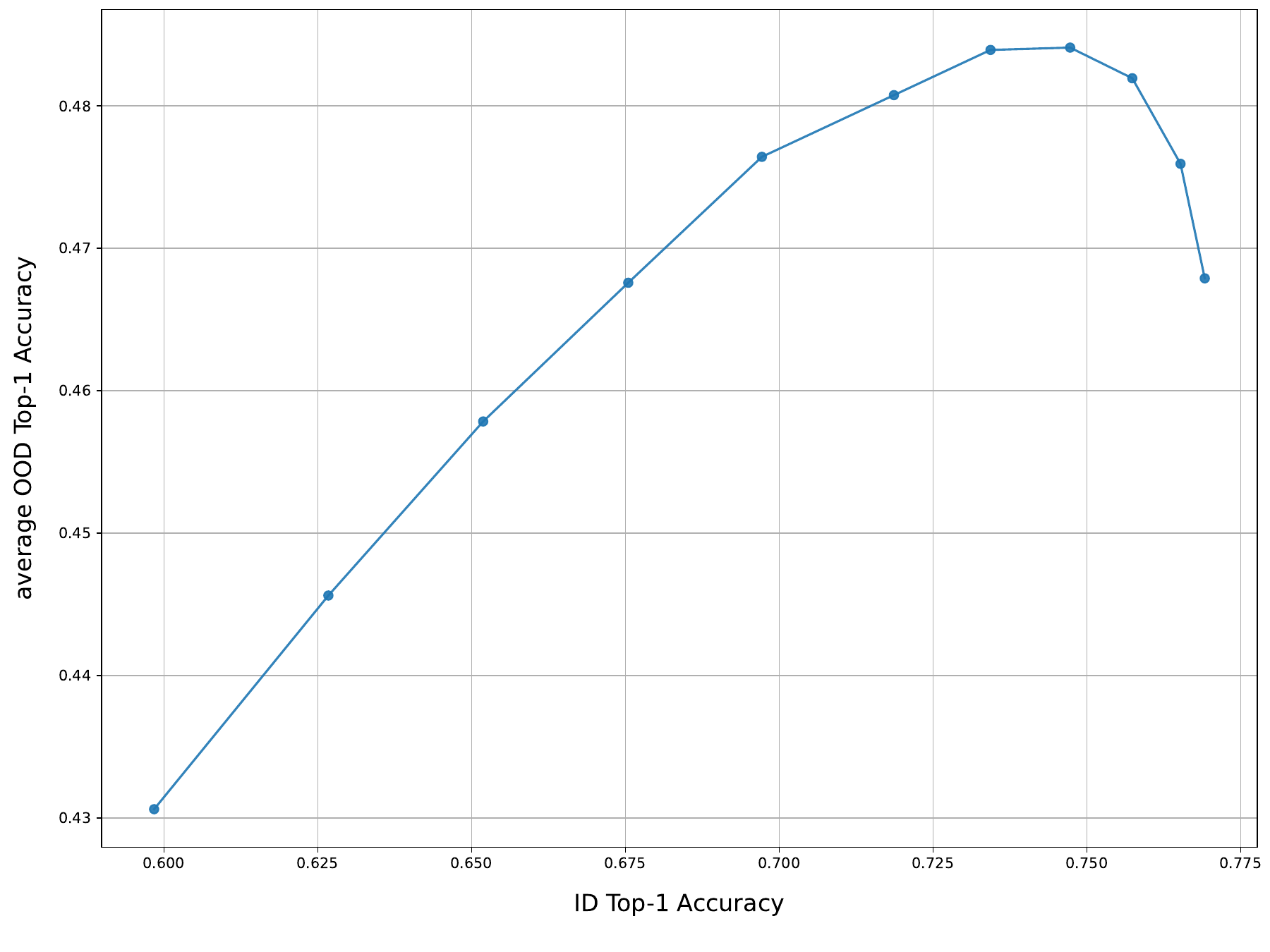}
	\end{subfigure}
	\vspace{1ex}	
	
	\caption{Ensemble results on the ViT-B/32, ViT-B/16 and ResNet-50. backbone obtained by interpolating the weights of our function-regularized fine-tuned model with those of the original pre-trained checkpoint \cite{wortsman2022robust}. This combined model yields additional gains in both in-distribution accuracy and out-of-distribution robustness, demonstrating the complementary strengths of our method and post hoc model-ensemble strategies.}
	\label{fig:ensemble}
\end{figure}

As detailed in the related work, robust fine-tuning strategies for foundation models fall into two broad paradigms: those that alter the fine-tuning process itself through regularization constraints, and those that leave the fine-tuning procedure unchanged but leverage post hoc model ensembles to improve OOD robustness \cite{wortsman2022robust,kumarfine2022}. Our method belongs to the former category, introducing a novel functional regularization term directly into the fine-tuning objective. We demonstrate that this approach yields state-of-the-art performance across multiple CLIP backbones and diverse downstream datasets, confirming its wide applicability and effectiveness. Moreover, to illustrate the complementarity of our strategy and ensemble-based strategies, we further combine our fine-tuned model with the original pre-trained checkpoint via the weight-interpolation technique proposed in \cite{wortsman2022robust}. As shown in Fig.~\ref{fig:ensemble}, this ensemble of our regularized fine-tuning and the pre-trained model produces additional improvements in both in-distribution accuracy and OOD generalization, highlighting the significant potential of integrating these two approaches.

\section{Experiments}

To validate the effectiveness of our proposed fine‐tuning method on multimodal models, we design experiments in two prototypical OOD scenarios:  
(1) \emph{class-invariant distribution shift}, where the label space remains unchanged but the input distribution varies; and  
(2) \emph{cross-class zero-shot transfer}, in which the downstream task’s label space is completely disjoint from that of the evaluation domains.  

We compare our method against a suite of state‐of‐the‐art baselines, including: zero‐shot CLIP; standard fine‐tuning (FT); Flyp \cite{goyal2023finetune}; L2SP \cite{xuhong2018explicit,mukhotifine}; LDIFS \cite{mukhotifine}; CAR‐FT \cite{mao2024context}; Lipsum‐FT \cite{nam2024lipsum}; and CaRot \cite{oh2024towards}.  
Our results show that, across both OOD settings, the proposed method consistently outperforms these strong baselines—achieving substantial gains in robustness under distribution shifts and in transfer accuracy to novel label spaces.  

\subsection{Class-Invariant Distribution Shift Results}

\subsubsection{Distribution Shift on ImageNet Variants}

In the main paper, we reported only the overall average accuracy under ID and OOD conditions. In this appendix, we extend those results by including the CLIP ViT-L/14 backbone and by providing a breakdown of OOD performance across four ImageNet variants:  
\begin{itemize}
	\item \textbf{ImageNet-V2}: a natural re-sampling of the original validation set;  
	\item \textbf{ImageNet-R}: 200 “rendition” classes (e.g., paintings, sculptures);  
	\item \textbf{ImageNet-A}: adversarially curated examples;  
	\item \textbf{ImageNet-Sketch}: hand-drawn sketches covering all 1,000 classes.  
\end{itemize}

Table~\ref{tab:ImageNet_comparison1} presents the mean ID and OOD Top-1 accuracies for four CLIP variants—ViT-L/14, ViT-B/32, ViT-B/16, and ResNet-50, under a unified fine-tuning protocol. Detailed per-benchmark results for each architecture are provided in Table~\ref{tab:imagenet_all_avg_shifts}.
These results demonstrate that our function-space regularization consistently improves OOD accuracy across all variants, with ViT-L/14 exhibiting the largest absolute gains on ImageNet-A and ImageNet-Sketch.  

\begin{table}[htbp]
	\centering
	\caption{ViT- L/14, ViT-B/32, ViT-B/16, ResNet50 on ImageNet}
	\label{tab:ImageNet_comparison1}
	\resizebox{\textwidth}{!}{
		\begin{tabular}{l|cc|cc|cc|cc}
			\toprule
			\multirow{2}{*}{Methods} 
			& \multicolumn{2}{c}{ViT-L/14}        
			& \multicolumn{2}{c}{ViT-B/32} 
			& \multicolumn{2}{c}{ViT-B/16} 
			& \multicolumn{2}{c}{ResNet-50} \\
			& ID    & OOD Avg.\ & ID    & OOD Avg.\ & ID    & OOD Avg.\ & ID    & OOD Avg.\ \\
			\midrule
			ZS        & 72.55 & 72.02 & 63.34 & 49.79     & 68.33 & 59.44     & 59.84 & 43.56     \\
			FT        & 85.26 & 67.44 & 76.17 & 45.39     & 81.27 & 54.10     & 76.23 & 41.59     \\
			FLYP  \cite{goyal2023finetune}    & 86.19 & 72.77 & 76.39    & 46.54       & 82.31    & 56.30       & 76.06    & 41.07   \\
			L2SP    \cite{xuhong2018explicit,mukhotifine}    & 85.63 & 70.53 & 77.62 & 47.46     & 75.52 & 62.37     & 64.42 & 42.42     \\
			LDIFS   \cite{mukhotifine}  & 85.19 & 73.05 & 77.33 & 50.09 & 81.43    & 60.49        & 76.38 & 41.69     \\
			CAR-FT \cite{mao2024context}   & 85.77 & 71.92 & 77.60 & 48.31     & 82.66 & 59.26     & 76.60 & 43.42     \\
			Lipsum-FT \cite{nam2024lipsum} & 86.10 & 72.68 & 76.33 & 45.27     & 81.39 & 54.62     & 76.02 & 41.77     \\
			CaRot  \cite{oh2024towards}   & 86.45 & 73.52 & 77.41 & 47.59     & 82.76 & 58.94     & 76.20   & 41.80       \\
			\midrule
			Ours      & \textbf{86.77} & \textbf{74.25} & \textbf{78.98} & \textbf{52.17}     & \textbf{83.28} & \textbf{63.04}     & \textbf{76.92} & \textbf{46.29}     \\
			\bottomrule
		\end{tabular}
	}
\end{table}

\begin{table}[htbp]
	\centering
	\caption{Top-1 ID and OOD accuracies (\%) on ImageNet variants for four CLIP backbones across different fine-tuning methods. “Avg.\ shifts” denotes the mean over all OOD datasets; higher is better.}
	\small
	\begin{tabular}{l|c|cccc|c}
		\toprule
		\multicolumn{7}{c}{\textbf{Architecture: ViT-L/14}}\\
		\midrule
		Method         & IN$\uparrow$ & IN-V2$\uparrow$ & IN-R$\uparrow$ & IN-A$\uparrow$ & IN-S$\uparrow$ & Avg.\ shifts$^{*}$$\uparrow$ \\
		\midrule
		ZS             & 75.55 & 69.85 & 87.85 & 70.76 & 59.61 & 72.02 \\
		FT             & 85.26 & 76.76 & 80.21 & 55.95 & 56.84 & 67.44 \\
		FLYP           & 86.19 & 78.21 & 83.81 & 68.85 & 60.20 & 72.77 \\
		L2SP           & 85.63 & 77.95 & 82.35 & 63.97 & 57.86 & 70.53 \\
		LDIFS          & 85.19 & 78.46 & 83.97 & 69.42 & 60.36 & 73.05 \\
		CAR-FT         & 85.77 & 76.35 & 84.59 & 66.71 & 60.05 & 71.92 \\
		Lipsum-FT      & 86.10 & 77.65 & 85.67 & 67.00 & 60.41 & 72.68 \\
		CaRot          & 86.45 & 78.87 & 86.13 & 68.14 & 60.75 & 73.52 \\
		Ours           & 86.77 & 78.29 & 86.61 & 69.51 & 62.57 & 74.25 \\
		\midrule
		\multicolumn{7}{c}{\textbf{Architecture: ViT-B/32}}\\
		\midrule
		Method         & IN$\uparrow$ & IN-V2$\uparrow$ & IN-R$\uparrow$ & IN-A$\uparrow$ & IN-S$\uparrow$ & Avg.\ shifts$^{*}$$\uparrow$ \\
		\midrule
		ZS             & 63.34 & 55.95 & 69.35 & 31.55 & 42.30 & 49.79 \\
		FT             & 76.17 & 64.47 & 57.17 & 20.28 & 39.62 & 45.39 \\
		FLYP           & 76.39 & 64.94 & 58.12 & 21.32 & 41.80 & 46.54 \\
		L2SP           & 77.62 & 65.37 & 60.15 & 23.40 & 40.92 & 47.46 \\
		LDIFS          & 77.33 & 61.67 & 65.05 & 32.82 & 40.80 & 50.09 \\
		CAR-FT         & 77.60 & 65.94 & 61.71 & 23.76 & 41.83 & 48.31 \\
		Lipsum-FT      & 76.33 & 64.66 & 57.15 & 20.18 & 39.07 & 45.27 \\
		CaRot          & 77.41 & 66.17 & 61.51 & 21.17 & 41.51 & 47.59 \\
		Ours           & 78.75 & 67.61 & 66.30 & 28.92 & 45.85 & 52.17 \\
		\midrule
		\multicolumn{7}{c}{\textbf{Architecture: ViT-B/16}}\\
		\midrule
		Method         & IN$\uparrow$ & IN-V2$\uparrow$ & IN-R$\uparrow$ & IN-A$\uparrow$ & IN-S$\uparrow$ & Avg.\ shifts$^{*}$$\uparrow$ \\
		\midrule
		ZS             & 68.33 & 61.91 & 77.70 & 49.93 & 48.22 & 59.44 \\
		FT             & 81.27 & 70.60 & 66.03 & 36.08 & 45.68 & 54.10 \\
		FLYP           & 82.31 & 71.44 & 70.03 & 36.55 & 47.18 & 56.30 \\
		L2SP           & 75.52 & 68.35 & 78.73 & 52.12 & 50.26 & 62.37 \\
		LDIFS          & 81.43 & 67.01 & 73.57 & 50.11 & 51.03 & 60.43 \\
		CAR-FT         & 82.66 & 72.90 & 71.05 & 43.85 & 49.22 & 59.26 \\
		Lipsum-FT      & 81.39 & 71.22 & 65.46 & 36.06 & 45.74 & 54.62 \\
		CaRot          & 82.76 & 73.06 & 71.03 & 41.05 & 48.63 & 58.94 \\
		Ours           & 83.28 & 73.72 & 75.54 & 50.16 & 52.73 & 63.04 \\
		\midrule
		\multicolumn{7}{c}{\textbf{Architecture: ResNet-50}}\\
		\midrule
		Method         & IN$\uparrow$ & IN-V2$\uparrow$ & IN-R$\uparrow$ & IN-A$\uparrow$ & IN-S$\uparrow$ & Avg.\ shifts$^{*}$$\uparrow$ \\
		\midrule
		ZS             & 59.84 & 52.88 & 60.72 & 23.21 & 35.44 & 43.56 \\
		FT             & 76.23 & 64.52 & 50.55 & 17.84 & 33.44 & 41.59 \\
		FLYP           & 76.06 & 63.91 & 50.07 & 17.55 & 32.75 & 41.07 \\
		L2SP           & 64.42 & 56.06 & 58.98 & 20.28 & 34.34 & 42.42 \\
		LDIFS          & 76.38 & 64.55 & 50.92 & 17.81 & 33.47 & 41.69 \\
		CAR-FT         & 76.60 & 65.09 & 53.26 & 20.33 & 34.98 & 43.42 \\
		Lipsum-FT      & 76.02 & 64.86 & 50.40 & 17.76 & 34.04 & 41.77 \\
		CaRot          & 76.20 & 65.07 & 49.55 & 18.09 & 34.49 & 41.80 \\
		Ours           & 76.92 & 66.46 & 59.16 & 21.60 & 39.93 & 46.29 \\
		\bottomrule
	\end{tabular}
	\label{tab:imagenet_all_avg_shifts}
\end{table}

\subsubsection{Distribution Shift on WILDS-iWildCam Dataset}
WILDS-iWildCam is a large‐scale camera‐trap wildlife dataset designed to evaluate geographic domain shift. Each “domain” corresponds to a unique trap location: the training split comprises approximately 201 399 images from 323 locations, while the test split contains 60 029 images from 91 held‐out locations, covering 182 animal species. By construction, no location in the test set overlaps with those in training, providing a rigorous benchmark for unseen environments. Table~\ref{tab:WILDS-iWildCam_comparison1} presents the ID and OOD Top-1 accuracies for all four CLIP backbones—ViT-L/14, ViT-B/32, ViT-B/16, and ResNet-50, under our fine-tuning protocol, with the ViT-L/14 results newly added to complete the comparison across architectures.  

\begin{table}[t]
	\centering
	\caption{Performance on WILDS-iWildCam: ID and OOD metrics for ViT-L/14, ViT-B/32, ViT-B/16, and ResNet50}
	\label{tab:WILDS-iWildCam_comparison1}
	\small
	\resizebox{\textwidth}{!}{
		\begin{tabular}{lcccccccccccc}
			\toprule
			Method  & \multicolumn{3}{c}{ViT-L/14} & \multicolumn{3}{c}{ViT-B/32} & \multicolumn{3}{c}{ViT-B/16} & \multicolumn{3}{c}{ResNet50} \\
			\cmidrule(lr){2-4} \cmidrule(lr){5-7} \cmidrule(lr){8-10} \cmidrule(lr){11-13}
			& Acc    & Recall   & F1       & Acc     & Recall   & F1       & Acc     & Recall   & F1 & Acc     & Recall   & F1       \\
			\midrule
			\multicolumn{13}{c}{\textbf{ID Metrics}} \\
			ZS       & 10.04 & 11.54 & 11.77 & 7.46  & 8.64  & 8.03  & 10.55 & 10.22 & 8.81  & 6.09  & 8.23  & 7.23  \\
			FT      & 76.24 & 53.28 & 53.86 &    64.21    &    30.21   &  29.55     &   69.59     &    31.53   &  32.77     &     63.49  &    22.74   &    23.10   \\
			FLYP \cite{goyal2023finetune} & 76.85 & 53.76 & 54.70 &   64.31    &   30.47    &  59.76     &   71.47    &   32.19    &  32.51     &   63.21   &   20.14    &    24.19   \\
			L2SP \cite{xuhong2018explicit,mukhotifine} & 71.67 & 40.60 & 41.10 & 73.05 & 35.29 & 35.78 & 75.79 & 38.16 & 38.52 & 67.74 & 26.25 & 25.35 \\
			LDIFS  \cite{mukhotifine}   & 76.36 & 54.91 & 54.88 & 77.76 & 49.68 & 44.30 & 80.69 & 48.15 & 48.31 & 78.75 & 41.34 & 42.56 \\
			CAR-FT \cite{mao2024context}  & 77.05 & 56.07 & 56.73 & 77.00 & 40.29 & 40.87 & 80.35 & 44.92 & 45.64 & 78.88 & 42.08 & 42.42 \\
			Lipsum-FT \cite{nam2024lipsum} & 76.44 & 55.18 & 55.39 & 77.34 & 40.77 & 41.29 & 79.98 & 45.86 & 45.62 & 78.53 & 40.95 & 41.08 \\
			CaRot \cite{oh2024towards} & 77.11 & 55.79 & 56.07  &    77.53   &   40.95    &   41.57    &   79.47    &    45.14   &    45.27   &    78.61   &  41.83     &   41.09    \\
			\midrule
			Ours     &  \textbf{77.43} & 53.98 & 55.12 & \textbf{79.87} & {40.99} & {41.70} & \textbf{81.06} & \textbf{51.09} & \textbf{51.69} & \textbf{81.13} & \textbf{46.43} & \textbf{45.94} \\
			\midrule
			\multicolumn{13}{c}{\textbf{OOD Metrics}} \\
			ZS       & 15.46 & 15.85 & 12.10 & 12.89 & 7.80  & 7.32  & 15.23 & 13.24 & 10.99 & 10.30 & 8.39  & 6.22  \\
			FT       & 76.34 & 43.69 & 45.35 &  60.15    &   19.40    &  20.41    &    64.13   &    21.11   &   22.30    &   58.14    &  15.96    &   15.43    \\
			FLYP   \cite{goyal2023finetune}   & 76.55 & 43.56 & 45.95 &   60.21    &   20.12    &   20.08    &   65.90     &   22.15    &    22.08   &   27.61    &   17.49    &  18.05     \\
			L2SP \cite{xuhong2018explicit,mukhotifine}     & 74.52 & 35.84 & 35.21 & 68.94 & 27.87 & 28.74 & 68.03 & 27.33 & 26.07 & 63.52 & 17.70 & 17.37 \\
			LDIFS \cite{mukhotifine}   & 76.71 & 40.15 & 39.77 & 65.21 & 22.49 & 22.34 & 74.25 & 36.03 & 34.36 & 69.21 & 27.60 & 27.17 \\
			CAR-FT  \cite{mao2024context} & 77.79 & 43.58 & 44.29 & 65.55 & 27.88 & 26.81 & 74.85 & 34.47 & 33.67 & 69.31 & 29.42 & 29.03 \\
			Lipsum-FT \cite{nam2024lipsum}  & 77.61 & 43.62 & 44.61 & 63.77 & 24.48 & 24.75 & 75.16 & 38.01 & 36.91 & 69.17 & 28.62 & 25.53 \\
			CaRot \cite{oh2024towards}   & 77.65 & 43.55 & 44.57 &    65.74   &   28.91    &   28.88    &   75.33    &   38.13    &   37.24    &  69.14   &   29.51    &    26.33   \\
			\midrule
			Ours     &  \textbf{77.81} & 42.62 & 43.44 & \textbf{72.49} & \textbf{31.58} & \textbf{32.41} & \textbf{78.29} & \textbf{38.56} & \textbf{39.31} & \textbf{69.44} & \textbf{31.52} & \textbf{29.07} \\
			\bottomrule
		\end{tabular}
	}
\end{table}

\subsubsection{Distribution Shift on  WILDS-FMoW Dataset}
WILDS-FMoW is a large‐scale satellite imagery dataset designed to evaluate both geographic and temporal domain shifts in land‐use classification. It comprises high‐resolution images drawn from the Functional Map of the World (FMoW) corpus, spanning 62 countries and multiple time periods between 2014 and 2017. The training split includes over 340 000 images from 45 countries, while the OOD test split contains approximately 72 000 images from 17 held‐out countries and later timestamps, covering 62 land‐use categories such as “airport,” “meadow,” and “bridge.” By withholding entire countries and time periods at test time, WILDS-FMoW provides a challenging benchmark for evaluating a model’s ability to generalize under realistic distributional shifts.  
Table~\ref{tab:WILDS_FMoW_comparison} summarizes the ID and OOD Top‐1 accuracies of four CLIP-based architectures (ViT-L/14, ViT-B/32, ViT-B/16, and ResNet-50) under our unified fine‐tuning protocol, demonstrating that our function‐space regularization consistently improves robustness across both spatial and temporal shifts.  

\begin{table}[htbp]
	\centering
	\caption{ViT-B/32 ViT-B/16 ResNet50 on WILDS-FMoW}
	\label{tab:WILDS_FMoW_comparison}
	\resizebox{\textwidth}{!}{
		\begin{tabular}{l|ccc|ccc|ccc|ccc}
			\toprule
			\multirow{2}{*}{Methods} 
			& \multicolumn{3}{c}{ViT-L/14} 
			& \multicolumn{3}{c}{ViT-B/32} 
			& \multicolumn{3}{c}{ViT-B/16} 
			& \multicolumn{3}{c}{ResNet-50} \\
			& ID & ID   & OOD & ID & ID   & OOD & ID & ID   & OOD & ID & ID   & OOD \\
			\midrule
			ZS        & 26.06 & 27.13 & 28.06 & 15.82 & 16.41 & 18.76 & 19.15 & 20.37 & 23.53 & 12.19 & 12.91 & 15.67 \\
			FT        & 75.15 & 73.27 & 66.32 & 65.25 & 63.93 & 56.45 & 70.86 & 69.37 & 61.19 & 61.57 & 59.98 & 52.35 \\
			FLYP      & 75.02 & 73.04 & 65.98 & 65.76 & 63.81 & 56.99 & 70.96 & 68.44 & 61.10 & 60.43 & 58.57 & 50.68 \\
			L2SP      & 67.42 & 66.31 & 60.44 & 60.66 & 58.77 & 52.87 & 65.65 & 64.57 & 57.26 & 46.79 & 47.03 & 41.30 \\
			LDIFS     & 74.96 & 73.05 & 66.22 & 65.76 & 63.81 & 56.99 & 70.97  & 68.97 & 60.82 & 60.97 & 59.18 & 51.57 \\
			CAR-FT    & 74.88 & 73.52 & 66.72 & 65.17 & 63.87 & 56.92 & 70.51 & 68.85 & 61.41 & 61.77 & 60.39 & 52.54 \\
			Lipsum-FT & 74.91 & 72.79 & 66.08 & 65.41 & 64.15 & 56.52 & 70.96 & 68.80 & 60.70 & 61.62 & 59.52 & 51.85 \\
			CaRot     & 52.67 & 52.19 & 48.58 & 40.55 & 40.28 & 37.53 & 47.12 & 46.00 & 43.08 & 30.54 & 30.49 & 28.73 \\
			\midrule
			Ours      & \textbf{75.35} & \textbf{73.08} & \textbf{67.28} & \textbf{66.57} & \textbf{64.60} & \textbf{58.22} & \textbf{71.33} & \textbf{69.43} & \textbf{62.17} & \textbf{62.88} & \textbf{61.61} & \textbf{54.66} \\
			\bottomrule
		\end{tabular}
	}
\end{table}

\subsection{Cross-Class Transfer}
We further assess our method’s ability to transfer to domains whose label sets are entirely disjoint from those seen during fine-tuning. Following the protocol of Mukhoti et al. \cite{mukhotifine}, we fine-tune each CLIP variant end-to-end on the EuroSAT remote-sensing dataset, which contains ten land-use classes (e.g., “AnnualCrop,” “Forest,” “Residential”) that do not overlap with any of our evaluation domains. Training uses AdamW (peak learning rate $1\times10^{-7}$ with a 4-epoch linear warm-up followed by cosine decay), weight decay $0.5$, batch size 128, for 20 epochs.  
We then evaluate on eleven held-out datasets spanning both general-vision and specialized benchmarks:  
\begin{itemize}
	\item \textbf{CIFAR-10 / CIFAR-100}: small natural images in 10 and 100 classes, respectively;  
	\item \textbf{DTD}: Describable Textures Dataset, with 47 texture categories;  
	\item \textbf{GTSRB}: German Traffic Sign Recognition Benchmark, 43 sign classes;  
	\item \textbf{RESISC45}: remote-sensing scenes across 45 land-use categories;  
	\item \textbf{STL-10}: 10-class natural images with fewer training samples;  
	\item \textbf{ImageNet variants}: ImageNet, ImageNetV2, ImageNet-R, ImageNet-A, ImageNet-Sketch.  
\end{itemize}

Tables~\ref{tab:cross_class_transfer_b32} and~\ref{tab:cross_class_transfer_b16} report Top-1 accuracy for ViT-B/32 and ViT-B/16, respectively. Notably, most alternative fine-tuning methods degrade performance relative to the zero-shot CLIP baseline, underscoring the challenge of cross-class transfer on this benchmark. Our method, by contrast, consistently matches or slightly exceeds zero-shot accuracy—achieving small but positive gains on datasets such as CIFAR-100, RESISC45, while never falling below the zero-shot level on any test set. This behavior highlights both the difficulty of the task and the robustness of our function-space regularization, demonstrating its strong potential for preserving out-of-distribution performance in real-world zero-shot transfer scenarios.  

\begin{table}[ht]
	\centering
	\caption{zero-shot classification on different datasets on VIT-B/32}
	\label{tab:cross_class_transfer_b32}
	\resizebox{0.9\textwidth}{!}{
		\begin{tabular}{lccccccccc}
			\toprule
			Datasets   & ZS    & FT  & FLYP  & L2SP & LDIFS & CAR-FT  & Lipsum-FT & CaRot & Ours  \\
			\midrule
			CIFAR10    & 89.83 & 82.29 & 82.54 & 83.90 & 87.18 & 83.77 & 83.96 & 89.85 & \textbf{89.87} \\
			CIFAR100   & 64.23 & 57.95 & 57.44 & 59.58 & 62.13 & 58.93 & 59.14 & 64.33 & \textbf{64.51} \\
			DTD        & 44.41 & 42.61 & 43.78 & 43.56 & 43.19 & 43.30 & 43.24 & 44.41 & \textbf{44.43} \\
			GTSRB      & 32.60 & 31.65 & 32.09 & 31.67 & 33.10 & 31.74 & 31.76 & 32.57 & {32.61} \\
			RESISC45   & 60.26 & 56.84 & 54.81 & 57.89 & 60.48 & 57.92 & 57.60  & 60.87 & \textbf{62.21} \\
			STL10      & 97.13 & 96.08 & 96.26 & 96.16 & 96.85 & 96.13 & 96.10 & 97.08 & \textbf{97.16} \\
			ImageNet   & 63.34 & 62.44 & 62.85 & 62.61 & 62.89 & 62.63 & 62.66 & 63.21 & \textbf{63.36} \\
			ImageNetV2 & 55.95 & 54.68 & 55.12 & 54.90 & 55.33 & 54.86 & 54.90 & 55.78 & \textbf{55.96} \\
			ImageNetR  & 69.35 & 68.58 & 68.50 & 68.70 & 69.18 & 68.60 & 68.67 & 69.22 & \textbf{69.41} \\
			ImageNetA  & 31.55 & 30.59 & 29.69 & 30.87 & 30.56 & 30.87 & 30.83 & 30.80 & \textbf{31.57} \\
			ImageNetS  & 42.30 & 40.98 & 41.30 & 41.20 & 41.78 & 41.17 & 41.23 & 42.09 & \textbf{42.31} \\
			\bottomrule
		\end{tabular}
	}
\end{table}

\begin{table}[htbp]
	\centering
	\caption{zero-shot classification on different datasets on VIT-B/16}
	\label{tab:cross_class_transfer_b16}
	\resizebox{0.9\textwidth}{!}{
		\begin{tabular}{lccccccccc}
			\toprule
			Datasets   & ZS    & FT  & FLYP  & L2SP & LDIFS & CAR-FT  & Lipsum-FT & CaRot & Ours  \\
			\midrule
			CIFAR10    & 90.77 & 88.66 & 89.74 & 90.49 & 90.76 & 90.44 & 90.39 & 90.89 & \textbf{90.91}  \\
			CIFAR100   & 66.95 & 63.55 & 66.01 & 66.88 & 67.55 & 66.91 & 66.58 & 67.31 & \textbf{67.65} \\
			DTD        & 44.68 & 44.84 & 44.95 & 45.43 & 45.53 & 45.48 & 45.53 & 45.48 & \textbf{45.55} \\
			GTSRB      & 43.37 & 42.94 & 43.98 & 43.39 & 43.91 & 43.75 & 43.45 & 44.51 & \textbf{43.54}  \\
			RESISC45   & 66.38 & 58.68 & 59.60 & 61.56 & 62.30 & 61.48 & 61.27 & 66.06 & \textbf{66.94}  \\
			STL10      & 98.25 &97.80 & 98.23 & 98.05 & 98.08 & 98.05 & 98.06 & 98.21 & \textbf{98.27} \\
			ImageNet   & 68.33 &67.71 & 68.09 & 68.09 & 68.15 & 67.98 & 68.06 & 68.31 & \textbf{68.34}  \\
			ImageNetV2 & 61.91 & 61.33 & 61.57 & 61.65 & 61.46 & 61.55 & 61.56 & 61.86 & \textbf{61.90}  \\
			ImageNetR  & 77.71 & 76.52 & 76.84 & 76.94 & 77.18 & 76.88 & 76.93 & 77.61 & \textbf{77.73} \\
			ImageNetA  & 49.93 &47.53 & 48.47 & 49.15 & 48.81 & 48.97 & 48.99 & 49.73 & \textbf{50.21} \\
			ImageNetS  & 48.22 & 47.51 & 47.69 & 47.80 & 47.82 & 47.78 & 47.72 & 48.06 & \textbf{48.19}  \\
			\bottomrule
		\end{tabular}
	}
\end{table}

Overall, across both settings, our approach consistently outperforms naive FT and remains competitive with or better than strong regularization-based baselines like LDIFS, L2SP, CAR-FT, and Lipsum-FT. These results validate the effectiveness of our fine-tuning strategy for robust cross-domain transfer on top of the CLIP backbone.

\subsection{Ablation Study}
Table~\ref{tab:Ablation_All}  and Figure~\ref{fig:far_fcr_results} reports Top-1 ID and OOD accuracies for three CLIP backbones (ViT-B/32, ViT-B/16, ResNet-50) under four training variants: standard fine‐tuning (FT), FAR only, FCR only, and FAR+FCR, confirming removing either component results in noticeable degradation in either OOD robustness or ID accuracy, demonstrating the complementary roles of the two regularization strategies.
We observe that introducing functional regularization alone substantially enhances the model’s generalization under distributional shifts, underscoring the critical role of function-space constraints in preserving robust OOD performance. Moreover, functional consistency regularization further bolsters stability against input perturbations, yielding additional gains across diverse OOD datasets.

\begin{table}[htbp]
	\centering
	\caption{Top-1 ID and OOD accuracies (\%) for three CLIP backbones (ViT-B/32, ViT-B/16, ResNet-50) under four training variants: standard fine-tuning (FT), functional regularization (FAR), consistency regularization (FCR), and combined FAR+FCR.}
	
	\label{tab:Ablation_All}
	\begin{tabular}{l|cc|cc|cc}
		\toprule
		\multirow{2}{*}{Methods} 
		& \multicolumn{2}{c}{ViT-B/32} 
		& \multicolumn{2}{c}{ViT-B/16} 
		& \multicolumn{2}{c}{ResNet50} \\
		& ID & OOD & ID & OOD & ID & OOD \\
		\midrule
		\multicolumn{7}{c}{\textbf{ImageNet}} \\
		ZS        & 63.34 & 49.79  & 68.33 & 59.44     & 59.84 & 43.56     \\
		FT        & 76.17 & 45.39  & 81.27 & 54.10     & 76.23 & 41.59     \\
		FAR      & 78.55 & 48.50 & 83.10 & 58.77 & 78.16 & 44.24 \\
		FCR      & 78.06 & 51.67 & 82.65 & 62.66 & 71.73 & 46.06 \\
		FAR + FCR      & 78.98 & 52.17 & 83.28 & 63.04 & 76.92 & 46.29 \\
		\midrule
		\multicolumn{7}{c}{\textbf{WILDS-iWildCam}} \\
		ZS        & 7.46 & 12.89 & 10.55 & 15.23 & 6.09 & 10.30 \\ 
		FT        & 64.21 & 60.15 & 69.59 & 64.13 & 63.49 & 58.14  \\ 
		FAR      & 72.33 & 70.56 & 75.32 & 75.74 & 73.09 & 68.79 \\
		FCR      & 70.97 & 70.84 & 75.46 & 76.99 & 79.51 & 48.93 \\
		FAR + FCR     & 79.87 & 72.49 & 81.06 & 78.29 & 81.13 & 69.44      \\
		\midrule
		\multicolumn{7}{c}{\textbf{WILDS-FMoW}} \\
		ZS        & 16.11 & 18.76 & 19.76 & 23.53 & 12.55 & 15.67 \\
		FT        & 64.59 & 56.45 & 70.12 & 61.19 & 60.78 & 52.35 \\
		FAR      & 66.15 & 58.24 & 70.37 & 62.38 & 62.18 & 54.08 \\
		FCR      & 32.32 & 56.82 & 70.06 & 60.93 & 41.32 & 38.47 \\   
		FAR + FCR     & 65.59 & 58.22 & 70.38 & 62.17 & 62.25 & 54.66 \\            
		\bottomrule
	\end{tabular}
\end{table}

\newpage
\nocite{*}
\bibliography{reference}

\end{document}